\documentclass{article}

\usepackage{microtype}
\usepackage{booktabs} 
\usepackage{adjustbox}
\usepackage{amsmath,amssymb,amsfonts,mathtools, subcaption}
\usepackage{algorithmic}
\usepackage{multirow}
\usepackage{graphicx}
\usepackage{algorithm}
\usepackage{color}
\usepackage{bm}
\usepackage[usenames,dvipsnames,svgnames]{xcolor}

\DeclareMathOperator*{\argmax}{argmax}

\newtheorem{assumption}{Assumption}

\newtheorem{theorem}{Theorem}[section]

\newtheorem{lemma}[theorem]{Lemma}
\newtheorem{remark}[theorem]{Remark}

\graphicspath{{figures/}}

\def \tw{\tilde{w}}
\def \S{\mathcal{S}}

\def \E{\mathbb{E}}

\def \bdelta{\bm{\delta}}

\def \tg{\tilde{g}}

\usepackage{hyperref}

\usepackage[accepted]{icml2022}
\icmltitlerunning{Generalized Federated Learning via Sharpness Aware Minimization}

\begin{document}
	
	\twocolumn[\icmltitle{Generalized Federated Learning via Sharpness Aware Minimization}
	
	
	
	\icmlsetsymbol{equal}{*}
	
	\begin{icmlauthorlist}
		\icmlauthor{Zhe Qu}{equal,to}
		\icmlauthor{Xingyu Li}{equal,goo}
		\icmlauthor{Rui Duan}{to}
		\icmlauthor{Yao Liu}{ed}
		\icmlauthor{Bo Tang}{goo}
		\icmlauthor{Zhuo Lu}{to}
	\end{icmlauthorlist}
	
	\icmlaffiliation{to}{Department of Electrical Engineering, University of South Florida, Tampa, USA}
	\icmlaffiliation{goo}{Department of Electrical and Computer Engineering, Mississippi State University, Starkville, USA}
	\icmlaffiliation{ed}{Department of Computer Science and Engineering, University of South Florida, Tampa, USA}
	
	\icmlcorrespondingauthor{Zhe Qu}{zhequ@usf.edu}
	
	\icmlkeywords{Machine Learning, ICML}
	
	\vskip 0.3in
	
	]

	
	
	\printAffiliationsAndNotice{\icmlEqualContribution} 
	
	\begin{abstract}
		Federated Learning (FL) is a promising framework for performing privacy-preserving, distributed learning with a set of clients. However, the data distribution among clients often exhibits non-IID, i.e., distribution shift, which makes efficient optimization difficult. To tackle this problem, many FL algorithms focus on mitigating the effects of data heterogeneity across clients by increasing the performance of the global model. However, almost all algorithms leverage Empirical Risk Minimization (ERM) to be the local optimizer, which is easy to make the global model fall into a sharp valley and increase a large deviation of parts of local clients. Therefore, in this paper, we revisit the solutions to the distribution shift problem in FL with a focus on local learning generality. To this end, we propose a general, effective algorithm, \texttt{FedSAM}, based on Sharpness Aware Minimization (SAM) local optimizer, and develop a momentum FL algorithm to bridge local and global models, \texttt{MoFedSAM}. Theoretically, we show the convergence analysis of these two algorithms and demonstrate the generalization bound of \texttt{FedSAM}. Empirically, our proposed algorithms substantially outperform existing FL studies and significantly decrease the learning deviation.
	\end{abstract}
	
	\section{Introduction}\label{Sec:introduction}
	Federated Learning (FL) \cite{mcmahan2017communication} is a collaborative training framework that enables a large number of clients, which can be phones, network sensors, or alternative local information sources \cite{kairouz2019advances, mohri2019agnostic}. FL trains machine learning models without transmitting client data over the network, and thus it can protect data privacy at some basic levels. Two important settings are introduced in FL \cite{kairouz2019advances}: the \textit{cross-device} FL and the \textit{cross-silo} FL. The cross-silo FL is related to a small number of reliable clients, e.g., medical or financial institutions. By contrast, the cross-device FL includes a large number of clients, e.g., billion-scale android phones \cite{hard2018federated}. In cross-device FL, clients are usually deployed in various environments. It is unavoidable that the distribution of the local dataset of each client varies considerably and incurs a distribution shift problem, highly degrading the learning performance.
	
	Many existing FL studies focus on the distribution shift problem mainly based on the following three directions: (i) The most popular solution to address this problem is to set the number of local training epochs performed between each communication round \cite{Li2020On, yang2021achieving}. (ii) Many algorithmic solutions in \cite{li2018federated, karimireddy2020scaffold, acar2021federated} mainly focus on mitigating the influence of heterogeneity across clients via giving a variety of proximal terms to control the local model updates close to the global model. (iii) Knowledge distillation based techniques \cite{lin2020ensemble, gong2021ensemble, zhu2021data} aggregate locally-computed logits for building global models, helping eliminate the need for each local model to follow the same architecture to the global model.
	
	\textbf{Motivation.} In centralized learning, the network generalization technique has been well studied to overcome the overfitting problem \cite{lakshminarayanan2017simple, woodworth2020kernel}. Even in standard settings where the training and test data are drawn from a similar distribution, models still overfit the training data and the training model will fall into a sharp valley of the loss surface by using Empirical Risk Minimization (ERM) \cite{chaudhari2019entropy}. This effect is further intensified when the training and test data are of different distributions. Similarly, in FL, overfitting the local training data of each client is detrimental to the performance of the global model, as the distribution shift problem creates conflicting objectives among local models. The main strategy to improve the FL performance is to mitigate the local models to the global model from the average perspective \cite{karimireddy2020scaffold, yang2021achieving, li2018federated}, which has been demonstrated to accelerate the convergence of FL. However, fewer existing FL studies focus on how to protect the learning performance of the clients with poor performance, and hence parts of clients may lose their unique properties and incur large performance deviation. Therefore, a focus on improving global model generality should be of primary concern in the presence of the distribution shift problem. Improving local training generality would inherently position the objective of the clients closer to the global model objective. 
	
	Recently, efficient algorithms Sharpness Aware Minimization (SAM) \cite{foret2021sharpnessaware} have been developed to make the surface of loss function more smooth and generalized. It does not need to solve the min-max objectives as adversarial learning \cite{goodfellow2014explaining, shafahi2020universal}; instead, it leverages linear approximation to improve the efficiency. As we discussed previously, applying SAM to be the local optimizer for generalizing the global model in FL should be an effective approach. We first introduce a basic algorithm adopting SAM in FL settings, called \texttt{FedSAM}, where each local client trains the local model with the same perturbation bound.
	
	Although \texttt{FedSAM} can help to make the global model generalization and improve the training performance, they do not affect the global model directly. In order to bridge the smooth information on both local and global models without accessing others' private data, we develop our second and more important algorithm in our framework, termed Momentum \texttt{FedSAM} (\texttt{MoFedSAM}) by additionally downloading the global model updates of the previous round, and then letting clients perform local training on both local dataset and global model updates by SAM.

	\noindent\textbf{Contributions.} We summarize our contributions as follows: 
	(1) We approach one of the most troublesome cross-device FL challenges, i.e., distribution shift caused by data heterogeneity. To generalize the global model, we first propose a simple algorithm \texttt{FedSAM} performing SAM to be the local optimizer.
	
	(2) We prove the convergence results $\mathcal{O}(\frac{L}{\sqrt{RKN}})$ and $\mathcal{O}(\frac{\sqrt{K}}{\sqrt{RS}})$ for \texttt{FedSAM} algorithm, which matches the best convergence rate of existing FL studies. For the part of local training in the convergence rate, our proposed algorithms show speedup. Moreover, the generalization bound of \texttt{FedSAM} is also presented.
	
	(3) To directly smooth the global model, we develop \texttt{MoFedSAM} algorithm, which performs local training with both local dataset and global model updates by SAM optimizer. Then, we present the convergence rates are $\mathcal{O}(\frac{\sqrt{\beta L}}{\sqrt{RKN}})$ and $\mathcal{O}(\frac{\sqrt{\beta K}}{\sqrt{RS}})$ on full and partial client participation strategies, which achieves speedup and implies that \texttt{MoFedSAM} is a more efficient algorithm to address the distribution shift problem.

	\textbf{Related work.} In this paper, we aim to evaluate and distinguish the generalization performance of clients. Throughout this paper, we only focus on the classic cross-device FL setting \cite{mcmahan2017communication, li2018federated, karimireddy2020scaffold} in which a single global model is learned from and served to all clients. In the Personalized FL (PFL) setting \cite{t2020personalized, fallah2020personalized, singhal2021federated}, the goal is to learn and serve different models for different clients. While related, our focus and contribution are orthogonal to personalization. In fact, our proposed algorithms are easy to extend to the PFL setting. For example, by solving a hyperparameter to control the interpolation between local and global models \cite{deng2020adaptive, li2021ditto}, the participating clients can be defined as the clients that contribute to the training of the global model. We can use SAM to develop the global model and generate the local models by ERM to improve the performance.
	
	Momentum FL is an effective way to address the distribution shift problem and accelerate the convergence, which is based on injecting the global information into the local training directly. Momentum can be set on the server \cite{wang2019slowmo, reddi2020adaptive}, client \cite{karimireddy2021breaking, xu2021fedcm} or both \cite{khanduri2021stem}. As we introduce previously, while these algorithms accelerate the convergence, the global model will locate in a sharp valley and overfit. As such, the global model may not be efficient for all clients and generate a large deviation.

	We propose to train global models using a set of participating clients and examine their performance both on training and validation datasets. In the centralized learning, some studies\cite{keskar2016large, lakshminarayanan2017simple, woodworth2020kernel} consider the out-of-distribution generalization problem, which shows on centrally trained models that even small deviations in the morphology of deployment examples can lead to severe performance degradation. The sharpness minimization is an efficient way to deal with this problem \cite{foret2021sharpnessaware, kwon21b, zhuang2022surrogate, du2021efficient}. The FL setting differs from these other settings in that our problem assumes data is drawn from a distribution of client distributions even if the union of these distributions is stationary. Therefore, in FL settings, we consider the training performance and validation. It incurs more challenges than centralized learning. Although some studies develop algorithms to generalize the global model in FL \cite{mendieta2021local, yuan2021we, yoon2021fedmix}, they lack theoretical analysis of how the proposed algorithm can improve the generalization and may incur privacy issues. A recent study \cite{caldarola2022improving} shows via empirical experiments that using SAM to be the local optimizer can improve the generalization of FL.

	\section{Preliminaries and Proposed Algorithms}
	\subsection{\texttt{FedAvg} Algorithm}
	Consider a FL setting with a network including $N$ clients connected to one aggregator. We assume that for every $ i \in [N]$ the $i$-th client holds $m$ training data samples $\xi_i = (\mathbf{X}_{i}, Y)$ drawn from distribution $\mathcal{D}_i$. Note that $\mathcal{D}_i$ may differ across different clients, which corresponds to client heterogeneity. Let $F_i (w)$ be the training loss function of the client $i$, i.e., $F_i (w) \triangleq \mathbb{E}_{\xi_i \sim \mathcal{D}_i}[\mathcal{L}_i (w, \xi_i )]$, where $\mathcal{L}_i (w, \xi_i )$ is the per-data loss function. The classical FL problem \cite{mcmahan2017communication, Li2020On, karimireddy2020scaffold} is to fit the best model $w$ to all samples via solving the following empirical risk minimization (ERM) problem on each client:
	\begin{equation}\label{Eq:goal}
		\min_{w} \bigg\{F(w) := \frac{1}{N}\sum_{i\in[N]} F_i (w)\bigg\}.
	\end{equation}
	where $F(w)$ is the loss function of the global model. \texttt{FedAvg} \cite{mcmahan2017communication} is one of the most popular algorithms to address \eqref{Eq:goal}. In the communication round $r$, the server randomly samples $\mathcal{S}^r$ clients with the number of $S$ and downloads the global model $w^r$ to them. After receiving the global model, these sampled clients run $K$ times local Stochastic Gradient Descent (SGD) epochs using their local dataset in parallel, and upload the local model updates $w_{i,K}^r$ to the server. When the server receives all the local model updates, it averages these to obtain the new global model $w^{r+1}$. The pesudocode of \texttt{FedAvg} is shown in Algorithm~\ref{ALG:three}.

	\subsection{\texttt{FedSAM} Algorithm}
	Statistically heterogeneous local training dataset across the clients is one of the most important problems in FL studies. By capturing the Non-IID nature of local datasets in FL, the common assumption in existing FL studies \cite{mohri2019agnostic, Li2020Fair, karimireddy2020scaffold, reisizadeh2020robust} considers that the data samples of each client have a local distribution shift from a common unknown mixture distribution $\mathcal{D}$, i.e., $\mathcal{D}_i \neq \mathcal{D}$. While training via minimizing ERM by SGD searches for a single point $w$ with a low loss, which can perfectly fit the distribution $\mathcal{D}$, it often falls into a sharp valley of the loss surface \cite{chaudhari2019entropy}. As a result, the global model $w$ may be biased to parts of clients (i.e., low heterogeneity compared to the mixture distribution $\mathcal{D}$) and cannot guarantee enough generalization that makes all clients perform well. Moreover, since the training dataset distribution of each client may be different from the validation dataset with high probability, i.e., $\mathcal{D}_i^{\text{tra}} \neq \mathcal{D}_i^{\text{val}}$, and the validation dataset cannot be accessible during the training process, the global model $w$ may not guarantee the learning performance of every client even for the clients working well during the training process. To address this problem, some FL algorithms with fairness guarantee have been developed \cite{Li2020Fair, du2021fairness}, but they only consider the learning performance from the average perspective and do not protect the worse clients. In order to focus on the average and deviation for all clients at the same time, it is necessary to create a more general global model to serve all clients.

	Instead of searching for a single point solution such as ERM, the state-of-the-art algorithm Sharpness Aware Minimization (SAM) \cite{foret2021sharpnessaware} aims to seek a region with low loss values via adding a small perturbation to the models, i.e., $w + \delta$ with less performance degradation. Due to the linear property of the FL optimization in \eqref{Eq:goal}, it is not difficult to observe that training the perturbed loss via SAM, i.e., $\tw = w +\delta_i$, on each client should reduce the impact on the distribution shift and improve the generalization of the global model. Based on this observation, we design a more general FL algorithm called \texttt{FedSAM} in this paper. The optimization of \texttt{FedSAM} is formulated as follows:
	\begin{equation}\label{Eq:sam}
		\min_{w}\max_{\|\delta_i \|_2^2 \leq \rho}\bigg\{f(\tw) := \frac{1}{N}\sum_{i\in[N]} f_i (\tw)\bigg\},
	\end{equation}
	where $f(\tw) \triangleq \max_{\|\delta\|\leq \rho}F(w + \delta)$, $f_i (\tw) \triangleq \max_{\|\delta_i \| \leq \rho} F_i (w + \delta_i )$, $\rho$ is a predefined constant controlling the radius of the perturbation and $\|\cdot\|_2^2$ is a $l_2$-norm, which will be simplified to $\|\cdot\|$ in the rest paper. Next, we take a close look at the local perturbed loss function $F_i (w + \delta_i )$ and introduce how to use SAM to approach it. For a small value of $\rho$, using first order Taylor expansion around $w$, the inner maximization in \eqref{Eq:sam} turns into the following linear constrained optimization:
	\begin{equation}
		\begin{split}
			\delta_i & = \argmax_{\|\delta_i \| \leq \rho} F_i (w + \delta_i ) \\
			& \approx \argmax_{\|\delta_i \| \leq \rho}F_i (w) + \delta^{\top}_i \nabla F_i (w) + O(\rho^2 ) \\
			& = \rho \text{sign}(\nabla F_i (w)) \frac{\nabla F_i (w)}{\|\nabla F_i (w)\|},
		\end{split}
	\end{equation}
	where $\text{sign}(\cdot)$ denotes element-wise signum function.
	Therefore, the local optimizer of \texttt{FedSAM} changes to $\min_{w} F_i (w) = \min_{\tw} f_i (\tw)$, where $\tw \triangleq w + \rho\frac{\nabla F_i (w)}{\|\nabla F_i (w)\|}$. We call $\tw$ is the perturbed model with the highest loss within the neighborhood. Local SAM optimizer solves the min-max problem by iteratively applying the following two-step procedure for epoch $k = 0,\dots, K-1$ in communication round $r$: 
	\begin{equation}
		\left\{
		\begin{aligned}\label{Eq:SAMpro}
			\tw_{i,k}^r & = w_{i,k}^r + \rho\frac{g_{i,k}^r}{\|g_{i,k}^r\|} \\
			w_{i,k+1}^r & = w_{i,k}^r - \eta_l \tg_{i,k}^r ,
		\end{aligned}
		\right.
	\end{equation}
	where $\eta_l$ is the learning rate of local model updates on each client, $g_{i,k}^r = \nabla F_i (w_{i,k}^r , \xi_{i}^r )$ of $\nabla F_i (w_{i,k})$ and $\tg_{i,k}^r = \nabla f(\tw_{i,k}^r , \xi_{i}^r )$ of $f_i (\tw_{i,k}^r )$. We can see that from \eqref{Eq:SAMpro}, local training of each client estimates the point $w_{i,k}^r + \delta_i^r$ at which the local loss is maximized around $w_{i,k}^r$ in a region with a fixed perturbed radius approximately by using gradient ascent, and calculates gradient descent at $w_{i,k}^r$ based on the gradient at the maximum point $w_{i,k}^r + \delta_i^r$. 
	
	\begin{algorithm}[t!]
		\caption{\colorbox[rgb]{1.0, 0.55, 0.41}{FedAvg} and \colorbox[rgb]{0.74,0.83,1}{FedSAM}}
		\begin{algorithmic}
			\STATE Initialization: $w_0$, $\rho_0$ $\Delta^0 = 0$, learning rates $\eta_l$, $\eta_g$ and the number of epochs $K$.
			\FOR {$r = 0, \dots, R-1$}
			\STATE Sample subset $\S^r \subseteq [N]$ of clients.
			\STATE $w^t_{i,0} = w^r$.
			\FOR {each client $i \in \S^r$ in parallel}
			\FOR {$k=0,\dots, K-1$}
			\STATE Compute a local training estimate $g_{i,k}^r = \nabla F_{i}(w_{i,k}^r , \xi_{i,k}^r )$ of $\nabla F_i (w_{i,k}^r )$.
			\STATE \colorbox[rgb]{1.0, 0.55, 0.41}{$w_{i,k}^r = w_{i,k}^r - \eta_l g_{i,k}^r$.} 
			\STATE \colorbox[rgb]{0.74,0.83,1}{Compute local model $w_{i,k}^r$ from \eqref{Eq:SAMpro}.}
			\ENDFOR
			\STATE $\Delta_i^r = w_{i,K}^r - w^r$.
			\ENDFOR
			\STATE $\Delta^{r+1} = \frac{1}{S}\sum_{i\in\mathcal{S}^r}\Delta_i^r$.
			\STATE $w^{r+1} = w^r + \eta_g \Delta^r$.
			\ENDFOR
		\end{algorithmic}
		\label{ALG:three}
	\end{algorithm}
	
	To present the difference between \texttt{FedAvg} and \texttt{FedSAM}, we summarize the training procedures in Algorithm~\ref{ALG:three}. SAM optimizer comes from the similar idea of adversarial training, and it has been used in FL \cite{reisizadeh2020robust} called \texttt{FedRobust}. It is based on solving min-max objectives, which brings up more computational cost for local training and the worse convergence performance than our proposed algorithms. We will show the comparison both on theoretical and empirical perspectives.

	\begin{remark}
		\normalfont Here, we briefly mention the SAM local optimizer can improve the generalization and help convergence from the smoothness perspective. We assume that the local loss function $f(w)$ is $L$-smooth. Clearly, the loss function $f$ is smoother, when $L$ is smaller. Assume that $f(\tw)$ is $G$-Lipschitz continuous, and $\delta \sim \mathcal{N}(0,\epsilon^2 I)$, by leveraging \cite{nesterov2017random}, we obtain that the perturbed loss function $f (\tw)$ of \texttt{FedSAM} is $\frac{2G}{\epsilon}$-smooth. Based on the analysis in \cite{lian2017can, goyal2017accurate}, the best convergence rate should be $\frac{1}{L}$. For SGD based FL with the original loss surface, $L$ can be very high (even close to $+\infty$ due to the non-smooth nature of the ReLU activation). Obviously, $L$ of the perturbed loss $f(\tw)$ in \texttt{FedSAM} should be much smaller due to the loss region. This can explain the intuition why increasing smoothness can significantly improve the convergence of FL.
	\end{remark}

	\section{Theoretical Analysis}
	In what follows, we show the convergence results of \texttt{FedSAM} algorithm for general non-convex FL settings. In order to propose the convergence analysis, we first state our assumptions as follows.
	\begin{assumption}\label{ass:smooth}
		(Smoothness). $f_i$ is $L$-smooth for all $i \in [N]$, i.e., 
		\begin{equation}
			\|\nabla f_i (w) - \nabla f_i (v) \| \leq L \|w-v \|,\nonumber
		\end{equation}
		for all $w, v$ in its domain and $i \in [N]$.
	\end{assumption}
	
	\begin{assumption}\label{ass:sigmag}
		(Bounded variance of global gradient without perturbation). The global variability of the local gradient of the loss function without perturbation $\delta_i$ is bounded by $\sigma_g^2$, i.e., 
		\begin{equation}
			\|\nabla F_i (w^r ) - \nabla F(w^r )\|^2 \leq \sigma_g^2, \nonumber
		\end{equation}
		for all $i \in [N]$ and $r$.
	\end{assumption}

	\begin{assumption}\label{ass:sigmal}
		(Bounded variance of stochastic gradient). The stochastic gradient $\nabla f_i (w, \xi_i )$, computed by the $i$-th client of model parameter $w$ using mini-batch $\xi_i$ is an unbiased estimator $\nabla F_i (w)$ with variance bounded by $\sigma^2$, i.e., \begin{equation}
			\mathbb{E}_{\xi_i}\left\|\frac{\nabla F_i (w,\xi_i )}{\|\nabla F_i (w,\xi_i )\|} - \frac{\nabla F_i (w)}{\|\nabla F_i (w)\|}\right\|^2 \leq \sigma_l^2,\nonumber
		\end{equation}
		$\forall i \in [N]$, where the expectation is over all local datasets.
	\end{assumption}

	Assumptions~\ref{ass:smooth} and \ref{ass:sigmag} are standard in general non-convex FL studies \cite{Li2020On, karimireddy2020scaffold, reddi2020adaptive, karimireddy2021breaking, yang2021achieving} in order to assume the loss function continuous and bound the heterogeneity of FL systems. Note that we consider that $\sigma_g^2$ mainly depends on the data-heterogeneity, and the perturbation should be calculated. Hence, we only bound it without perturbation. We will present the upper bound of $\|\nabla f_i (\tw^r ) - \nabla f(\tw^r )\|^2$ in Appendix~\ref{Sec:Lemmas}. Assumption~\ref{ass:sigmal} bounds the variance of stochastic gradient. Although many FL studies use the similar assumption to bound the stochastic gradient variance \cite{Li2020On, karimireddy2020scaffold}, the definition is $\mathbb{E}_{\xi_i}\|\nabla F_i (w,\xi_i ) - \nabla F_i (w) \|^2 \leq \sigma_l^2$, which is not easy to measure the value of $\sigma_l^2$, and the upper bound of $\sigma_l^2$ may be closed to $+\infty$. In this paper, Assumption~\ref{ass:sigmal} is considered as the norm of difference in unit vectors that can be upper bounded by the arc length on a unit circle. Therefore, $\sigma^2$ should be less than $\pi^2$. Clearly, this assumption is tighter than existing FL studies.
	
	\subsection{Convergence Analysis of \texttt{FedSAM}}
	We now state our convergence results for \texttt{FedSAM} algorithm. The detailed proof is in Appendix~B.
	\begin{theorem}\label{The:fedsamconvergence}
		Let the learning rates be chosen as $\eta_l = \mathcal{O}(\frac{1}{\sqrt{R}KL})$, $\eta_g = \sqrt{KN}$ and the perturbation amplitude $\rho$ proportional to the learning rate, e.g., $\rho = \mathcal{O}(\frac{1}{\sqrt{R}})$. Under Assumptions~\ref{ass:smooth}-\ref{ass:sigmal} and full client participation, the sequence of iterates generated by \texttt{FedSAM} in Algorithm~\ref{ALG:three} satisfies:
		\begin{equation}
			\mathcal{O}\bigg(\frac{LF}{\sqrt{RKN}} + \frac{\sigma_g^2}{R} + \frac{L^2 \sigma_l^2}{R^{3/2}\sqrt{KN}} + \frac{L^2}{R^{2}} \bigg),\nonumber
		\end{equation}
		where $F = f(\tw^0 ) - f(\tw^* )$ and $f(\tw^* ) = \min_{\tw} f(\tw)$.
		
		For the partial client participation strategy and $S\geq K$, if we choose the learning rates $\eta_g = \mathcal{O}(\frac{1}{\sqrt{R}KL})$, $\eta_g = \sqrt{KS}$ and $\rho = \mathcal{O}(\frac{1}{\sqrt{R}})$, the sequence of iterates generated by \texttt{FedSAM} in Algorithm~\ref{ALG:three} satisfies:
		\begin{equation}
			\mathcal{O}\bigg(\frac{LF}{\sqrt{RKS}} + \frac{\sqrt{K}G^2}{\sqrt{RS}} + \frac{L^2 \sigma^2}{R^{3/2}K} + \frac{L^2}{R^2}\bigg).\nonumber
		\end{equation}
	\end{theorem}

	\begin{remark}
		\normalfont For the full and partial client participation strategies of \texttt{FedSAM} algorithm in this theorem, the dominant terms of the convergence rate are $\mathcal{O}(\frac{L}{\sqrt{RKN}})$ and $\mathcal{O}(\frac{\sqrt{K}}{\sqrt{SR}})$ by properly choosing the learning rates $\eta_l$ and $\eta_g$, which match the best convergence rate in existing general non-convex FL studies \cite{karimireddy2020scaffold, yang2021achieving, acar2021federated}. Since the convergence rate structures in this theorem of these two strategies are similar, it indicates that uniformly sampling does not result in fundamental changes of convergence. In addition, both convergence rates include four main terms with an additional term compared to \cite{karimireddy2020scaffold, yang2021achieving, acar2021federated}. Note that we only show the dominant part of each term in the main paper. The detailed proof can be found in Appendix.
	\end{remark}
	\begin{remark}
		\normalfont The additional term $\mathcal{O}(\frac{L^2}{R^2})$ comes from the additional SGD step for smoothness via SAM local optimizer in \eqref{Eq:SAMpro}. However, this term can be negligible due to its higher order. More specifically, since the smoothness is due to the local training, we can combine it with the local training term, i.e., $\mathcal{O}(\frac{\sigma^2}{R^{3/2}K} + \frac{1}{R^2})$. Clearly, this term also achieves speedup than the existing best rate, i.e., $\mathcal{O}(\frac{1}{R})$. For the partial participation strategy, the dominant term is due to fewer clients participating and random sampling (heterogeneity), i.e., $\mathcal{O}(\frac{\sqrt{K}G^2}{RS})$. The convergence rate improves substantially as the number of clients increases, which matches the results of partial client participation FL \cite{karimireddy2020scaffold, yang2021achieving, acar2021federated}. Intuitively, increasing the convergence rate of this term is because SAM optimizer can make the global model more generalization and reduce the distribution shift. 
	\end{remark}
	
	\begin{remark}
		\normalfont The \texttt{FedRobust} \cite{reisizadeh2020robust} algorithm is an adversarial learning framework in FL setting, which is based on the similar idea of \texttt{FedSAM}. It has the convergence rate of $\mathcal{O}(\frac{Lf}{(RN)^{1/3}} + \frac{L^2}{R^{1/3}N^{2/3}})$, and it does not perform well from the convergence perspective compared with \texttt{FedSAM}. Since multiple gradient descents steps should be computed in each local training epoch, \texttt{FedRobust} will waste more running time and computational cost to process the local training.
	\end{remark}

	\subsection{Generalization Bounds of \texttt{FedSAM}}
	Based on the margin-based generalization bounds in \cite{neyshabur2018pac, bartlett2017spectrally, farnia2018generalizable, reisizadeh2020robust}, we propose the generalization error of \texttt{FedSAM} algorithm with the general neural network as follows:
	\begin{equation}\label{Eq:marginmain}
		\begin{split}
			\mathcal{L}_\gamma^{\text{SAM}} (F(w)) := \frac{1}{N}&\sum_{i=1}^{N}\mathbb{P}_i \bigg(F_i (w+\delta_i , \mathbf{X}) [Y]  \\
			&  - \max_{j \neq Y}F_i (w+\delta_i ,\mathbf{X})[j] \leq \gamma \bigg).
		\end{split}
	\end{equation}
	Here, $F_i (w+\delta_i , \mathbf{X})$ is the loss function solving by SAM local optimizer for client $i$ in \eqref{Eq:sam}, $\mathbf{X}$ is an input, $\mathbb{P}_i$ is the probability of the underlying distribution of client $i$, and $F_i (w+\delta_i ,\mathbf{X})[j]$ is the output of the last softmax layer for label $j$ about the training neural network. It is worth noting that $\gamma$ is a constant, and for $\gamma = 0$, \eqref{Eq:marginmain} can be simplified to the average misclassification rate with the distribution shift, which is denoted by $\mathcal{L}^{\text{SAM}}$. In addition, we use $\hat{\mathcal{L}}_\gamma^{\text{SAM}}(w)$ as the above margin risk to represent the empirical distribution of training samples, and hence we use $\hat{\mathbb{P}}_i$ to replace the underlying $\mathbb{P}_i$ to be the empirical probability, which is calculated by the $m$ training samples on client $i$. 
	
	The following theorem aims to bound the difference of the empirical and the margin-based error defined in \eqref{Eq:marginmain} under a general deep neural network. We use the spectral norm based generalization bound framework \cite{neyshabur2018pac, farnia2018generalizable, chatterji2019intriguing} to prove the next theorem. In order to demonstrate the margin-based error bounds, we assume that the neural network with smooth ReLU activation functions $\theta$ are $1$-Lipschitz activation functions. The detailed proof is shown in Appendix~\ref{Sec:generalizationbound}.
	
	\begin{theorem}\label{The:generalization}
		Let input $\mathbf{X}$ be an $n\times n$ image whose norm is bounded by $A$, $f(w)$ be the classification function with $d$ hidden-layer neural network with $h$ units per hidden-layer, and satisfy $1$-Lipschitz activation $\theta (0) =0$. We assume the constant $M\geq 1$ for each layer $W_j$ satisfies $\frac{1}{M} \leq \frac{\|W_j \|}{\phi_{w}} \leq M$, where $\phi_{w} :=(\prod_{j=1}^{d}\|W_j \| )^{1/d}$ denotes the geometric mean of $f(w)$'s spectral norms across all layers. Then, for any margin value $\gamma$, size of local training dataset on each client $m$, $\zeta >0$, with probability $1-\zeta$ over the training set, any parameter of SAM local optimizer $\tw = w + \delta$ such that $\max_{\mathbf{X} \in \mathcal{D}_i}\|F_i (w) - f(\tw)\| \leq \frac{\gamma}{8}$, we can obtain the following generalization bound:
		\begin{equation}
			\begin{split}
				& \mathcal{L}^{\text{SAM}} (F(w)) \leq \hat{\mathcal{L}}^{\text{SAM}}_{\gamma}(F(w + \delta))\\
				& + \mathcal{O}\bigg(\frac{32Ad^2 h \log(dh)Q(F(w)) + d\log\frac{Nmd\log (M)}{\zeta}}{\gamma^2 m}\bigg),\nonumber
			\end{split}
		\end{equation}
		where $Q(F(w)) := \prod_{j=1}^{d}\|W_j \|\sum_{i=1}^{d}\frac{\|W_j \|_F^2}{\|W_j \|}$ and $\|W_j \|_F^2$ is the Frobenius norm.
	\end{theorem}
	
	Theorem~\ref{The:generalization} proposes a non-asymptotic bound on the generalization risk of \texttt{FedSAM} for general neural networks. The PAC-Bayesian bounds of SAM \cite{foret2021sharpnessaware, kwon21b, zhuang2022surrogate, du2021efficient} does not provide the insight about the underlying reason that results in generalization, i.e., how to choose the value of $\lambda$ in the Gaussian noise $\mathcal{N}(0, \lambda I)$ to be the perturbation. In Theorem~\ref{The:generalization}, we present the dependence of the perturbation $\delta$ and the different neural network parameters in which we can enforce the loss surface around a point in order to guarantee the smoothness.

	\begin{algorithm}[t!]
		\caption{\texttt{MoFedSAM} algorithm.}
		\begin{algorithmic}[1]
			\STATE Initialization: $w^0$, $\Delta^0 = 0$, $\rho^0$, momentum parameter $\beta$ the number of local updates $K$.
			\FOR {$r = 0, \dots, R-1$}
			\STATE Sample subset $\S^r \subseteq [N]$ of clients.
			\STATE $w^t_{i,0} = w^r$.
			\FOR {each client $i \in \S^r$ in parallel}
			\FOR {$k=0,\dots, K-1$}
			\STATE Compute a local training estimate $g_{i,k}^r = \nabla F_{i}(w_{i,k}^r , \xi_{i,k}^r )$ of $\nabla F_i (w_{i,k}^r )$.
			\STATE Compute local model $w_{i,k}^r$ from \eqref{Eq:MofedSAMpro}.
			\ENDFOR
			\STATE $\Delta_i^r = w_{i,K}^r - w^r$.
			\ENDFOR
			\STATE $\Delta^{r+1} = -\frac{1}{\eta_l K S}\sum_{i \in \mathcal{S}^r}\Delta_i^r$.
			\STATE $w^{r+1} = w^r - \eta_g \Delta^{r+1}$.
			\ENDFOR
		\end{algorithmic}
		\label{ALG:MoFedSAM}
	\end{algorithm}

	\section{Momentum \texttt{FedSAM} (\texttt{MoFedSAM})}
	\subsection{Algorithm of \texttt{MoFedSAM}}
	Since $\Delta^r$ serves as the direction for the global model, while \texttt{FedSAM} algorithm achieves efficient convergence rate theoretically, the influence of local optimizer cannot directly affect the global model, i.e., the term including $\sigma_g^2$ in the convergence rate. Note that $\Delta^r$ aggregates the global model information of participating clients, and reusing this information should be useful to guide the local training on the participated clients in next communication round, which is similar to momentum FL \cite{wang2019slowmo, reddi2020adaptive, karimireddy2021breaking, khanduri2021stem, xu2021fedcm}. Inspired by this motivation, we now provide our second algorithm, termed \texttt{MoFedSAM}, which aims to smooth and generalize the global model directly. The training procedure of $k$-th local training epoch in round $r$ is formulated as follows:
	\begin{equation}
		\left\{
		\begin{aligned}\label{Eq:MofedSAMpro}
			\tw_{i,k}^r & = w_{i,k}^r + \rho\frac{g_{i,k}^r}{\|g_{i,k}^r\|} \\
			v_{i,k}^r & = \beta \tg_{i,k}^r + (1-\beta)\Delta^r \\
			w_{i,k}^r & = w_{i,k}^r - \eta_l v_{i,k}^r ,
		\end{aligned}
		\right.
	\end{equation}
	where $\beta$ is the momentum rate. If $\beta = 1$, \texttt{MoFedSAM} is equivalent to \texttt{FedSAM}. From \eqref{Eq:MofedSAMpro}, we can see that the global model information $\Delta^r$ directly contributes the local training epoch, since $w_{i,k}^r$ includes $\tg_{i,k}^r$ and $\Delta^r$ at the same time. Therefore, it indicates that \texttt{MoFedSAM} make the local and global models smoothness at the same time. Especially, even if only a subset of clients are sampled in each communication round, the information of gradients of previous local model updates can be still contained in $\Delta^r$. Therefore, \texttt{MoFedSAM} also works well of partial client participation FL. More specifically, the global model information term $\Delta^r$ is considered as an approximation to the gradient of the global model $\nabla f(\tw)$, i.e., $\Delta^r \approx \nabla f(\tw^r )$. One advantage is that \texttt{MoFedSAM} adds a correction term to the local gradient direction, and it also asymptotically aligns with the difference between global and local gradient. It is worth noting that we use $(G,B)$-BGD in Assumption~\ref{ass:sigmag} to prove the convergence rate, which is tighter than $(G,0)$-BGD in \cite{xu2021fedcm}.

	\begin{figure*}[t!]
		\centering
		\begin{minipage}{0.64\columnwidth}
			\centering
			\includegraphics[width=\textwidth]{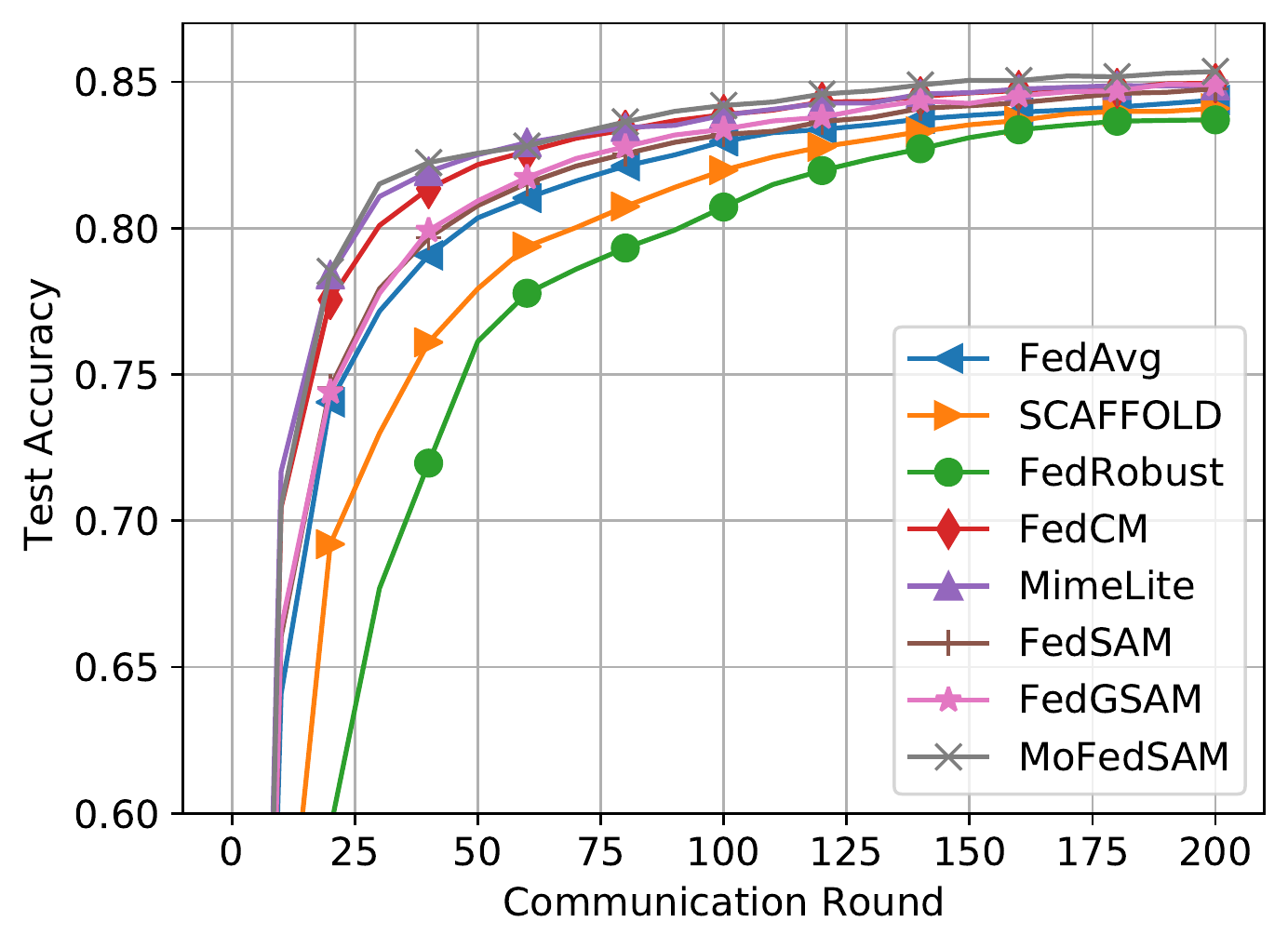}
			\subcaption[first]{EMNIST dataset.}
		\end{minipage}
		\hfill
		\begin{minipage}{0.64\columnwidth}
			\centering
			\includegraphics[width=\textwidth]{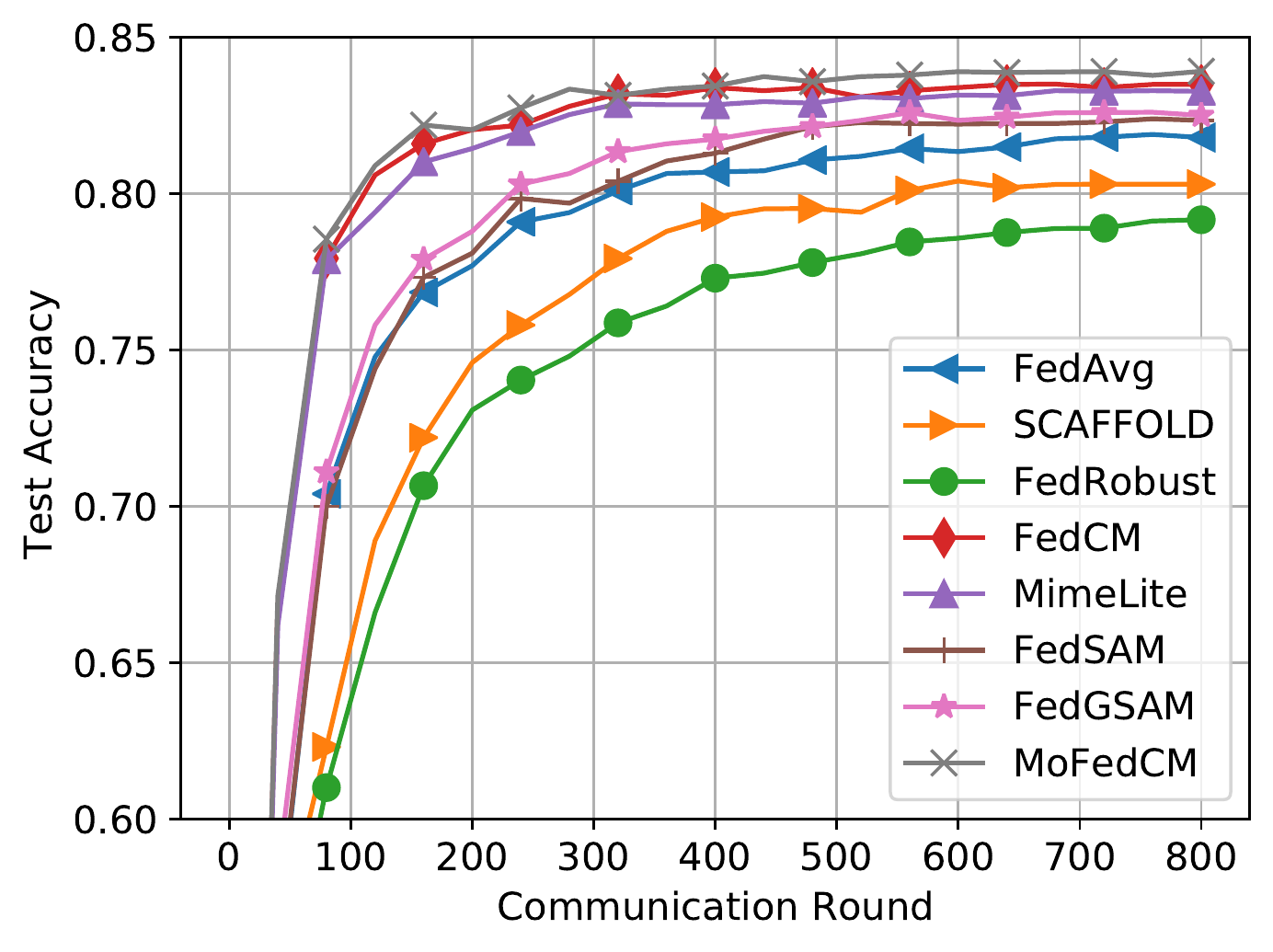}
			\subcaption[second]{CIFAR-10 dataset.}
		\end{minipage}%
		\hfill
		\begin{minipage}{0.64\columnwidth}
			\centering
			\includegraphics[width=\textwidth]{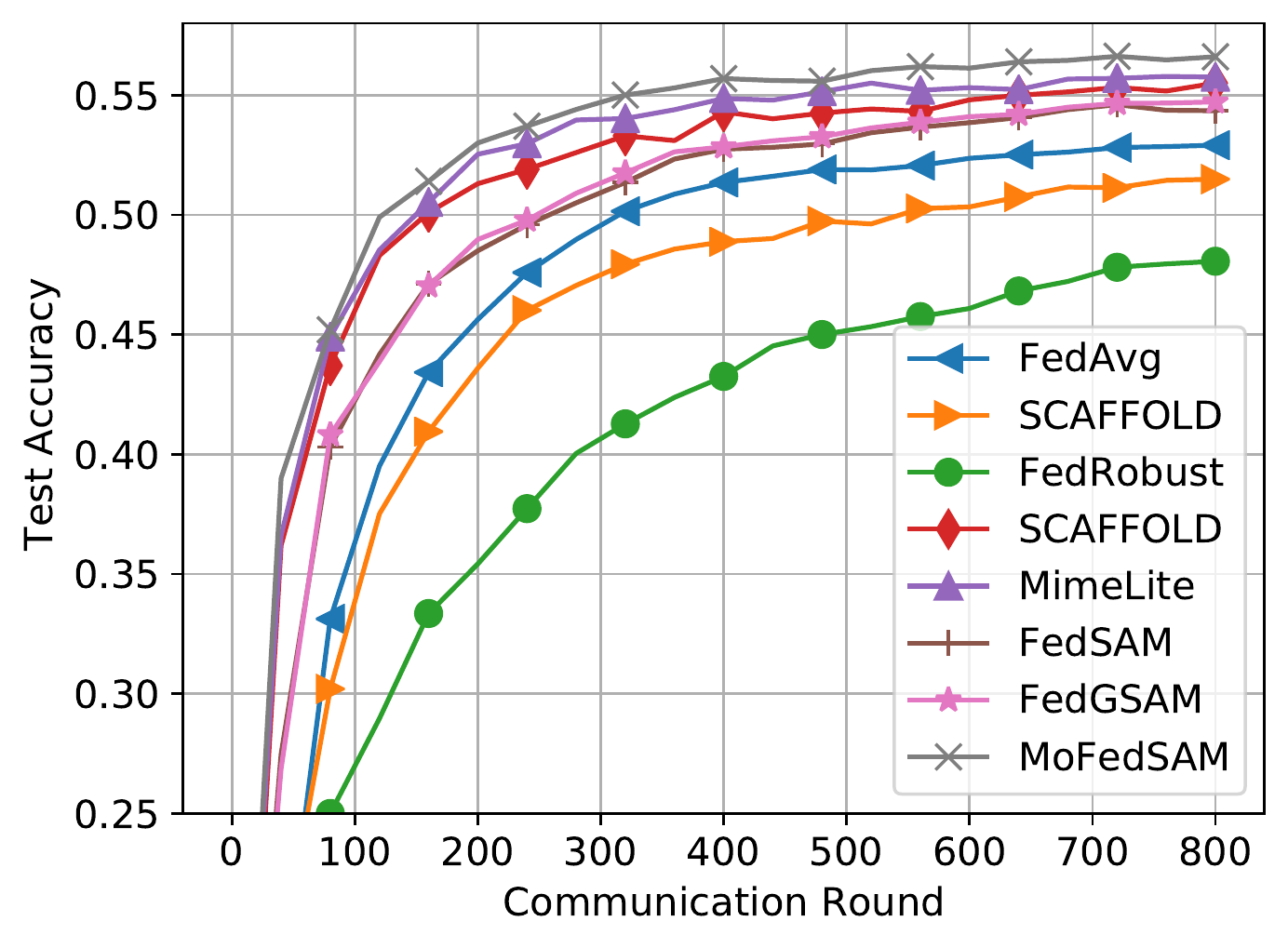}
			\subcaption[third]{CIFAR-100 dataset.}
		\end{minipage}
		\caption{Testing accuracy on different datasets.}
		\label{Fig:dataset}
	\end{figure*}
	
	\begin{table*}[t!]
		\centering
		\caption{Average (standard deviation) training accuracy and testing accuracy. Communication round to achieve the targeted testing accuracy: EMNIST 80\%, CIFAR-10 80\% and CIFAR-100 50\%.}
		\begin{adjustbox}{max width=0.97\textwidth}
			\begin{tabular}{*{10}{c}}
				\toprule 
				\multirow{2}{*}{Algorithm}          & \multicolumn{3}{c}{EMNIST}              & \multicolumn{3}{c}{CIFAR-10}            & \multicolumn{3}{c}{CIFAR-100}           \\\cmidrule{2-10}
				& Train          & Validation    & Round & Train          & Validation     & Round & Train & Validation     & Round \\ \midrule
				\texttt{FedAvg}   & 95.07 (0.94) & 84.38 (4.03) & 43 & 93.15 (1.44) & 81.87 (5.09) & 307 & 79.57 (1.84) & 53.57 (5.40) & 302 \\
				\texttt{SCAFFOLD} & 93.85 (1.31) & 84.09 (4.56) & 69 & 91.76 (1.89) & 80.61 (5.64) & 546 & 78.49 (2.02) & 51.49 (5.87) & 551 \\
				\texttt{FedRobust} & 93.17 (0.62) & 83.70 (3.37) & 91 & 90.82 (1.27) & 79.63 (4.21) & 847 & 76.80 (1.70) & 49.06 (4.75) & 893 \\
				\texttt{FedCM}   & 96.16 (1.14) & 84.85 (4.11) & 28 & 95.61 (1.50) & 83.30 (4.77) & 136 & 82.13 (1.96) & 55.50 (5.04) & 182 \\
				\texttt{MimeLite}  & 96.22 (1.16) & 84.88 (4.22) & 25 & 95.73 (1.56) & 83.18 (4.65) & 152 & 82.46 (2.00) & 55.73 (5.11)& 189 \\
				\texttt{FedSAM}   & 95.73 (0.49) & 84.75 (3.04) & 38 & 94.20 (1.08) & 83.06 (3.87) & 269 & 81.04 (1.59) & 54.69 (4.36) & 245 \\
				\texttt{MoFedSAM}  & 96.42 (0.42) & 85.07 (2.95) & 24 & 95.67 (1.16) & 83.92 (3.65) & 124 & 82.62 (1.53) & 56.60 (4.42) & 124 \\
				\toprule
			\end{tabular}
		\end{adjustbox}
		\label{Tab:compare}
	\end{table*}
	
	\subsection{Convergence Analysis of \texttt{MoFedSAM}}
	Next theorem is the convergence rate of \texttt{MoFedSAM} algorithm, and the detailed proof is in Appendix~D.
	\begin{theorem}\label{The:fedmosamconvergence}
		Let the learning rates be chosen as $\eta_l = O(\frac{1}{\sqrt{R \beta K}L})$, $\eta_g = O(\frac{\sqrt{KN}}{\sqrt{R}\beta L})$ and the perturbation amplitude $\rho$ proportional to the learning rate, e.g., $\rho = \mathcal{O}(\frac{1}{\sqrt{R}})$. Under the Assumptions~\ref{ass:smooth}-\ref{ass:sigmal}, any momentum parameter $\beta \leq \frac{1}{2}$ and the full client participation strategy, the sequence of $\{\tw^r \}$ generated by \texttt{MoFedSAM} in Algorithm~\ref{ALG:MoFedSAM} satisfies:
		\begin{equation}
			\mathcal{O}\bigg(\frac{\beta LF}{\sqrt{RKN}} + \frac{\beta \sigma_g^2}{R L^2} + \frac{L \sigma^2}{R^2 \beta} + \frac{\beta L^2}{R^2} \bigg),\nonumber
		\end{equation}
		where $F = f(\tw^0 ) - f(\tw^* )$ and $f(\tw^* ) = \min_{\tw} f(\tw)$.
		
		For the partial client participation strategy, if we choose the learning rates $\eta_g = \mathcal{O}(\frac{1}{\sqrt{R \beta K}L})$, $\eta_g = \mathcal{O}(\frac{\sqrt{KS}}{\sqrt{R}\beta L})$ and $\rho = \mathcal{O}(\frac{1}{\sqrt{R}})$, the following convergence holds:
		\begin{equation}
			\mathcal{O}\bigg(\frac{\beta LF}{\sqrt{RKS}} + \frac{\beta \sqrt{K}\sigma_g^2}{\sqrt{RS}} + \frac{L^2 \sigma^2}{R^{3/2}K} + \frac{\sqrt{K}L^2}{R^{3/2} \sqrt{S}}\bigg).\nonumber
		\end{equation}
	\end{theorem}
	
	\begin{remark}
		\normalfont When $T$ is sufficiently large compared to $K$, convergence rates under full and partial client participation strategies of \texttt{MoFedSAM} algorithm are $\mathcal{O}(\frac{\sqrt{\beta L}}{\sqrt{RKN}} + \frac{\beta}{R L^2})$ and $\mathcal{O}(\frac{\beta L}{\sqrt{RKS}} + \frac{\beta \sqrt{K}}{\sqrt{RS}})$. The momentum parameter $\beta$ is small enough, i.e., 0.1 \cite{karimireddy2021breaking, xu2021fedcm}, from which the effect is important for convergence, due to the number of local epochs setting less than $20$ in usual \cite{reddi2020adaptive, yang2021achieving, acar2021federated}. Therefore, our convergence results achieve speedup compared with \texttt{FedSAM}. We also note that the convergence related to the local training is $\mathcal{O}(\frac{L}{R^2 \beta} + \frac{\beta L^2}{R^2})$ and $\mathcal{O}(\frac{L^2}{R^{3/2}K} + \frac{\sqrt{K}L^2}{R^{3/2} \sqrt{S}})$, where the second part comes from sharpness, and it can be negligible. From the convergence analysis of \texttt{FedCM} \cite{xu2021fedcm}, i.e., $\mathcal{O}(\frac{\sqrt{KSL}}{\sqrt{R}} + \frac{L}{\beta^{2/3}R^{2/3}})$, we can see that \texttt{MoFedSAM} achieves speedup both on the dominant part and local training part. The analysis indicates the benefit of bridging the sharpness between local and global models.
	\end{remark}

	\begin{table*}[t!]
		\centering
		\caption{Impact of the heterogeneity on CIFAR-10 dataset (IID, Dirichlet 0.6 and Dirichlet 0.3).}
		\begin{adjustbox}{max width=0.97\textwidth}
			\begin{tabular}{*{10}{c}}
				\toprule 
				\multirow{2}{*}{Algorithm}       & \multicolumn{3}{c}{IID}              & \multicolumn{3}{c}{Dirichlet 0.6}            & \multicolumn{3}{c}{Dirichlet 0.3}           \\\cmidrule{2-10}
				& Train          & Validation     & Round & Train          & Validation     & Round & Train          & Validation     & Round \\ \midrule
				\texttt{FedAvg} & 94.95 (1.01) & 85.97 (3.53) & 238 & 93.15 (1.44) & 81.87 (5.09) & 307 & 91.89 (1.63) & 77.39 (5.62) & -   \\
				\texttt{SCAFFOLD}  & 93.04 (1.13) & 83.82 (3.72) & 290 & 91.76 (1.89) & 80.61 (5.64) & 546 & 90.02 (2.08) & 75.67 (5.93) & - \\
				\texttt{FedRobust} & 91.63 (0.91) & 82.44 (3.15) & 361 & 90.82 (1.27) & 79.63 (4.21) & 847 & 89.72 (1.42) & 73.11 (5.11) & - \\
				\texttt{FedCM}     & 97.02 (1.10) & 88.14 (3.33) & 87 & 95.61 (1.50) & 83.30 (4.77) & 136 & 93.88 (1.67) & 81.34 (5.50) & 583 \\
				\texttt{MimeLite}  & 97.16 (1.08) & 88.53 (3.53) & 82 & 95.73 (1.56) & 83.18 (4.65) & 152 & 93.97 (1.72) & 81.83 (5.53) & 548  \\
				\texttt{FedSAM}    & 95.42 (0.81) & 87.36 (2.85) & 205 & 94.20 (1.08) & 83.06 (3.87) & 269 & 92.90 (1.26) & 79.82 (4.98) & 816 \\
				\texttt{MoFedSAM}  & 97.22 (0.88) & 88.96 (2.94) & 75 & 95.67 (1.16) & 83.92 (3.65) & 124 & 94.12 (1.31) & 83.35 (5.06) & 490 \\\toprule
			\end{tabular}
		\end{adjustbox}
		\label{Tab:heterogeneous}
	\end{table*}

	\section{Experiments}\label{Sec:experiment}
	We evaluate our proposed algorithms on extensive and representative datasets and learning models to date. To accomplish this, we conduct experiments on three learning models across three datasets comparing to five FL benchmarks with varying different parameters.
	\subsection{Experimental Setup}
	\textbf{Benchmarks and hyper-parameters.} We consider five FL benchmarks: without momentum FL \texttt{FedAvg} \cite{mcmahan2017communication}, \texttt{SCAFFOLD} \cite{karimireddy2020scaffold}, \texttt{FedRobust} \cite{reisizadeh2020robust}; momentum FL \texttt{MimeLite} \cite{karimireddy2021breaking} and FedCM \cite{xu2021fedcm}. The learning rates are individually tuned and other optimizer hyper-parameters such as $\rho = 0.5$ for SAM and $\beta =0.1$ for momentum, unless explicitly stated otherwise. We refer to Appendices~\ref{Sec:Setup}-\ref{Sec:add} for detailed experimental setup and additional ablation studies.
	
	\textbf{Datasets and models.} We use three images datasets: EMNIST \cite{cohen2017emnist}, CIFAR-10, and CIFAR-100 \cite{krizhevsky2009learning}. Our cross-device FL setting includes 100 clients in total with participation rate $20$\%. In each communication round, each client is sampled independently of each other, with probability 0.2. We simulate the data heterogeneity by sampling the label ratios from a Dirchlet distribution with parameter 0.6 \cite{acar2021federated}, the number of local epochs is set as $K =10$ by default. We adopt two learning models on each dataset: (i) CNN on EMNIST with batch $32$ and (ii) ResNet-18 \cite{he2016deep} on CIFAR-10 and CIFAR-100 with batch $128$. The detailed experimental setup and other additional experiments and ablation studies will be shown in Appendices~\ref{Sec:Setup}-\ref{Sec:add}.
	
	\begin{figure*}[t!]
		\centering
		\begin{minipage}{0.65\columnwidth}
			\centering
			\includegraphics[width=\textwidth]{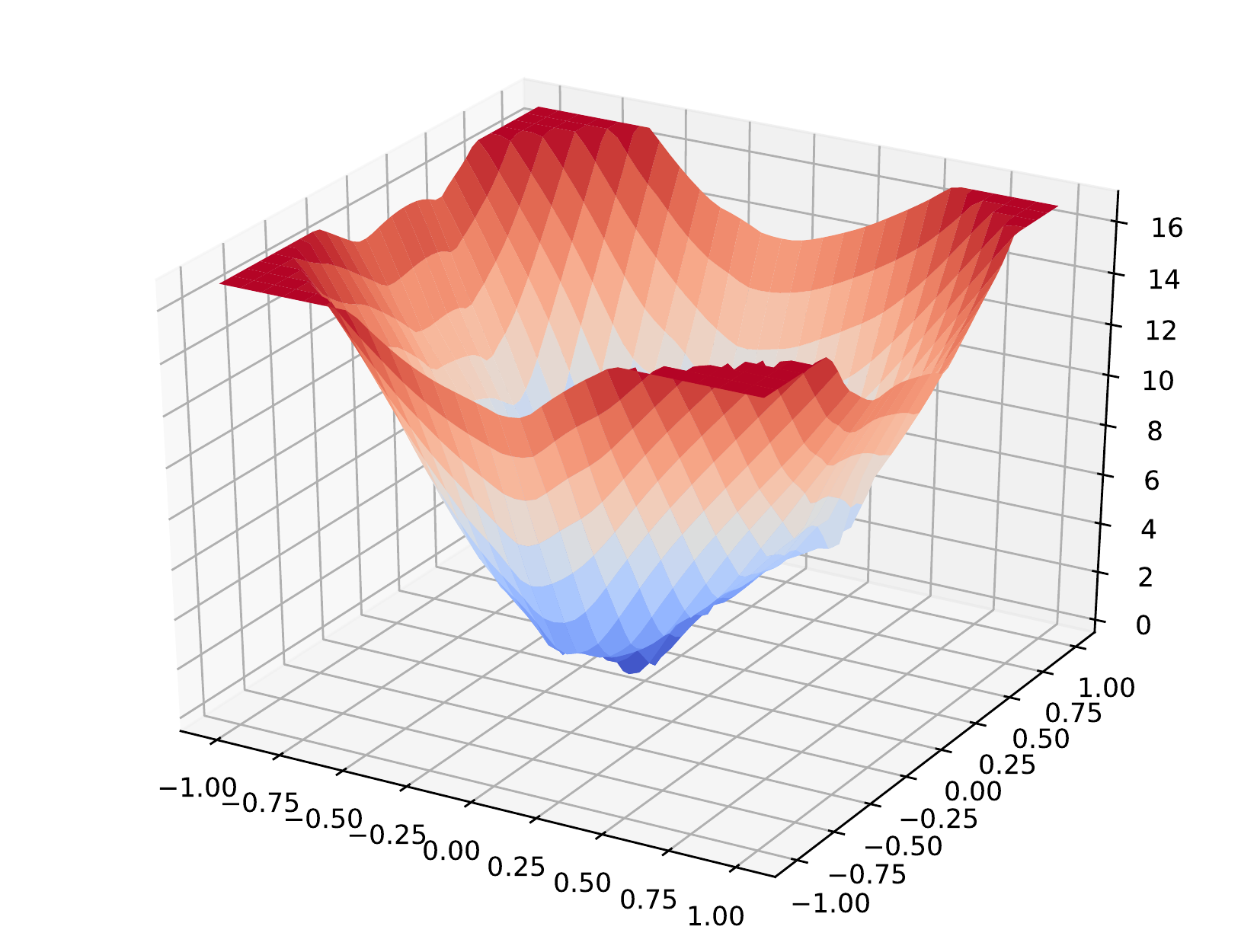}
			\subcaption[first]{\texttt{FedAvg}.}
		\end{minipage}
		\hfill
		\begin{minipage}{0.65\columnwidth}
			\centering
			\includegraphics[width=\textwidth]{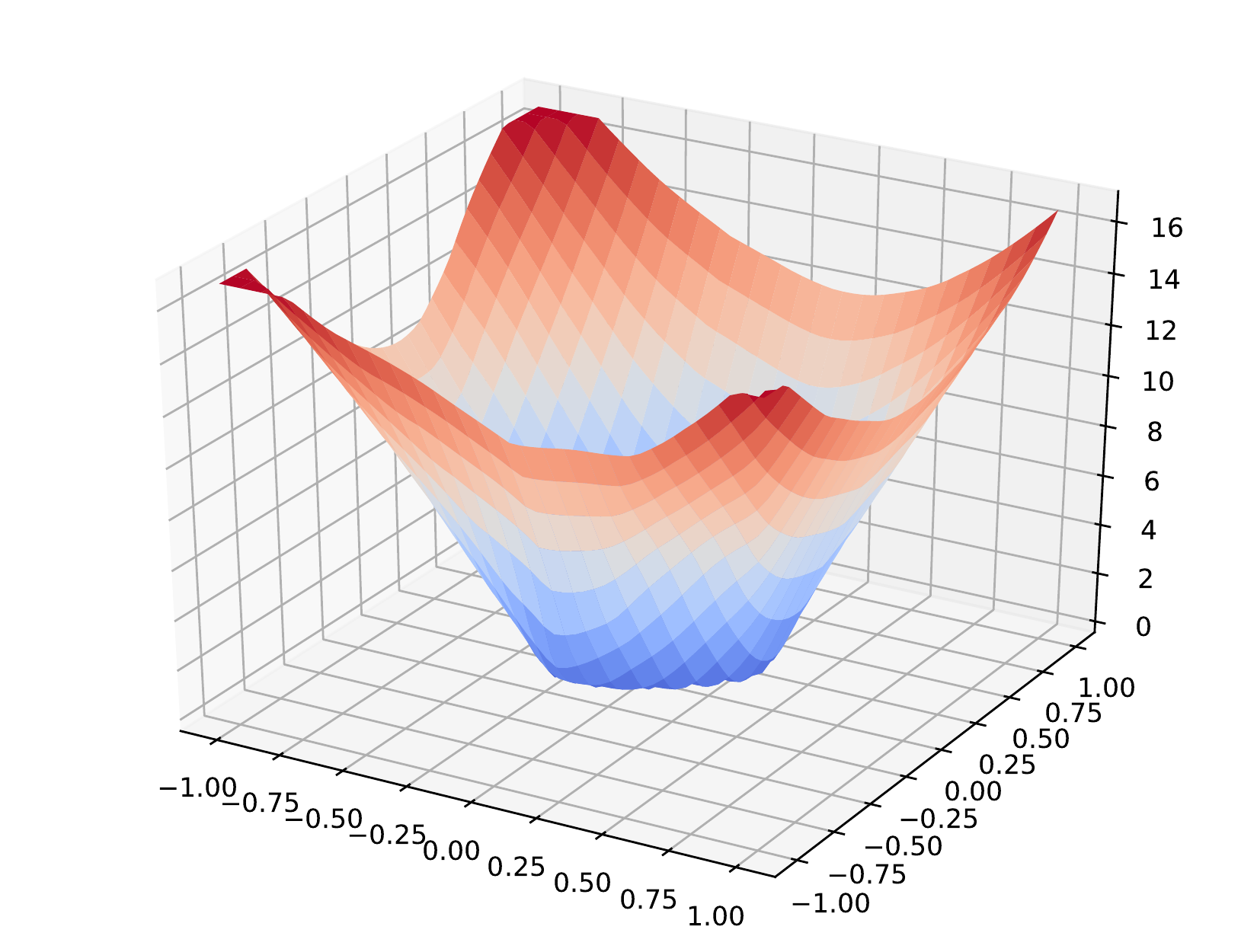}
			\subcaption[second]{\texttt{FedSAM}.}
		\end{minipage}%
		\hfill
		\begin{minipage}{0.65\columnwidth}
			\centering
			\includegraphics[width=\textwidth]{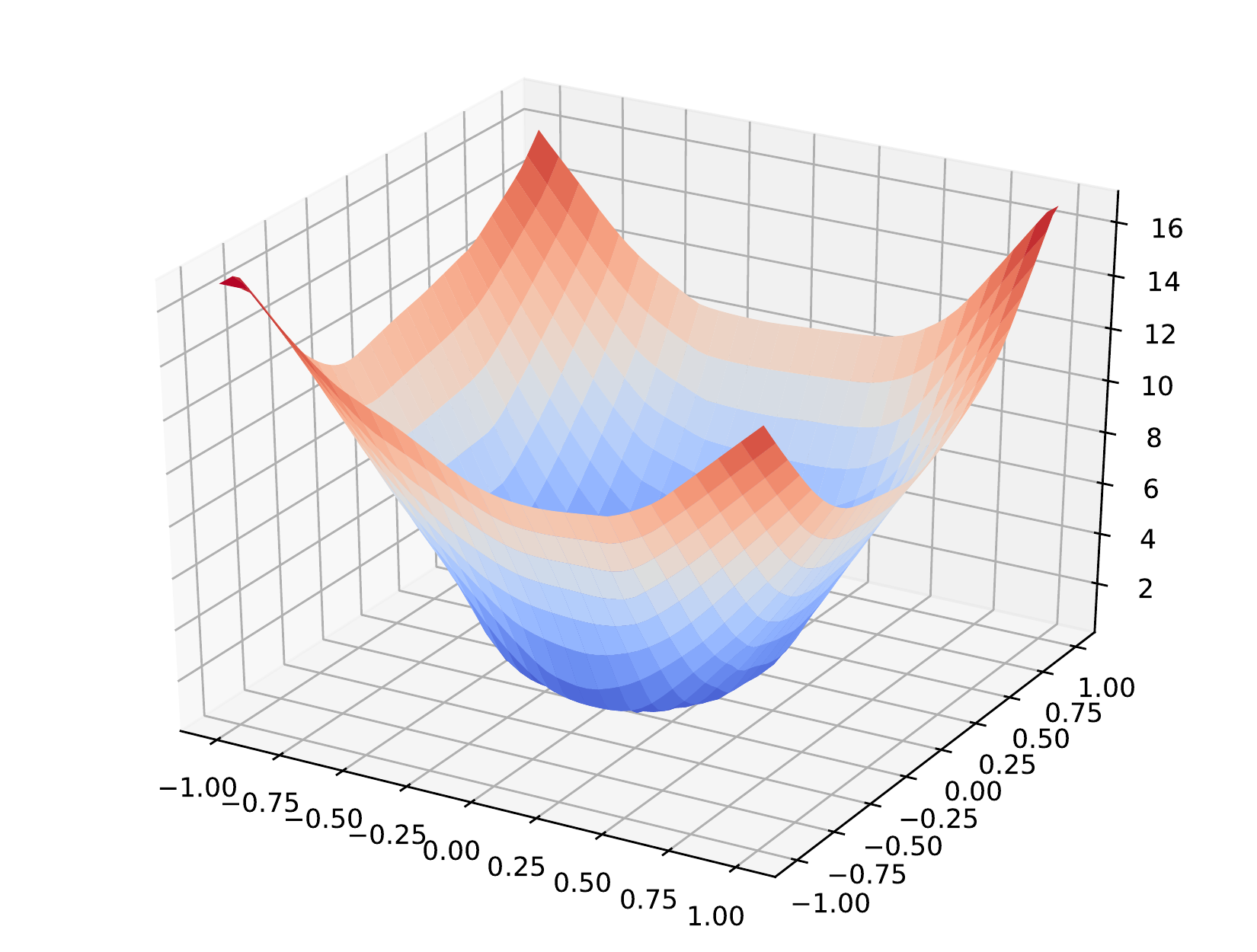}
			\subcaption[third]{\texttt{MoFedSAM}.}
		\end{minipage}
		\caption{Loss surface of \texttt{FedAvg}, \texttt{FedSAM} and \texttt{MoFedSAM} algorithm with ResNet-18 on CIFAR-10 dataset.}
		\label{Fig:losssurface}
	\end{figure*}

	\begin{figure*}[t!]
		\centering
		\begin{minipage}{0.48\columnwidth}
			\centering
			\includegraphics[width=\textwidth]{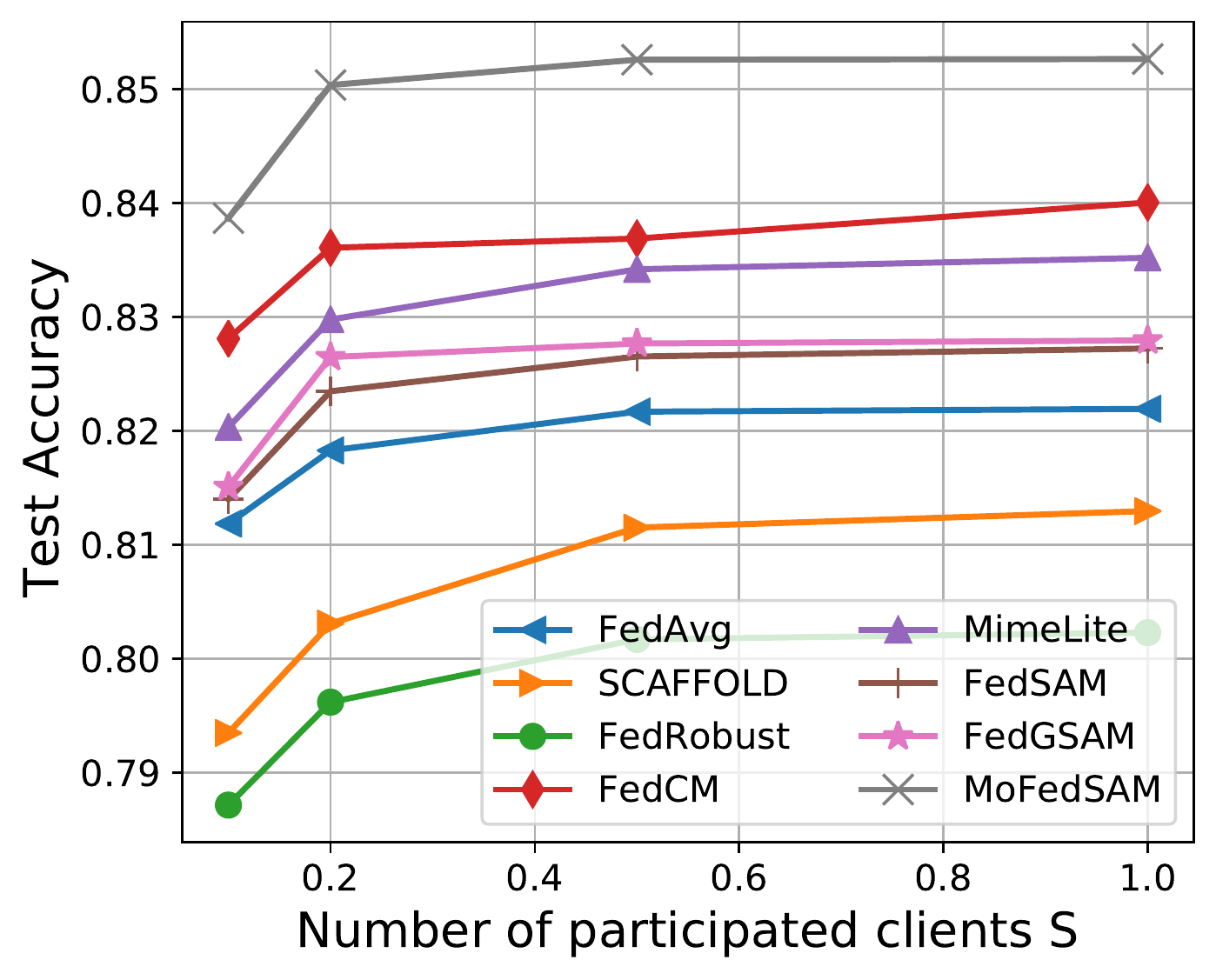}
			\subcaption[second]{Impact of $S$.}
			\label{Fig:rhocifar10}
		\end{minipage}%
		\hfill
		\begin{minipage}{0.48\columnwidth}
			\centering
			\includegraphics[width=\textwidth]{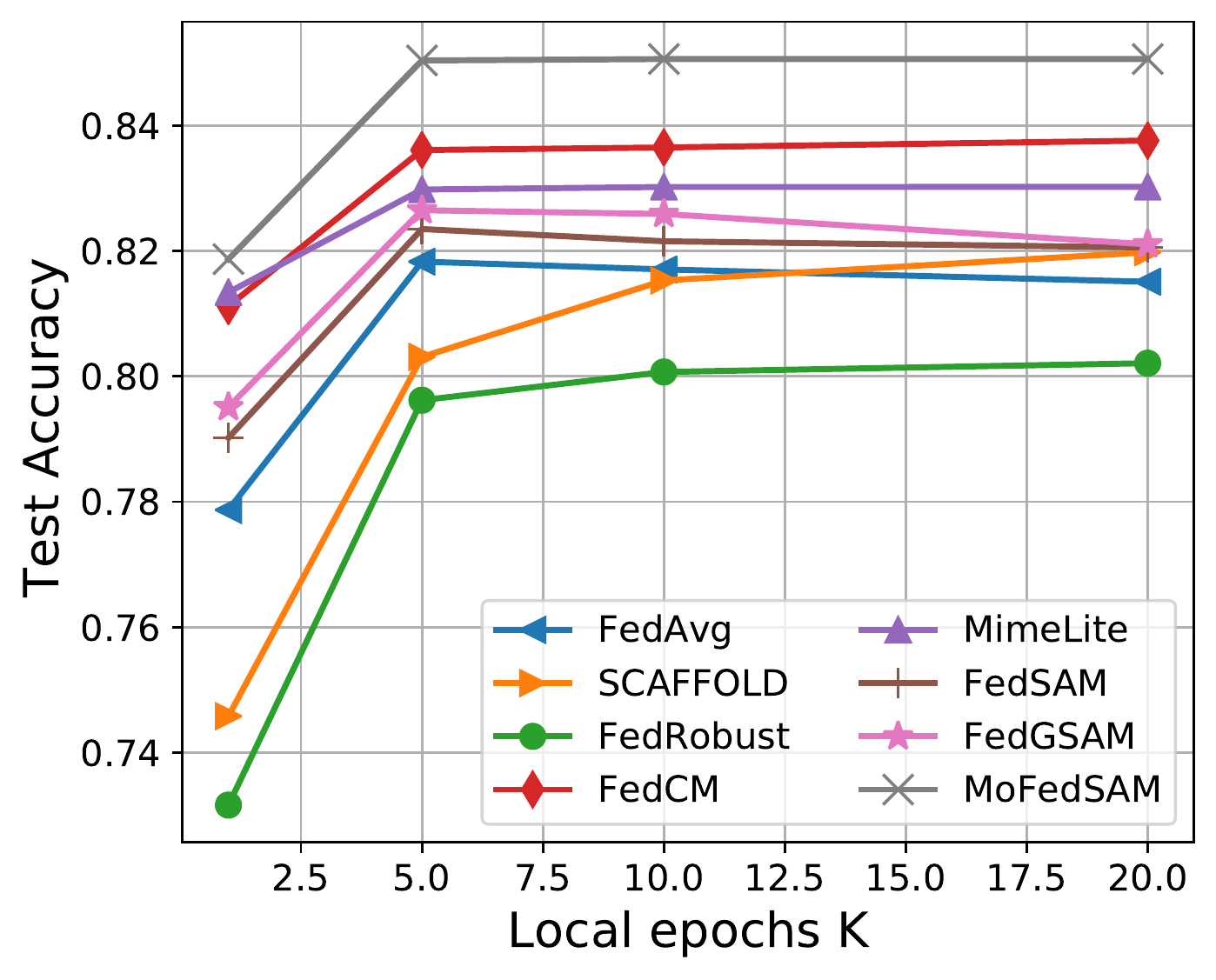}
			\subcaption[third]{Impact of $K$.}
			\label{Fig:Kcifar10}
		\end{minipage}
		\hfill
		\begin{minipage}{0.48\columnwidth}
			\centering
			\includegraphics[width=\textwidth]{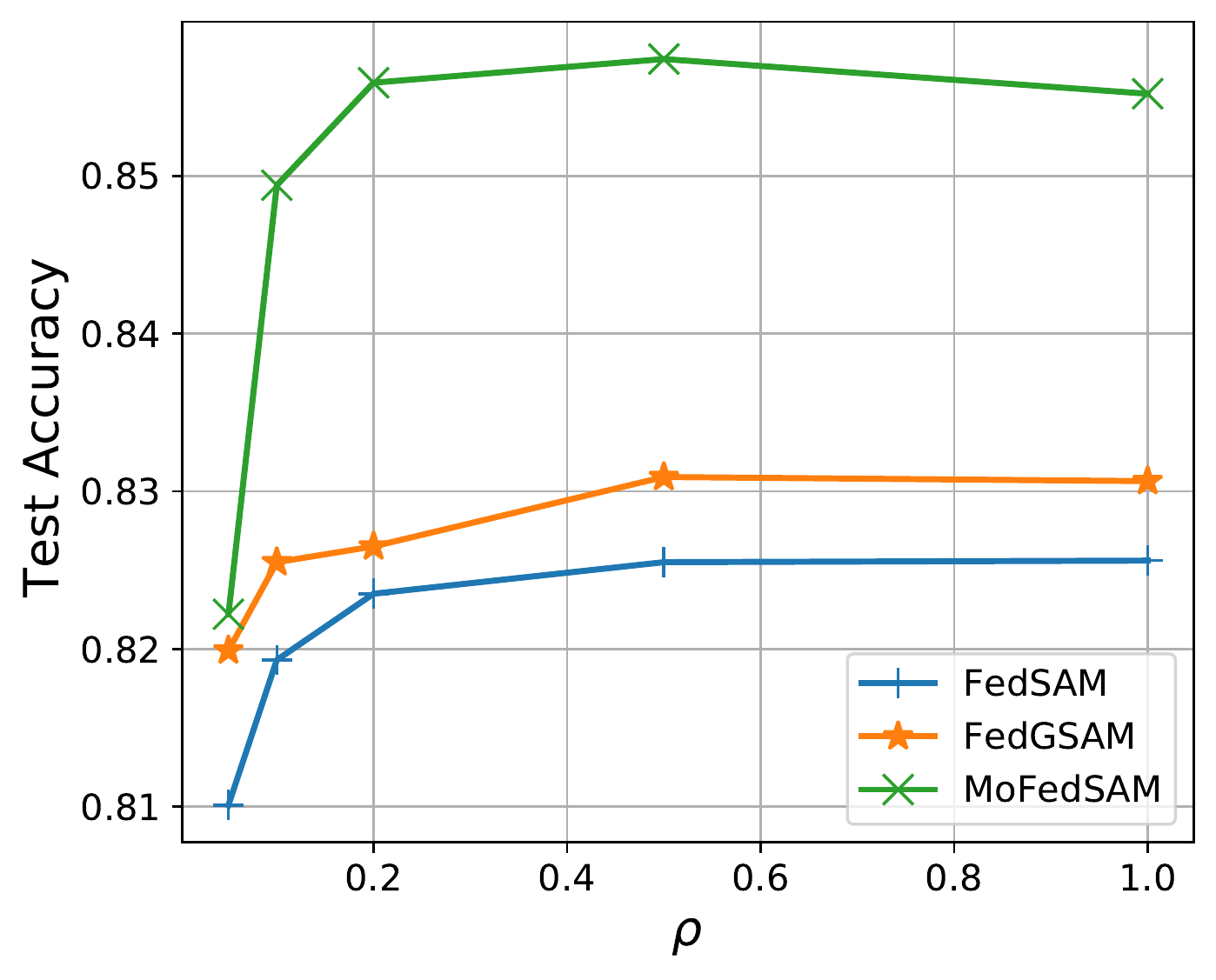}
			\subcaption[fourth]{Impact of $\rho$.}
			\label{Fig:Scifar10}
		\end{minipage}
		\hfill
		\begin{minipage}{0.48\columnwidth}
			\centering
			\includegraphics[width=\textwidth]{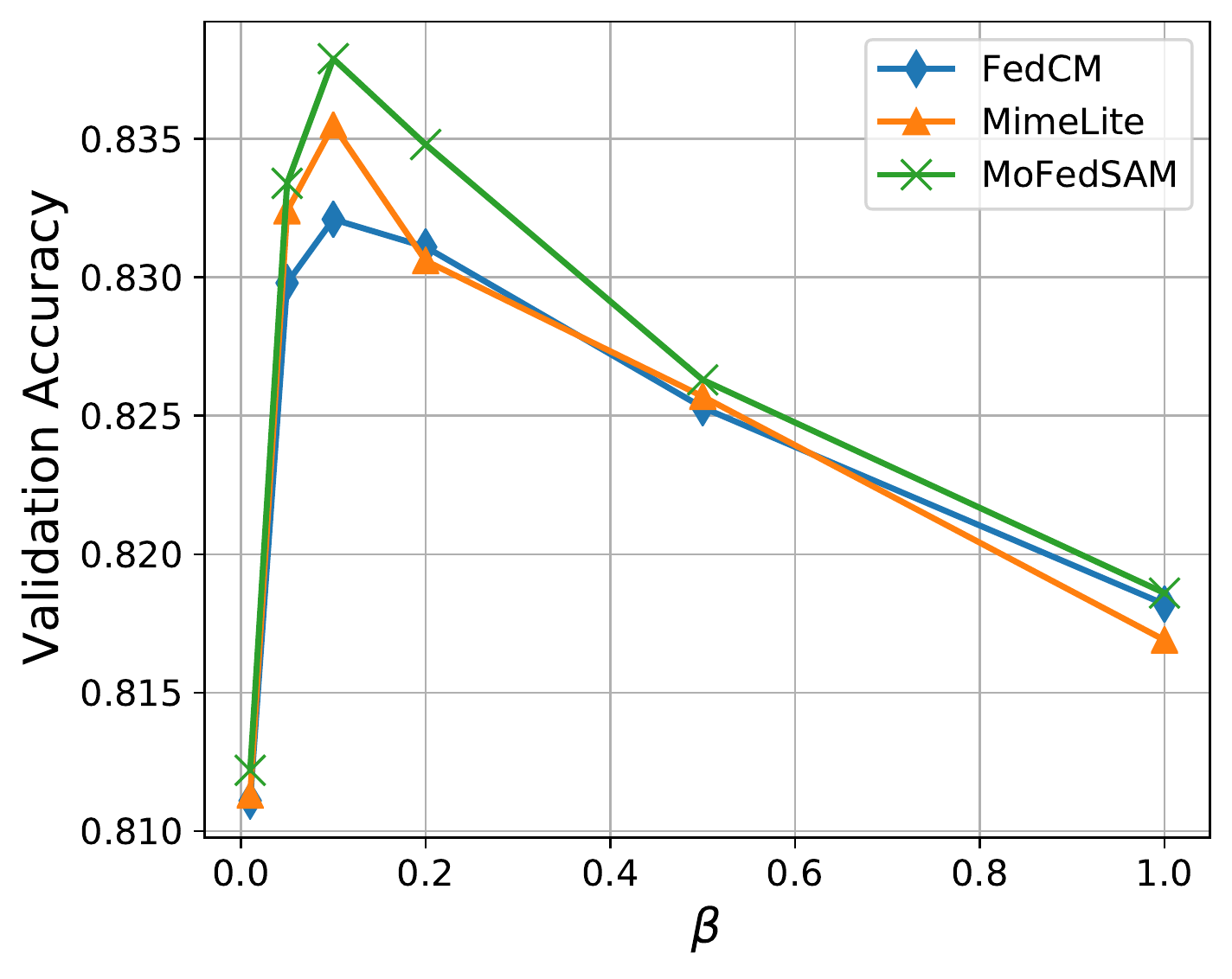}
			\subcaption[fourth]{Impact of $\beta$.}
			\label{Fig:betacifar10}
		\end{minipage}
		\caption{Impacts of different parameters on CIFAR-10 dataset.}
		\label{Fig:impactcifar10}
	\end{figure*}
	
	\subsection{Performance Evaluation}
	\textbf{(1) Performance with compared benchmarks.} We first investigate the effect of our proposed algorithms with compared benchmarks on different datasets in Figure~\ref{Fig:dataset} and Table~\ref{Tab:compare}. From these results, we can clearly see that for the performance of without momentum FL: \texttt{FedSAM} $>$ \texttt{FedAvg} $>$ \texttt{SCAFFOLD} $>$ \texttt{FedRobust}, and the performance momentum FL: \texttt{MoFedSAM} $>$ \texttt{MimeLite} $>$ \texttt{FedCM}. Our proposed algorithms outperform other benchmarks both on accuracy and convergence perspectives. We do not compare the FL algorithms with momentum FL, since momentum FL is required to transmit more information than FL, e.g., $\Delta^{r+1}$. This is the reason why momentum FL outperforms FL benchmarks. More specifically, to present the generalization performance, we show the deviation, i.e., best and worst local accuracy. In addition, the performance improvement on CIFAR-100 dataset is more obvious than others, since SAM optimizers perform more efficiently on more complicated datasets.

	\textbf{(2) Impact of Non-IID levels.} In Tables~\ref{Tab:heterogeneous}, 3, 4 and 5, we can see that our proposed algorithms outperforms the benchmarks across different client distribution levels on the same FL categories. We consider heterogeneous client distributions by varying balanced-unbalanced, number of clients and participation levels settings on various datasets. Client distributions become more non-IID as we go from IID, Dirichlet 0.6 to Dirichlet 0.3 splits which makes global optimization more difficult. For example, as non-IID levels increasing, \texttt{MoFedSAM} achieves a higher test accuracy 0.43\%, 1.24\% and 1.52\% and saving communication round 7, 40, and 59 than \texttt{MimeLite} on CIFAR-10 dataset. In summary, although almost all the algorithms perform well enough for training dataset, the testing accuracy usually has a significant degradation especially the deviation of local clients. In Table~\ref{Tab:compare}, we can see that our proposed algorithms significantly decrease the deviation of local clients, which indicates that our proposed algorithms show enough generalization of the global model.

	\textbf{(3) Loss surface visualization.} To visualize the sharpness of the flat minima obtained by \texttt{FedAvg}, \texttt{FedSAM} and \texttt{MoFedSAM}, we show the loss surface, which are trained with ResNet-18 under the CIFAR-10 dataset. We display the loss surfaces in  Figure~\ref{Fig:impactcifar10}, following the plotting algorithm in \cite{li2018visualizing}. The $x$- and $y$-axes are two random sampled orthogonal Gaussian perturbations. We can clearly see that both \texttt{FedSAM} and \texttt{MoFedSAM} improve the sharpness significantly in comparison to \texttt{FedAvg}, which indicates that our proposed algorithms perform more generalization.
	
	\textbf{(4) Impact of other parameters.} Here, we show the impact of different parameters, e.g., number of participated clients $S$, number of epochs $K$, perturbation radius $\rho$ for our proposed algorithms and momentum value $\beta$ in Figures~3, 7, 8 and 9. Our proposed algorithms outperform the same FL categories, i.e., with or without momentum. Similar to existing FL studies, increasing batch size and number of participated clients can improve the learning performance. Increasing the number of epochs $K$ cannot guarantee better accuracy substantially, however, all the benchmarks perform worst when $K=1$. The best $\rho$ for each dataset is different, the best performance of $\rho$ value is set as $0.2$ for EMNIST, $0.5$ for CIFAR-10 and $0.6$ for CIFAR-100.
	
	\section{Conclusion}
	In this paper, we study the distribution shift coming from the data heterogeneity challenge of cross-device FL from a simple yet unique perspective by making global model generality. To this end, we propose two algorithms \texttt{FedSAM} and \texttt{MoFedSAM}, which do not generate more communication costs compared with existing FL studies. By deriving the convergence of general non-convex FL settings, these algorithms achieve competitive performance. Furthermore, we also provide the generalization bound of \texttt{FedSAM} algorithm. The extensive experiments strongly support that our proposed algorithms decrease the performance deviation among all local clients significantly.
	
	
	\bibliography{example_paper}

\begin{thebibliography}{53}
\providecommand{\natexlab}[1]{#1}
\providecommand{\url}[1]{\texttt{#1}}
\expandafter\ifx\csname urlstyle\endcsname\relax
  \providecommand{\doi}[1]{doi: #1}\else
  \providecommand{\doi}{doi: \begingroup \urlstyle{rm}\Url}\fi

\bibitem[Acar et~al.(2021)Acar, Zhao, Matas, Mattina, Whatmough, and
  Saligrama]{acar2021federated}
Acar, D. A.~E., Zhao, Y., Matas, R., Mattina, M., Whatmough, P., and Saligrama,
  V.
\newblock Federated learning based on dynamic regularization.
\newblock In \emph{International Conference on Learning Representations}, 2021.

\bibitem[Bartlett et~al.(2017)Bartlett, Foster, and
  Telgarsky]{bartlett2017spectrally}
Bartlett, P.~L., Foster, D.~J., and Telgarsky, M.~J.
\newblock Spectrally-normalized margin bounds for neural networks.
\newblock \emph{Advances in Neural Information Processing Systems},
  30:\penalty0 6240--6249, 2017.

\bibitem[Caldarola et~al.(2022)Caldarola, Caputo, and
  Ciccone]{caldarola2022improving}
Caldarola, D., Caputo, B., and Ciccone, M.
\newblock Improving generalization in federated learning by seeking flat
  minima.
\newblock \emph{arXiv preprint arXiv:2203.11834}, 2022.

\bibitem[Chatterji et~al.(2019)Chatterji, Neyshabur, and
  Sedghi]{chatterji2019intriguing}
Chatterji, N., Neyshabur, B., and Sedghi, H.
\newblock The intriguing role of module criticality in the generalization of
  deep networks.
\newblock In \emph{International Conference on Learning Representations}, 2019.

\bibitem[Chaudhari et~al.(2019)Chaudhari, Choromanska, Soatto, LeCun, Baldassi,
  Borgs, Chayes, Sagun, and Zecchina]{chaudhari2019entropy}
Chaudhari, P., Choromanska, A., Soatto, S., LeCun, Y., Baldassi, C., Borgs, C.,
  Chayes, J., Sagun, L., and Zecchina, R.
\newblock Entropy-sgd: Biasing gradient descent into wide valleys.
\newblock \emph{Journal of Statistical Mechanics: Theory and Experiment},
  2019\penalty0 (12):\penalty0 124018, 2019.

\bibitem[Cohen et~al.(2017)Cohen, Afshar, Tapson, and
  Van~Schaik]{cohen2017emnist}
Cohen, G., Afshar, S., Tapson, J., and Van~Schaik, A.
\newblock Emnist: Extending mnist to handwritten letters.
\newblock In \emph{2017 International Joint Conference on Neural Networks
  (IJCNN)}, pp.\  2921--2926. IEEE, 2017.

\bibitem[Deng et~al.(2020)Deng, Kamani, and Mahdavi]{deng2020adaptive}
Deng, Y., Kamani, M.~M., and Mahdavi, M.
\newblock Adaptive personalized federated learning.
\newblock \emph{arXiv preprint arXiv:2003.13461}, 2020.

\bibitem[Dieuleveut et~al.(2021)Dieuleveut, Fort, Moulines, and
  Robin]{dieuleveut2021federated}
Dieuleveut, A., Fort, G., Moulines, E., and Robin, G.
\newblock Federated-em with heterogeneity mitigation and variance reduction.
\newblock \emph{Advances in Neural Information Processing Systems}, 34, 2021.

\bibitem[Du et~al.(2021{\natexlab{a}})Du, Yan, Feng, Zhou, Zhen, Goh, and
  Tan]{du2021efficient}
Du, J., Yan, H., Feng, J., Zhou, J.~T., Zhen, L., Goh, R. S.~M., and Tan, V.~Y.
\newblock Efficient sharpness-aware minimization for improved training of
  neural networks.
\newblock \emph{arXiv preprint arXiv:2110.03141}, 2021{\natexlab{a}}.

\bibitem[Du et~al.(2021{\natexlab{b}})Du, Xu, Wu, and Tong]{du2021fairness}
Du, W., Xu, D., Wu, X., and Tong, H.
\newblock Fairness-aware agnostic federated learning.
\newblock In \emph{Proceedings of the 2021 SIAM International Conference on
  Data Mining (SDM)}, pp.\  181--189. SIAM, 2021{\natexlab{b}}.

\bibitem[Fallah et~al.(2020)Fallah, Mokhtari, and
  Ozdaglar]{fallah2020personalized}
Fallah, A., Mokhtari, A., and Ozdaglar, A.
\newblock Personalized federated learning with theoretical guarantees: A
  model-agnostic meta-learning approach.
\newblock \emph{Advances in Neural Information Processing Systems},
  33:\penalty0 3557--3568, 2020.

\bibitem[Farnia et~al.(2018)Farnia, Zhang, and Tse]{farnia2018generalizable}
Farnia, F., Zhang, J., and Tse, D.
\newblock Generalizable adversarial training via spectral normalization.
\newblock In \emph{International Conference on Learning Representations}, 2018.

\bibitem[Foret et~al.(2021)Foret, Kleiner, Mobahi, and
  Neyshabur]{foret2021sharpnessaware}
Foret, P., Kleiner, A., Mobahi, H., and Neyshabur, B.
\newblock Sharpness-aware minimization for efficiently improving
  generalization.
\newblock In \emph{International Conference on Learning Representations}, 2021.

\bibitem[Gong et~al.(2021)Gong, Sharma, Karanam, Wu, Chen, Doermann, and
  Innanje]{gong2021ensemble}
Gong, X., Sharma, A., Karanam, S., Wu, Z., Chen, T., Doermann, D., and Innanje,
  A.
\newblock Ensemble attention distillation for privacy-preserving federated
  learning.
\newblock In \emph{Proceedings of the IEEE/CVF International Conference on
  Computer Vision}, pp.\  15076--15086, 2021.

\bibitem[Goodfellow et~al.(2014)Goodfellow, Shlens, and
  Szegedy]{goodfellow2014explaining}
Goodfellow, I.~J., Shlens, J., and Szegedy, C.
\newblock Explaining and harnessing adversarial examples.
\newblock \emph{arXiv preprint arXiv:1412.6572}, 2014.

\bibitem[Goyal et~al.(2017)Goyal, Doll{\'a}r, Girshick, Noordhuis, Wesolowski,
  Kyrola, Tulloch, Jia, and He]{goyal2017accurate}
Goyal, P., Doll{\'a}r, P., Girshick, R., Noordhuis, P., Wesolowski, L., Kyrola,
  A., Tulloch, A., Jia, Y., and He, K.
\newblock Accurate, large minibatch sgd: Training imagenet in 1 hour.
\newblock \emph{arXiv preprint arXiv:1706.02677}, 2017.

\bibitem[Hard et~al.(2018)Hard, Rao, Mathews, Ramaswamy, Beaufays, Augenstein,
  Eichner, Kiddon, and Ramage]{hard2018federated}
Hard, A., Rao, K., Mathews, R., Ramaswamy, S., Beaufays, F., Augenstein, S.,
  Eichner, H., Kiddon, C., and Ramage, D.
\newblock Federated learning for mobile keyboard prediction.
\newblock \emph{arXiv preprint arXiv:1811.03604}, 2018.

\bibitem[He et~al.(2016)He, Zhang, Ren, and Sun]{he2016deep}
He, K., Zhang, X., Ren, S., and Sun, J.
\newblock Deep residual learning for image recognition.
\newblock In \emph{Proceedings of the IEEE conference on computer vision and
  pattern recognition}, pp.\  770--778, 2016.

\bibitem[Kairouz et~al.(2019)Kairouz, McMahan, Avent, Bellet, Bennis, Bhagoji,
  Bonawitz, Charles, Cormode, Cummings, et~al.]{kairouz2019advances}
Kairouz, P., McMahan, H.~B., Avent, B., Bellet, A., Bennis, M., Bhagoji, A.~N.,
  Bonawitz, K., Charles, Z., Cormode, G., Cummings, R., et~al.
\newblock Advances and open problems in federated learning.
\newblock \emph{arXiv preprint arXiv:1912.04977}, 2019.

\bibitem[Karimireddy et~al.(2020)Karimireddy, Kale, Mohri, Reddi, Stich, and
  Suresh]{karimireddy2020scaffold}
Karimireddy, S.~P., Kale, S., Mohri, M., Reddi, S., Stich, S., and Suresh,
  A.~T.
\newblock Scaffold: Stochastic controlled averaging for federated learning.
\newblock In \emph{International Conference on Machine Learning}, pp.\
  5132--5143. PMLR, 2020.

\bibitem[Karimireddy et~al.(2021)Karimireddy, Jaggi, Kale, Mohri, Reddi, Stich,
  and Suresh]{karimireddy2021breaking}
Karimireddy, S.~P., Jaggi, M., Kale, S., Mohri, M., Reddi, S.~J., Stich, S.~U.,
  and Suresh, A.~T.
\newblock Breaking the centralized barrier for cross-device federated learning.
\newblock In \emph{Thirty-Fifth Conference on Neural Information Processing
  Systems}, 2021.

\bibitem[Keskar et~al.(2016)Keskar, Mudigere, Nocedal, Smelyanskiy, and
  Tang]{keskar2016large}
Keskar, N.~S., Mudigere, D., Nocedal, J., Smelyanskiy, M., and Tang, P. T.~P.
\newblock On large-batch training for deep learning: Generalization gap and
  sharp minima.
\newblock \emph{arXiv preprint arXiv:1609.04836}, 2016.

\bibitem[Khanduri et~al.(2021)Khanduri, SHARMA, Yang, Hong, Liu, Rajawat, and
  Varshney]{khanduri2021stem}
Khanduri, P., SHARMA, P., Yang, H., Hong, M., Liu, J., Rajawat, K., and
  Varshney, P.
\newblock {STEM}: A stochastic two-sided momentum algorithm achieving
  near-optimal sample and communication complexities for federated learning.
\newblock In \emph{Advances in Neural Information Processing Systems}, 2021.

\bibitem[Krizhevsky et~al.(2009)Krizhevsky, Hinton,
  et~al.]{krizhevsky2009learning}
Krizhevsky, A., Hinton, G., et~al.
\newblock Learning multiple layers of features from tiny images.
\newblock 2009.

\bibitem[Kwon et~al.(2021)Kwon, Kim, Park, and Choi]{kwon21b}
Kwon, J., Kim, J., Park, H., and Choi, I.~K.
\newblock Asam: Adaptive sharpness-aware minimization for scale-invariant
  learning of deep neural networks.
\newblock In \emph{Proceedings of the 38th International Conference on Machine
  Learning}, pp.\  5905--5914. PMLR, 2021.

\bibitem[Lakshminarayanan et~al.(2017)Lakshminarayanan, Pritzel, and
  Blundell]{lakshminarayanan2017simple}
Lakshminarayanan, B., Pritzel, A., and Blundell, C.
\newblock Simple and scalable predictive uncertainty estimation using deep
  ensembles.
\newblock \emph{Advances in Neural Information Processing Systems}, 30, 2017.

\bibitem[Li et~al.(2018{\natexlab{a}})Li, Xu, Taylor, Studer, and
  Goldstein]{li2018visualizing}
Li, H., Xu, Z., Taylor, G., Studer, C., and Goldstein, T.
\newblock Visualizing the loss landscape of neural nets.
\newblock \emph{Advances in Neural Information Processing Systems}, 31,
  2018{\natexlab{a}}.

\bibitem[Li et~al.(2018{\natexlab{b}})Li, Sahu, Zaheer, Sanjabi, Talwalkar, and
  Smith]{li2018federated}
Li, T., Sahu, A.~K., Zaheer, M., Sanjabi, M., Talwalkar, A., and Smith, V.
\newblock Federated optimization in heterogeneous networks.
\newblock \emph{arXiv preprint arXiv:1812.06127}, 2018{\natexlab{b}}.

\bibitem[Li et~al.(2020{\natexlab{a}})Li, Sanjabi, Beirami, and
  Smith]{Li2020Fair}
Li, T., Sanjabi, M., Beirami, A., and Smith, V.
\newblock Fair resource allocation in federated learning.
\newblock In \emph{International Conference on Learning Representations},
  2020{\natexlab{a}}.

\bibitem[Li et~al.(2021)Li, Hu, Beirami, and Smith]{li2021ditto}
Li, T., Hu, S., Beirami, A., and Smith, V.
\newblock Ditto: Fair and robust federated learning through personalization.
\newblock In \emph{International Conference on Machine Learning}, pp.\
  6357--6368. PMLR, 2021.

\bibitem[Li et~al.(2020{\natexlab{b}})Li, Huang, Yang, Wang, and
  Zhang]{Li2020On}
Li, X., Huang, K., Yang, W., Wang, S., and Zhang, Z.
\newblock On the convergence of fedavg on non-iid data.
\newblock In \emph{International Conference on Learning Representations},
  2020{\natexlab{b}}.

\bibitem[Lian et~al.(2017)Lian, Zhang, Zhang, Hsieh, Zhang, and
  Liu]{lian2017can}
Lian, X., Zhang, C., Zhang, H., Hsieh, C.-J., Zhang, W., and Liu, J.
\newblock Can decentralized algorithms outperform centralized algorithms? a
  case study for decentralized parallel stochastic gradient descent.
\newblock \emph{arXiv preprint arXiv:1705.09056}, 2017.

\bibitem[Lin et~al.(2020)Lin, Kong, Stich, and Jaggi]{lin2020ensemble}
Lin, T., Kong, L., Stich, S.~U., and Jaggi, M.
\newblock Ensemble distillation for robust model fusion in federated learning.
\newblock In \emph{Advances in Neural Information Processing Systems}, 2020.

\bibitem[McMahan et~al.(2017)McMahan, Moore, Ramage, Hampson, and
  y~Arcas]{mcmahan2017communication}
McMahan, B., Moore, E., Ramage, D., Hampson, S., and y~Arcas, B.~A.
\newblock Communication-efficient learning of deep networks from decentralized
  data.
\newblock In \emph{Artificial intelligence and statistics}, pp.\  1273--1282.
  PMLR, 2017.

\bibitem[Mendieta et~al.(2021)Mendieta, Yang, Wang, Lee, Ding, and
  Chen]{mendieta2021local}
Mendieta, M., Yang, T., Wang, P., Lee, M., Ding, Z., and Chen, C.
\newblock Local learning matters: Rethinking data heterogeneity in federated
  learning.
\newblock \emph{arXiv preprint arXiv:2111.14213}, 2021.

\bibitem[Mohri et~al.(2019)Mohri, Sivek, and Suresh]{mohri2019agnostic}
Mohri, M., Sivek, G., and Suresh, A.~T.
\newblock Agnostic federated learning.
\newblock In \emph{International Conference on Machine Learning}, pp.\
  4615--4625. PMLR, 2019.

\bibitem[Nesterov \& Spokoiny(2017)Nesterov and Spokoiny]{nesterov2017random}
Nesterov, Y. and Spokoiny, V.
\newblock Random gradient-free minimization of convex functions.
\newblock \emph{Foundations of Computational Mathematics}, 17\penalty0
  (2):\penalty0 527--566, 2017.

\bibitem[Neyshabur et~al.(2018)Neyshabur, Bhojanapalli, and
  Srebro]{neyshabur2018pac}
Neyshabur, B., Bhojanapalli, S., and Srebro, N.
\newblock A pac-bayesian approach to spectrally-normalized margin bounds for
  neural networks.
\newblock In \emph{International Conference on Learning Representations}, 2018.

\bibitem[Paszke et~al.(2019)Paszke, Gross, Massa, Lerer, Bradbury, Chanan,
  Killeen, Lin, Gimelshein, Antiga, et~al.]{paszke2019pytorch}
Paszke, A., Gross, S., Massa, F., Lerer, A., Bradbury, J., Chanan, G., Killeen,
  T., Lin, Z., Gimelshein, N., Antiga, L., et~al.
\newblock Pytorch: An imperative style, high-performance deep learning library.
\newblock \emph{Advances in neural information processing systems},
  32:\penalty0 8026--8037, 2019.

\bibitem[Reddi et~al.(2020)Reddi, Charles, Zaheer, Garrett, Rush,
  Kone{\v{c}}n{\`y}, Kumar, and McMahan]{reddi2020adaptive}
Reddi, S.~J., Charles, Z., Zaheer, M., Garrett, Z., Rush, K.,
  Kone{\v{c}}n{\`y}, J., Kumar, S., and McMahan, H.~B.
\newblock Adaptive federated optimization.
\newblock In \emph{International Conference on Learning Representations}, 2020.

\bibitem[Reisizadeh et~al.(2020)Reisizadeh, Farnia, Pedarsani, and
  Jadbabaie]{reisizadeh2020robust}
Reisizadeh, A., Farnia, F., Pedarsani, R., and Jadbabaie, A.
\newblock Robust federated learning: The case of affine distribution shifts.
\newblock In \emph{NeurIPS}, 2020.

\bibitem[Shafahi et~al.(2020)Shafahi, Najibi, Xu, Dickerson, Davis, and
  Goldstein]{shafahi2020universal}
Shafahi, A., Najibi, M., Xu, Z., Dickerson, J., Davis, L.~S., and Goldstein, T.
\newblock Universal adversarial training.
\newblock In \emph{Proceedings of the AAAI Conference on Artificial
  Intelligence}, volume~34, pp.\  5636--5643, 2020.

\bibitem[Singhal et~al.(2021)Singhal, Sidahmed, Garrett, Wu, Rush, and
  Prakash]{singhal2021federated}
Singhal, K., Sidahmed, H., Garrett, Z., Wu, S., Rush, J.~K., and Prakash, S.
\newblock Federated reconstruction: Partially local federated learning.
\newblock In \emph{Advances in Neural Information Processing Systems}, 2021.

\bibitem[T~Dinh et~al.(2020)T~Dinh, Tran, and Nguyen]{t2020personalized}
T~Dinh, C., Tran, N., and Nguyen, T.~D.
\newblock Personalized federated learning with moreau envelopes.
\newblock \emph{Advances in Neural Information Processing Systems}, 33, 2020.

\bibitem[Tropp(2012)]{tropp2012user}
Tropp, J.~A.
\newblock User-friendly tail bounds for sums of random matrices.
\newblock \emph{Foundations of computational mathematics}, 12\penalty0
  (4):\penalty0 389--434, 2012.

\bibitem[Wang et~al.(2019)Wang, Tantia, Ballas, and Rabbat]{wang2019slowmo}
Wang, J., Tantia, V., Ballas, N., and Rabbat, M.
\newblock Slowmo: Improving communication-efficient distributed sgd with slow
  momentum.
\newblock In \emph{International Conference on Learning Representations}, 2019.

\bibitem[Woodworth et~al.(2020)Woodworth, Gunasekar, Lee, Moroshko, Savarese,
  Golan, Soudry, and Srebro]{woodworth2020kernel}
Woodworth, B., Gunasekar, S., Lee, J.~D., Moroshko, E., Savarese, P., Golan,
  I., Soudry, D., and Srebro, N.
\newblock Kernel and rich regimes in overparametrized models.
\newblock In \emph{Conference on Learning Theory}, pp.\  3635--3673. PMLR,
  2020.

\bibitem[Xu et~al.(2021)Xu, Wang, Wang, and Yao]{xu2021fedcm}
Xu, J., Wang, S., Wang, L., and Yao, A. C.-C.
\newblock Fedcm: Federated learning with client-level momentum.
\newblock \emph{arXiv preprint arXiv:2106.10874}, 2021.

\bibitem[Yang et~al.(2021)Yang, Fang, and Liu]{yang2021achieving}
Yang, H., Fang, M., and Liu, J.
\newblock Achieving linear speedup with partial worker participation in
  non-{IID} federated learning.
\newblock In \emph{International Conference on Learning Representations}, 2021.

\bibitem[Yoon et~al.(2021)Yoon, Shin, Hwang, and Yang]{yoon2021fedmix}
Yoon, T., Shin, S., Hwang, S.~J., and Yang, E.
\newblock Fedmix: Approximation of mixup under mean augmented federated
  learning.
\newblock In \emph{International Conference on Learning Representations}, 2021.

\bibitem[Yuan et~al.(2021)Yuan, Morningstar, Ning, and Singhal]{yuan2021we}
Yuan, H., Morningstar, W., Ning, L., and Singhal, K.
\newblock What do we mean by generalization in federated learning?
\newblock \emph{arXiv preprint arXiv:2110.14216}, 2021.

\bibitem[Zhu et~al.(2021)Zhu, Hong, and Zhou]{zhu2021data}
Zhu, Z., Hong, J., and Zhou, J.
\newblock Data-free knowledge distillation for heterogeneous federated
  learning.
\newblock \emph{arXiv preprint arXiv:2105.10056}, 2021.

\bibitem[Zhuang et~al.(2022)Zhuang, Gong, Yuan, Cui, Adam, Dvornek, sekhar
  tatikonda, s~Duncan, and Liu]{zhuang2022surrogate}
Zhuang, J., Gong, B., Yuan, L., Cui, Y., Adam, H., Dvornek, N.~C., sekhar
  tatikonda, s~Duncan, J., and Liu, T.
\newblock Surrogate gap minimization improves sharpness-aware training.
\newblock In \emph{International Conference on Learning Representations}, 2022.

\end{thebibliography}
	\bibliographystyle{icml2022}

	\clearpage
	\onecolumn
	\appendix
	\section{Preliminary Lemmas}\label{Sec:Lemmas}
	For giving the theoretical analysis of the convergence rate of all proposed algorithms, we firstly state some preliminary lemmas as follows:
	
	\begin{lemma}\label{lemma:triangleinequality}
		(Relaxed triangle inequality). Let $\{v_1 , \dots, v_\tau \}$ be $\tau$ vectors in $\mathbb{R}^d$. Then, the following are true: (1) $\|v_i + v_j \|^2 \leq (1+a)\|v_i \|^2 + (1+\frac{1}{a})\|v_j \|^2$ for any $a > 0$, and (2) $\|\sum_{i=1}^{\tau} v_i \|^2 \leq \tau \sum_{i=1}^{\tau}\|v_i \|^2$.
	\end{lemma}
	
	\begin{lemma}\label{lemma:dependent}
		For random variables $x_1 , \dots, x_n$, we have
		\begin{equation}
			\E [\|x_1 + \cdots + x_n \|^2 ] \leq n\E [\|x_1 \|^2 + \cdots + \|x_n \|^2 ].\nonumber
		\end{equation}
	\end{lemma}
	
	\begin{lemma}\label{lemma:independent}
		For independent, mean $0$ random variables $x_1 , \dots, x_n$, we have
		\begin{equation}
			\E [\|x_1 + \cdots + x_n \|^2 ] = \E [\|x_1 \|^2 + \cdots + \|x_n \|^2 ].\nonumber
		\end{equation}
	\end{lemma}

	\begin{lemma}\label{lemma:SAMvariance}
		(Separating mean and variance for \texttt{SAM}). The stochastic gradient $\nabla F_i (w,\xi_i )$ computed by the $i$-th client at model parameter $w$ using minibatch $\xi$ is an unbiased estimator of $\nabla F_i (w)$ with variance bounded by $\sigma^2$. The gradient of SAM is formulated by
		\begin{equation}
			\begin{split}
				\E\bigg[\bigg\|\sum_{k=0}^{K-1} g_{i,k}^r \bigg\|^2 \bigg]\leq K\sum_{k=0}^{K-1} \E [\|\nabla F_i (w_{r,k}^i )\|^2 ] +\frac{K L^2\rho^2}{N}\sigma_l^2 ,\\
				\E \bigg[\bigg\|\sum_{k=0}^{K-1} g_{i,k}^r \bigg\|^2 \bigg] \leq K\sum_{k=0}^{K-1} \E [\|\nabla F_i (w_{r,k}^i ) \|^2 ] + KL^2 \rho^2 \sigma_l^2 .\nonumber
			\end{split}
		\end{equation}
	\end{lemma}
	
	\noindent\textit{Proof.} For the first inequality, we can bound as follows
	\begin{equation}
		\begin{split}
			\E \bigg[\bigg\|\sum_{k=0}^{K-1} g_{i,k}^r \bigg\|^2 \bigg] & = \E\bigg[\bigg\|\sum_{k=0}^{K-1} g_{i,k}^r \bigg\|^2 \bigg] + \E\bigg[\bigg\|\sum_{k=0}^{K-1}( g_{i,k}^r - \nabla F(w_{i,k}^r ))\bigg\|^2 \bigg]\\
			&\overset{\text{(a)}}{\leq} K\sum_{k=0}^{K-1}\E [ \|g_{i,k}^r \|^2 ] + L^2 \sum_{k=0}^{K-1} \E\bigg[\bigg\|\frac{1}{N}\sum_{i\in[N]}(w_{i,k}^r + \delta_{i,k}^r (\tw_{i,k}^r ; \xi_{i,k}^r ) - w_{i,k}^r - \delta_{i,k}^r (\tw_{i,k}^r )) \bigg\|^2 \bigg] \\
			& \overset{\text{(b)}}{\leq} K\sum_{k=0}^{K-1}\E [ \|g_{i,k}^r \|^2 ] +  \frac{KL^2 \rho^2 \sigma_l^2}{N}.\nonumber
		\end{split}
	\end{equation}
	where (a) is from Assumption~\ref{ass:smooth} and (b) is from Assumption~\ref{ass:sigmal} and Lemma~\ref{lemma:independent}. Similarly, we can obtain the second inequality, and hence we omit it here. \hfill~$\Box$

	\begin{lemma}\label{lemma:sigmag}
		(Bounded global variance of $\|\nabla F_i (w + \delta_i ) - \nabla F(w + \delta)\|^2$.) An immediate implication of Assumptions~\ref{ass:smooth} and \ref{ass:sigmag}, the variance of local and global gradients with perturbation can be bounded as follows:
		\begin{equation}
			\|\nabla F_i (w + \delta_i ) - \nabla F(w + \delta)\|^2 \leq 3\sigma_g^2 + 6L^2 \rho^2 .\nonumber
		\end{equation}
	\end{lemma}
	
	\noindent\textit{Proof.} 
	\begin{equation}
		\begin{split}
			\|\nabla f_i (\tw) - \nabla f(\tw)\|^2 & = \|\nabla F_i (w + \delta_i ) - \nabla F(w + \delta)\|^2 \\
			& = \|\nabla F_i (w+\delta_i ) - \nabla F_i (w) + \nabla F_i (w) - \nabla F(w) + \nabla F(w) - \nabla F(w+\delta) \|^2 \\
			& \overset{\text{(a)}}{\leq} 3\|\nabla F_i (w+\delta_i ) - \nabla F_i (w)\|^2 + 3\|\nabla F_i (w) - \nabla F(w)\|^2 + 3\|\nabla F(w) - \nabla F(w+\delta)\|^2 \\
			& \overset{\text{(b)}}{\leq} 3\sigma_g^2 + 6L^2 \rho^2 ,\nonumber
		\end{split}
	\end{equation}
	where (a) is from Lemma~\ref{lemma:dependent} and (b) is from Assumption~\ref{ass:smooth}, \ref{ass:sigmag} and the perturbation is bounded by $\rho$. \hfill~$\Box$

	\section{Convergence Analysis for FedSAM}\label{Sec:convergencefedsam}
	
	\subsection{Description of \texttt{FedSAM} Algorithm and Key Lemmas}
	\begin{algorithm}[t!]
		\caption{\texttt{FedSAM}: Federated Sharpness Aware Minimization}
		\begin{algorithmic}[1]
			\STATE Initialization: $w_0$, $\rho_0$, $\gamma$ the number of local updates $K$, batch size $b$, local learning $\eta_l$ and global learning rate $\eta_g$.
			\FOR {each round $r = 0, \dots, R-1$}
			\STATE Sample subset $\S^r \subseteq [N]$ of clients.
			\STATE communicate $w^r$ to all clients $i \in \S^r$.
			\FOR {each client $i \in \S^r$ in parallel}
			\STATE initialize local model $w_{i,0}^r \gets w^r$.
			\FOR {$k=0,\dots, K-1$}
			\STATE Compute $g_{i,k-1}^r$ by taking an estimation $\nabla F_i (w_{i,k-1}^r , \xi_{i}^r)$ of $\nabla F_i (w_{i,k-1}^r )$.
			\STATE $\tw_{i,k-1}^r = w_{i,k-1}^r + \rho \frac{g_{i,k-1}^r}{\|g_{i,k-1}^r \|}$.
			\STATE Compute $\tg_{i,k-1}^r$ by taking an estimation $\nabla f_i (\tw_{i,k-1}^r , \xi_{i}^r)$ of $\nabla f_i (\tw_{i,k-1}^r , \xi_{i}^r)$.
			\STATE $w_{i,k}^r = w_{i,k-1}^r - \eta_l \tg_{i,k-1}^r$.
			\ENDFOR
			\STATE $\Delta_i^r = w_{i,K}^r - w^r$.
			\ENDFOR
			\STATE $\Delta^{r+1} = \frac{1}{S}\sum_{i \in \S^r}\Delta_i^r$.
			\STATE $w^{r+1} = w^r + \eta_g \Delta^r$.
			\ENDFOR
		\end{algorithmic}
		\label{Alg:FedSAM}
	\end{algorithm}

	We outline the \texttt{FedSAM} algorithm in Algorithm~\ref{Alg:FedSAM}. In round $r$, we sample $\S^r \subseteq [N]$ clients with $|\S^r | = S$ and then perform the following updates:
	
	\begin{itemize}
		\item Starting from the shared global parameters $w_{i,0}^r = w^{r-1}$, we update the local parameters for $k \in [K]$
		\begin{equation}
			\begin{split}
				\tw_{i,k}^r & = w_{i,k-1}^r + \rho \frac{g_{i,k-1}^r}{\|g_{i,k-1}^r \|} \\
				w_{i,k}^r & = w_{i,k-1}^r - \eta_l \tg_{i,k-1}^r ,\nonumber
			\end{split}
		\end{equation}
		\item After $K$ times local epochs, we obtain the following
		\begin{equation}
			\Delta_i^r = w_{i,K}^r - w^r .
		\end{equation}
		\item Compute the new global parameters using only updates from the clients $i \in \S^r$ and a global step-size $\eta_g$:
		\begin{equation}
			\begin{split}
				\Delta^{r+1} & = \frac{1}{S}\sum_{i\in\mathcal{S}^r}\Delta_i^r \\
				w^{r+1} & = w^{r} + \eta_g \Delta^r . \nonumber
			\end{split}
		\end{equation}
	\end{itemize}

	\begin{lemma}\label{lemma:deltadrift}
		(Bounded $\mathcal{E}_{\bdelta}$ of \texttt{FedSAM}). Suppose our functions satisfies Assumptions~\ref{ass:smooth}-\ref{ass:sigmag}. Then, the updates of \texttt{FedSAM} for any learning rate satisfying $\eta_l \leq \frac{1}{4KL}$ have the drift due to $\delta_{i,k} - \delta$:
		\begin{equation}
			\mathcal{E}_{\delta} = \frac{1}{N}\sum_{i}\E [\|\delta_{i,k} - \delta \|^2 ] \leq 2K^2 \beta^2 \eta_l^2 \rho^2 . \nonumber
		\end{equation}
	\end{lemma}

	\noindent\textit{Proof.} Recall the definitions of $\delta$ and $\delta_{i,k}$ as follows:
	\begin{equation}
		\delta = \rho \frac{\nabla F(w)}{\|\nabla F(w)\|}, ~~~ \delta_{i,k} = \rho \frac{\nabla F_i (w_{i,k} ,\xi_i )}{\|\nabla F_i (w_{i,k}, \xi_i )\|}. \nonumber
	\end{equation}
	If the local learning rate $\eta_l$ is small, the gradient of one epoch $\nabla F_i (w_{i,k}, \xi_i )$ is small. Based on the first order Hessian approximation, the expected gradient is
	\begin{equation}
		\nabla F_i (w_{i,k}) = \nabla F_i (w_{i,k-1} + g_{i,k-1}) = \nabla F_i (w_{i,k-1}) + H\eta_l g_{i,k-1} + O(\|\eta_l g_{i,k-1} \|^2 ),\nonumber
	\end{equation}
	where $H$ is the Hessian at $w_{i,k-1}$. Therefore, we have
	\begin{equation}\label{Eq:deltadrift}
		\E [\|\delta_{i,k} - \delta \|^2 ] = \rho^2 \E \bigg[\bigg\|\frac{\nabla F_i (w_{i,k})}{\|\nabla F_i (w_{i,k})\|} - \frac{\nabla F_i (w)}{\|\nabla F_i (w)\|} \bigg\|^2 \bigg] \leq \rho^2 \phi_{i,k}, 
	\end{equation}
	where $\phi_{i,k}$ is the square of the angle between the unit vector in the direction of $\nabla F_i (w_{i,k})$ and $\nabla F_i (w_{i,0})$. The inequality follows from that (1) $\bigg\|\frac{\nabla F_i (\cdot)}{\|\nabla F_i (\cdot)\|}\bigg\|^2 < 1$, and hence we replace $\delta$ with a unit vector in corresponding directions multiplied by $\rho^2$ and obtain the upper bound, (2) the norm of difference in unit vectors can be upper bounded by the square of the arc length on a unit circle. When the learning rate $\eta_l$ and the local model update of one epoch $\nabla F_i (w_{i,k})$ are small, $\phi_{i,k}$ is also small. Based on the first order Taylor series, i.e., $\tan x = x + O(x^2 )$, we have
	\begin{equation}
		\begin{split}
			& \tan \phi_{i,k} = \frac{\|\nabla F_i (w_{i,k}) - \nabla F_i (w_{i,0})\|^2}{\|\nabla F_i (w_{i,0})\|^2} + O(\phi_{i,k}^2 )\\
			& = \frac{\|\nabla F_{i}(w_{i,k-1}) - H\eta_l g_{i,k-1} - O(\|\eta_l g_{i,k-1}\|^2 ) - \nabla F_i (w_{i,0})\|^2}{\|\nabla F_i (w_{i,0})\|^2} + O(\phi_{i,k}^2 ) \\
			& \overset{\text{(a)}}{\leq} \bigg(1 + \frac{1}{K-1}\bigg)\frac{\|\nabla F_{i}(w_{i,k-1} )- \nabla F_i (w_{i,0})\|^2}{\|\nabla F_i (w_{i,0})\|^2} + \frac{K\|H\eta_l g_{i,k-1} + O(\|\eta_l g_{i,k-1}\|^2 )\|^2}{\|\nabla F_i (w_{i,0})\|^2}+ O(\phi_{i,k}^2 )\\
			& \overset{\text{(b)}}{\leq} \bigg(1 + \frac{1}{K-1} \bigg)\frac{\|\nabla F_{i}(w_{i,k-1} )- \nabla F_i (w_{i,0})\|}{\|\nabla F_i (w_{i,0})\|} + KL^2 \eta_l^2  ,\nonumber
		\end{split}
	\end{equation}
	where (a) is from Lemma~\ref{lemma:triangleinequality} with $a = \frac{1}{K-1}$ and (b) is due to maximum eigenvalue of $H$ is bounded by $L$ because $F$ function is $L$-smooth. Unrolling the recursion above, we have
	\begin{equation}\label{Eq:phiik}
		\frac{1}{N}\sum_{i\in [N]}\frac{\|\nabla F_i (w_{i,k}) - \nabla F_i (w_{i,0})\|^2}{\|\nabla F_i (w_{i,0})\|^2} + O(\phi_{i,k}^2 ) \leq \sum_{\tau = 1}^{k-1}\bigg(1 + \frac{1}{K-1}\bigg)^\tau KL^2 \eta_l^2 \leq 2K^2 L^2 \eta_l^2 .
	\end{equation}
	Plugging \eqref{Eq:phiik} into \eqref{Eq:deltadrift}, we have
	\begin{equation}\label{Eq:secondterm}
		\mathcal{E}_{\delta} = \frac{1}{N}\sum_{i\in[N]}\E [\|\delta_{i,k} - \delta \|^2 ]\leq 2K^2 L^2 \eta_l^2 \rho^2 .\nonumber
	\end{equation}

	This completes the proof. \hfill~$\Box$
	
	\begin{lemma}\label{lemma:xdrift}
		(Bounded $\mathcal{E}_{w}$ of \texttt{FedSAM}). Suppose our functions satisfies Assumptions~\ref{ass:smooth}-\ref{ass:sigmag}. Then, the updates of \texttt{FedSAM} for any learning rate satisfying $\eta_l \leq \frac{1}{10KL}$ have the drift due to $w_{i,k} - w$:
		\begin{equation}
			\mathcal{E}_{w} = \frac{1}{N}\sum_{i}\E [\|w_{i,k} - w \|^2 ] \leq 5K\eta_l^2 (2L^2 \rho^2 \sigma_l^2 + 6K(3\sigma_g^2 + 6L^2 \rho^2 ) + 6K\|\nabla f(\tw)\|^2 ) + 24K^3 \eta_l^4 L^4 \rho^2 . \nonumber
		\end{equation}
	\end{lemma}
	
	\noindent\textit{Proof.} Recall that the local update on client $i$ is $w_{i,k} = w_{i,k-1} - \eta_l \tg_{i,k-1}$. Then,
	\begin{equation}
		\begin{split}
			& \E \|w_{i,k} - w \|^2 = \E\|w_{i,k-1} - w - \eta_l \tg_{i,k-1} \|^2 \\
			& \overset{\text{(a)}}{\leq} \E\|w_{i,k-1} - w - \eta_l (\tg_{i,k-1} - \nabla f_i (\tw_{i,k-1}) + \nabla f_i (\tw_{i,k-1}) - \nabla f_i (\tw) + \nabla f_i (\tw)) - \nabla f(\tw) + \nabla f(\tw)\|^2 \\
			& \overset{\text{(b)}}{\leq} \bigg(1+\frac{1}{2K-1}\bigg) \E\|w_{i,k-1} - w\|^2 + \E\|\eta_l (\tg_{i,k-1} - \nabla f_i (\tw_{i,k-1}))\|^2 \\
			&+ 6K\E\|\eta_l (\nabla f_i (\tw_{i,k-1}) - \nabla f_i (\tw))\|^2 + 6K\E \|\eta_l (\nabla f_i (\tw) - \nabla f(\tw))\|^2 + 6K\|\eta_l \nabla f (\tw) \|^2 \\
			& \overset{\text{(c)}}{\leq} \bigg(1+\frac{1}{2K-1} + 2L^2 \eta_l^2 \bigg) \E\|w_{i,k-1} - w\|^2 + 2\eta_l^2 L^2 \rho^2 \sigma_l^2 + 12K\eta_l^2 L^2 \E\|w_{i,k-1} - w \|^2 \\
			& + 12K L^2 \eta_l^2 \E \|\delta_{i,k-1} - \delta \|^2 + 6K\eta_l^2 \E\|\nabla f_i (\tw) - \nabla f(\tw)\|^2 +6K\|\nabla f(\tw)\|^2 \\
			& \overset{\text{(d)}}{\leq} \bigg(1 + \frac{1}{2K-1} +12K\eta_l^2 L^2 + 2L^2 \eta_l^2 \bigg)\E\|w_{i,k-1} - w\|^2 + 2\eta_l^2 L^2 \rho^2 \sigma_l^2 + 12K L^2 \eta_l^2 \E \|\delta_{i,k} - \delta \|^2 \\
			& + 6K\eta_l^2 (3\sigma_g^2 + 6L^2 \rho^2) + 6K\|\nabla f(\tw)\|^2 ,\nonumber
		\end{split}
	\end{equation}
	where (a) follows from the fact that $\tg_{i,k-1}$ is an unbiased estimator of $\nabla f_i (\tw_{i,k-1})$ and Lemma~\ref{lemma:independent}; (b) is from Lemma~\ref{lemma:dependent}; (c) is from Assumption~\ref{ass:sigmal} and Lemma~\ref{lemma:dependent} and (d) is from Lemma~\ref{lemma:sigmag}.
	
	Averaging over the clients $i$ and learning rate satisfies $\eta_l \leq \frac{1}{10KL}$, we have
	\begin{equation}\label{Eq:firstterm}
		\begin{split}
			\frac{1}{N}\sum_{i\in[N]}&\E \|w_{i,k} - w\|^2 \leq \bigg(1 + \frac{1}{2K-1} +12K\eta_l^2 L^2 + 2L^2 \eta_l^2 \bigg)\frac{1}{N}\sum_{i\in [N]} \E\|w_{i,k-1} - w\|^2\\
			& + 2\eta_l^2 L^2 \rho^2 \sigma_l^2 +  12K L^2 \eta_l^2 \frac{1}{N}\sum_{i\in[N]} \E \|\delta_{i,k} - \delta \|^2 + 6K\eta_l^2 (3\sigma_g^2 + 6L^2 \rho^2 ) + 6K\|\nabla f(\tw)\|^2   \\
			& \overset{\text{(a)}}{\leq} \bigg(1+\frac{1}{K-1}\bigg)\frac{1}{N}\sum_{i\in [N]}\E\|w_{i,k-1} - w\|^2 + \eta_l^2 L^2 \rho^2 \sigma_l^2\\
			& + 12KL^2 \eta_l^2 \frac{1}{N}\sum_{i\in[N]} \E \|\delta_{i,k} - \delta \|^2 + 6K\eta_l^2 (3\sigma_g^2 +6L^2 \rho^2 ) +6K\|\nabla f(\tw)\|^2 \\
			& \leq \sum_{\tau = 0}^{k-1}\bigg(1+\frac{1}{K-1}\bigg)^\tau [2\eta_l^2 L^2 \rho^2 \sigma_l^2 + 6K\eta_l^2 (3\sigma_g^2 +6L^2 \rho^2 ) + 6K\|\nabla f(\tw)\|^2 ] + 12KL^2 \eta_l^2 \frac{1}{N}\sum_{i\in[N]} \E \|\delta_{i,k} - \delta \|^2 \\
			& \overset{\text{(b)}}{\leq} 5K\eta_l^2 (2L^2 \rho^2 \sigma_l^2 + 6K(3\sigma_g^2 + 6L^2 \rho^2 ) + 6K\|\nabla f(\tw)\|^2 ) + 24K^3 \eta_l^4 L^4 \rho^2 ,\nonumber
		\end{split}
	\end{equation}
	where (a) is due to the fact that $\eta_l \leq \frac{1}{10KL}$ and (b) is from Lemma~\ref{lemma:deltadrift}. \hfill~$\Box$
	
	\subsection{Convergence Analysis of Full client participation \texttt{FedSAM}}
	\begin{lemma}\label{lemma:A1}
		\begin{equation}
			\langle \nabla f(\tw^r ),\E_r [\Delta^r + \eta_l K\nabla f(\tw^r ) ]\rangle \leq \frac{\eta_l K}{2}\|\nabla f(\tw^r ))\|^2 + K \eta_l L^2 \mathcal{E}_{w} + K \eta_l L^2 \mathcal{E}_{\delta} - \frac{\eta_l}{2KN^2}\E_r \bigg\|\sum_{i,k} \nabla f_i (\tw_{i,k}) \bigg\|^2 .\nonumber
		\end{equation}
	\end{lemma}
	
	\noindent\textit{Proof.} 
	\begin{equation}
		\begin{split}
			& \langle \nabla f(\tw^r ) ,\E_r [\Delta^r + \eta_l K\nabla f(\tw^r ) ]\rangle \\
			& \overset{\text{(a)}}{=} \frac{\eta_l K}{2}\|\nabla f(\tw^r ))\|^2 + \frac{\eta_l}{2KN^2} \E_r \bigg\|\sum_{i,k}\nabla f_i (\tw_{i,k}^r ) - \nabla f_i (\tw^r )\bigg\|^2 - \frac{\eta_l}{2KN^2}\E_r \bigg\|\sum_{i,k}\nabla f_i (\tw_{i,k}^r ) \bigg\|^2\\
			& \overset{\text{(b)}}{\leq} \frac{\eta_l K}{2}\|\nabla f(\tw^r ))\|^2 + \frac{\eta_l}{2N}\sum_{i,k}\E_r \|\nabla f_i (\tw_{i,k}^r ) - \nabla f_i (\tw^r )\|^2 - \frac{\eta_l}{2KN^2}\E_r \bigg\|\sum_{i,k}\nabla f_i (\tw_{i,k}^r ) \bigg\|^2 \\
			& \overset{\text{(c)}}{\leq} \frac{\eta_l K}{2}\|\nabla f(\tw^r ))\|^2 + \frac{\eta_l \beta^2}{2N} \sum_{i,k}\E_r \|\tw_{i,k}^r - \tw^r \|^2 - \frac{\eta_l}{2KN^2}\E_r \bigg\|\sum_{i,k} \nabla f_i (\tw_{i,k}^r ) \bigg\|^2 \\
			& \overset{\text{(d)}}{\leq}  \frac{\eta_l K}{2}\|\nabla f(\tw^r ))\|^2 + \frac{\eta_l L^2}{N} \sum_{i,k}\E_r \|w_{i,k}^r - w^r \|^2 + \frac{\eta_l L^2}{N} \sum_{i,k}\E_r \|\delta_{i,k}^r - \delta^r \|^2 - \frac{\eta_l}{2KN^2}\E_r \bigg\|\sum_{i,k} \nabla f_i (\tw_{i,k}^r ) \bigg\|^2 \\
			& = \frac{\eta_l K}{2}\|\nabla f(\tw^r ))\|^2 + K \eta_l L^2 \mathcal{E}_{w} + K \eta_l L^2 \mathcal{E}_{\delta} - \frac{\eta_l}{2KN^2}\E_r \bigg\|\sum_{i,k} \nabla f_i (\tw_{i,k}^r ) \bigg\|^2 ,
		\end{split}
	\end{equation}
	where (a) is from that $\langle a , b \rangle = \frac{1}{2}(\|a\|^2 + \|b\|^2 - \|a - b\|^2 )$ with $a = \sqrt{\eta_l K}\nabla f(\tw^r )$ and $b = -\frac{\sqrt{\eta_l}}{N\sqrt{K}}\sum_{i,k}(\nabla f_i (\tw_{i,k}^r) - \nabla f_i (\tw^r ))$; (b) is from Lemma~\ref{lemma:dependent}; (c) is from Assumption~\ref{ass:smooth} and (d) is from Lemma~\ref{lemma:dependent}.
	
	\begin{lemma}\label{lemma:d2}
		For the full client participation scheme, we can bound $\E [\|\Delta^r \|^2 ]$ as follows:
		\begin{equation}
			\E_r [\|\Delta^r \|^2 ] \leq \frac{K\eta_l^2 L^2 \rho^2}{N}\sigma_l^2 + \frac{\eta_l^2}{N^2}\bigg[\bigg\|\sum_{i,k}\nabla f_i(\tw_{i,k}^r ) \bigg\|^2 \bigg]. \nonumber
		\end{equation}
	\end{lemma}
	
	\noindent\textit{Proof.} For the full client participation scheme, we have:
	\begin{equation}
		\begin{split}
			\E_r [\|\Delta^r \|^2 ] &\overset{\text{(a)}}{\leq} \frac{\eta_l^2}{N^2}\E_r \bigg[\bigg\|\sum_{i,k}\tg_{i,k}^r \bigg\|^2 \bigg] \overset{\text{(b)}}{=} \frac{\eta_l^2}{N^2} \E_r \bigg[\bigg\|\sum_{i,k}(\tg_{i,k}^r - \nabla f_i (\tw_{i,k}^r )) \bigg\|^2 \bigg] + \frac{\eta_l^2}{N^2}\E_r \bigg[\bigg\|\sum_{i,k}\nabla f_i(\tw_{i,k}^r ) \bigg\|^2 \bigg]\\
			& \overset{\text{(c)}}{\leq} \frac{K\eta_l^2 L^2 \rho^2}{N}\sigma_l^2 + \frac{\eta_l^2}{N^2}\bigg[\bigg\|\sum_{i,k}\nabla f_i(\tw_{i,k}^r ) \bigg\|^2 \bigg],\nonumber
		\end{split}
	\end{equation}
	where (a) is from Lemma~\ref{lemma:dependent}; (b) is from Lemma~\ref{lemma:independent} and (c) is from Lemma~\ref{lemma:SAMvariance}. \hfill~$\Box$

	\begin{lemma}\label{lemma:decentsam}
		(Descent Lemma). For all $r \in R-1$ and $i \in \S^r$, with the choice of learning rate , the iterates generated by \texttt{FedSAM} in Algorithm~\ref{Alg:FedSAM} satisfy:
		\begin{equation}
			\begin{split}
				\E_r [f(\tw^{r+1})] & \leq f(\tw^r ) - K\eta_g \eta_l \bigg(\frac{1}{2} - 30K^2 L^2 \eta_l^2 \bigg)\|\nabla f(\tw^r )\|^2 + K\eta_g \eta_l (10KL^4 \eta_l^2 \rho^2 \sigma_l^2 + 90K^2 L^2 \eta_l^2 \sigma_g^2 + 180 K^2 L^4 \eta_l^2 \rho^2 \\
				&+ 120 K^4 L^6 \eta_l^6 \rho^2 + 16K^3 \eta_l^4 L^6 \rho^2 + \frac{\eta_g \eta_l L^3 \rho^2}{N}\sigma_l^2 )  \nonumber
			\end{split}
		\end{equation}
		where the expectation is w.r.t. the stochasticity of the algorithm.
	\end{lemma}
	
	\noindent\textit{Proof.} We firstly propose the proof of full client participation scheme. Due to the smoothness in Assumption~\ref{ass:smooth}, taking expectation of $f(\tw^{r+1})$ over the randomness at communication round $r$, we have:
	\begin{equation}\label{Eq:fedsam}
		\begin{split}
			& \E_{r} [F(w^{r+1})] = \E_r [f(\tw^{r+1} )] \leq f(\tw^r ) + \E_r \langle \nabla f(\tw^r ), \tw^{r+1} - \tw^r ]\rangle +\frac{L}{2}\E_r [\|\tw^{r+1} - \tw^r \|^2 ]\\
			& \overset{\text{(a)}}{=} f(\tw^r ) + \E_r \langle \nabla f(\tw^r ) ,  -\Delta^r + K\eta_g \eta_l \nabla f(\tw^r ) - K\eta_g \eta_l \nabla f(\tw^r )\rangle + \frac{L}{2}\eta_g^2 \E_r [\|\Delta^r \|^2]\\
			& \overset{\text{(b)}}{=} f(\tw^r ) - K\eta_g \eta_l \|\nabla f(\tw^r )\|^2 + \eta_g \langle \nabla f(\tw^r ) ,\E_r [-\Delta^r + K\eta_l \nabla f(\tw^r ) ]\rangle + \frac{L}{2}\eta_g^2 \E_r [\|\Delta^r \|^2 ]\\
			& \overset{\text{(c)}}{\leq} f(\tw^r ) - \frac{K\eta_g \eta_l}{2} \|\nabla f(\tw^r )\|^2 + K\eta_g \eta_l L^2 \mathcal{E}_{w} + K\eta_g \eta_l L^2 \mathcal{E}_{\delta} - \frac{\eta_g \eta_l}{2KN}\E_r \bigg[\bigg\|\sum_{i,k}\nabla f_i (\tw_{i,k}^r )\bigg\|^2 \bigg] + \frac{L}{2}\eta_g^2 \E_r [\|\Delta^r \|^2 ]\\
			& \overset{\text{(d)}}{\leq} f(\tw^r ) - \frac{K\eta_g \eta_l}{2} \|\nabla f(\tw^r )\|^2 + K\eta_g \eta_l L^2 \mathcal{E}_{w} + K\eta_g \eta_l L^2 \mathcal{E}_{\delta} + \frac{K\eta_g^2 \eta_l^2 L^3 \rho^2}{N}\sigma_l^2 \\
			& \overset{\text{(e)}}{\leq} f(\tw^r ) - K\eta_g \eta_l \bigg(\frac{1}{2} - 30K^2 L^2 \eta_l^2 \bigg)\|\nabla f(\tw^r )\|^2 + K\eta_g \eta_l (10KL^4 \eta_l^2 \rho^2 \sigma_l^2 + 90K^2 L^2 \eta_l^2 \sigma_g^2 + 180 K^2 L^4 \eta_l^2 \rho^2 \\
			&+ 120 K^4 L^6 \eta_l^6 \rho^2 + 16K^3 \eta_l^4 L^6 \rho^2 + \frac{\eta_g \eta_l L^3 \rho^2}{N}\sigma_l^2 ) ,\nonumber
		\end{split}
	\end{equation}
	where (a) is from the iterate update given in Algorithm~\ref{Alg:FedSAM}; (b) results from the unbiased estimators; (c) is from Lemma~\ref{lemma:A1}; (d) is from Lemma~\ref{lemma:d2} and due to the fact that $\eta_g \eta_l \leq \frac{1}{KL}$ and (e) is from Lemmas~\ref{lemma:deltadrift} and \ref{lemma:xdrift}. \hfill~$\Box$

	\begin{theorem}
		Let constant local and global learning rates $\eta_l$ and $\eta_g$ be chosen as such that $\eta_l \leq \frac{1}{10KL}$, $\eta_g \eta_l \leq \frac{1}{KL}$. Under Assumption~\ref{ass:smooth}-\ref{ass:sigmag} and with full client participation, the sequence of outputs $\{w^r \}$ generated by \texttt{FedSAM} satisfies:
		\begin{equation}
			\min_{r\in [R]}\E\|\nabla F(w^r )\|^2 \leq \frac{F^0 - F^*}{CK\eta_g \eta_l} + \Phi ,\nonumber
		\end{equation}
		where $\Phi = \frac{1}{C}[10KL^4 \eta_l^2 \rho^2 \sigma_l^2 + 90K^2 L^2 \eta_l^2 \sigma_g^2 + 180 K^2 L^4 \eta_l^2 \rho^2 
		+ 120 K^4 L^6 \eta_l^6 \rho^2 + 16K^3 \eta_l^4 L^6 \rho^2 + \frac{\eta_g \eta_l L^3 \rho^2}{N}\sigma_l^2 ]$. If we choose the learning rates $\eta_l = \frac{1}{\sqrt{R}KL}$, $\eta_g = \sqrt{KN}$ and perturbation amplitude $\rho$ proportional to the learning rate, e.g., $\rho = \frac{1}{\sqrt{R}}$, we have 
		\begin{equation}
			\frac{1}{R}\sum_{r=1}^{R}\E [\|F(w^{r+1})\|] = \mathcal{O}\bigg(\frac{FL}{\sqrt{RKN}} + \frac{\sigma_g^2}{R} + \frac{L^2 \sigma^2}{R^{3/2}\sqrt{KN}} + \frac{L^2}{R^{3/2}} \bigg) .\nonumber
		\end{equation}
	\end{theorem}
	
	\noindent\textit{Proof.} For full client participation, summing the result of Lemma~\ref{lemma:decentsam} for $r = [R]$ and multiplying both sides by $\frac{1}{CK\eta_g \eta_l R}$ with $(\frac{1}{2} - 30K^2 L^2 \eta_l^2 ) >C>0$ if $\eta_l < \frac{1}{\sqrt{30}KL}$, we have
	\begin{equation}
		\begin{split}
			&\frac{1}{R}\sum_{r=1}^{R}\E [\|F(w^{r+1})\|^2 ] = \frac{1}{R}\sum_{r=1}^{R}\E [\|f(\tw^{r+1})\|^2 ] \\
			& \leq \frac{f(\tw^r) - f(\tw^{r+1})}{CK\eta_g \eta_l R} + \frac{1}{C} (10KL^4 \eta_l^2 \rho^2 \sigma_l^2 + 90K^2 L^2 \eta_l^2 \sigma_g^2 + 180 K^2 L^4 \eta_l^2 \rho^2 + 120 K^4 L^6 \eta_l^6 \rho^2 + 16K^3 \eta_l^4 L^6 \rho^2 + \frac{\eta_g \eta_l L^3 \rho^2}{N}\sigma_l^2 )\\
			& \leq \frac{f(\tw^0 ) - f^*}{CK\eta_g \eta_l R} + \frac{1}{C}(10KL^4 \eta_l^2 \rho^2 \sigma_l^2 + 90K^2 L^2 \eta_l^2 \sigma_g^2 + 180 K^2 L^4 \eta_l^2 \rho^2 + 120 K^4 L^6 \eta_l^6 \rho^2 + 16K^3 \eta_l^4 L^6 \rho^2 + \frac{\eta_g \eta_l L^3 \rho^2}{N}\sigma_l^2 ),\nonumber
		\end{split}
	\end{equation}
	where the second inequality uses $f(\tw^{r+1} ) \geq f^* $ and $f(\tw^0 ) \geq f(\tw^r )$. If we choose the learning rates $\eta_l = \frac{1}{\sqrt{R}KL}$, $\eta_g = \sqrt{KN}$ and perturbation amplitude $\rho$ proportional to the learning rate, e.g., $\rho = \frac{1}{\sqrt{R}}$, we have
	\begin{equation}
		\frac{1}{R}\sum_{r=1}^{R}\E [\|F(w^{r+1})\|] = \mathcal{O}\bigg(\frac{FL}{\sqrt{RKN}} + \frac{\sigma_g^2}{R} + \frac{L^2 \sigma_l^2}{R^2 K} + \frac{L^2 \sigma_l^2}{R^{3/2}\sqrt{KN}} + \frac{L^2}{R^2 K} + \frac{L^2}{R^{3/2}} + \frac{L^2}{R^3 K}\bigg).\nonumber
	\end{equation}
	Note that the term $\frac{G^2}{R}$ is due to the heterogeneity between each client, $(\frac{L^2}{R^2 K} + \frac{L^2}{R^{3/2}\sqrt{KN}})\sigma^2$ is due to the local SGD and $\frac{1}{R^{3/2}} + \frac{1}{R^3 K}$ is due to the local SAM. We can see that $\frac{L^2}{R^{3/2}} + \frac{L^2}{R^3 K}$ only obtains higher order, and hence SAM part does not take large influence of convergence. After omitting the higher order, we have
	\begin{equation}
		\frac{1}{R}\sum_{r=1}^{R}\E [\|F(w^{r+1})\|] = \mathcal{O}\bigg(\frac{FL}{\sqrt{RKN}} + \frac{\sigma_g^2}{R} + \frac{L^2 \sigma^2}{R^{3/2}\sqrt{KN}} + \frac{L^2}{R^{3/2}} \bigg) .\nonumber
	\end{equation}
	
	This completes the proof. \hfill~$\Box$
	
	\subsection{Convergence Analysis of Partial Client Participation \texttt{FedSAM}}
	\begin{lemma}
		For the partial client participation, we can bound $\E_r [\|\Delta^r \|^2 ]$:
		\begin{equation}
			\E_r [\|\Delta^r \|^2 ] \leq \frac{K\eta_l^2 L^2 \rho^2}{S}\sigma_l^2 + \frac{S}{N}\sum_i \bigg\|\sum_{j=1}^{K-1}\nabla f_i (\tw_{i,k}^r )\bigg\|^2 + \frac{S(S-1)}{N^2}\bigg\|\sum_{j=0}^{K-1}\nabla f_i (\tw_{i,j}^r )\bigg\|^2 . \nonumber
		\end{equation}
	\end{lemma}
	For the partial client participation scheme w/o replacement, we have:
	\begin{equation}
		\begin{split}
			\E_r [\|\Delta^r \|^2 ] & \overset{\text{(a)}}{\leq} \frac{\eta_l^2}{S^2}\E_r \bigg[\bigg\|\sum_{i\in\S^r}\sum_{k}\tg_{i,k} \bigg\|^2 \bigg] = \frac{\eta_l^2}{S^2}\E_r \bigg[\bigg\|\sum_{i} \mathbb{I}\{i\in\S^r \}\sum_{k}\tg_{i,k} \bigg\|^2 \bigg]\\
			& \overset{\text{(b)}}{=} \frac{\eta_l^2}{SN} \E_r \bigg[\bigg\|\sum_{i}\sum_{j=0}^{K-1}(\tg_{i,j}^r - \nabla f_i (\tw_{i,j}^r )) \bigg\|^2 \bigg] + \frac{\eta_l^2}{S^2}\E_r \bigg[\bigg\|\sum_{i}\mathbb{I}\{i\in\S^r \}\sum_{j=0}^{K-1} \nabla f_i(\tw_{i,j}^r ) \bigg\|^2 \bigg]\\
			& \overset{\text{(c)}}{\leq} \frac{K\eta_l^2 L^2 \rho^2}{S}\sigma_l^2 + \frac{\eta_l^2}{S^2}\E_r \bigg[\bigg\|\sum_{i=1}^{S}\sum_{j=0}^{K-1} \nabla f_i(\tw_{i,j}^r ) \bigg\|^2 \bigg]\\
			& = \frac{K\eta_l^2 L^2 \rho^2}{S}\sigma_l^2 + \frac{\eta_l^2}{NS}\sum_i \bigg\|\sum_{j=1}^{K-1}\nabla f_i (\tw_{i,k}^r )\bigg\|^2 + \frac{(S-1)\eta_l^2}{SN^2}\bigg\|\sum_{j=0}^{K-1}\nabla f_i (\tw_{i,j}^r )\bigg\|^2 ,\nonumber
		\end{split}
	\end{equation}
	where (a) is from Lemma~\ref{lemma:dependent}; (b) is from Lemma~\ref{lemma:independent} and (c) is from Lemma~\ref{lemma:SAMvariance}.\hfill~$\Box$

	\begin{lemma}\label{lemma:tt}
		For $\E [\|\sum_{k}\nabla f_i (\tw_{i,k})\|^2 ]$, where $\nabla f_i (\tw_{i,k})^2$ for all $k \in [K]$ and $i\in [N]$ is chosen according to \texttt{FedSAM}, we have:
		\begin{equation}
			\begin{split}
				\sum_i \E \bigg[\bigg\|\sum_{k}\nabla f_i (\tw_{i,k}) \bigg\|^2 \bigg] & \leq 30NK^2 L^2 \eta_l^2 (2L^2 \rho^2 \sigma_l^2 + 6K (3\sigma^2_g + 6L^2 \rho^2 ) + 6K\|\nabla f(\tw)\|^2 ) + 144K^4 L^6 \eta_l^4 \rho^2 \\
				& + 12NK^4 L^2 \eta_l^2 \rho^2 + 3NK^2 (3\sigma_g^2 + 6L^2 \rho^2 ) + 3NK^2 \|\nabla f(\tw)\|^2,\nonumber
			\end{split}
		\end{equation}
		where the expectation is w.r.t the stochasticity of the algorithm.
	\end{lemma}

	\noindent\textit{Proof.} 
	\begin{equation}
		\begin{split}
			& \sum_i \E \bigg[\bigg\|\sum_{k}\nabla f_i (\tw_{i,k}) \bigg\|^2 \bigg] = \sum_i \E \bigg[\bigg\|\sum_k \nabla f_i (\tw_{i,k}) - \nabla f_i (\tw) + \nabla f_i (\tw)- \nabla f(\tw) + \nabla f(\tw) \bigg\|^2  \bigg]\\
			& \overset{\text{(a)}}{\leq}6KL^2 \sum_{i,k} \E [\|w_{i,k} - w \|^2 ] + 6KL^2 \sum_{i,k} \E [\|\delta_{i,k} - \delta \|^2 ] + 3NK^2 (3\sigma_g^2 + 6L^2 \rho^2 ) + 3NK^2 \|\nabla f(\tw)\|^2 \\
			& \overset{\text{(b)}}{\leq} 30NK^2 L^2 \eta_l^2 (2L^2 \rho^2 \sigma_l^2 + 6K (3\sigma^2_g + 6L^2 \rho^2 ) + 6K\|\nabla f(\tw)\|^2 ) + 144K^4 L^6 \eta_l^4 \rho^2 \\
			& + 12NK^4 L^2 \eta_l^2 \rho^2 + 3NK^2 (3\sigma_g^2 + 6L^2 \rho^2 ) + 3NK^2 \|\nabla f(\tw)\|^2. \nonumber
		\end{split}
	\end{equation}
	where (a) is from Assumption~\ref{ass:smooth}, Lemmas~\ref{lemma:dependent} and \ref{lemma:sigmag}; (b) is from Lemmas~\ref{lemma:deltadrift} and \ref{lemma:xdrift}. \hfill~$\Box$
	
	\begin{theorem}
		Let constant local and global learning rates $\eta_l$ and $\eta_g$ be chosen as such that $\eta_l \leq \frac{1}{10KL}$, $\eta_g \eta_l \leq \frac{1}{KL}$ and the condition $(\frac{1}{2} - 30K^2 L^2 \eta_l^2 - \frac{L\eta_g \eta_l}{2S}(3K + 180K^3 L^2 \eta_l^2 ))>0$ holds. Under Assumption~\ref{ass:smooth}-\ref{ass:sigmal} and with partial client participation, the sequence of outputs $\{w^r \}$ generated by \texttt{FedSAM} satisfies:
		\begin{equation}
			\min_{r\in [R]}\E\|\nabla F(w^r )\|^2 \leq \frac{F^0 - F^*}{CK\eta_g \eta_l} + \Phi ,\nonumber
		\end{equation}
		where $\Phi = \frac{1}{C}[10KL^4 \eta_l^2 \rho^2 \sigma_l^2 + 90K^2 L^2 \eta_l^2 \sigma_g^2 + 180 K^2 L^4 \eta_l^2 \rho^2 
		+ 120 K^4 L^6 \eta_l^6 \rho^2 + 16K^3 \eta_l^4 L^6 \rho^2 +\frac{L^3 \eta_g \eta_l \rho^2}{2S}\sigma^2 
		+ \frac{\eta_g \eta_l}{S}(30KL^5 \eta_l^2 \rho^2 \sigma_l^2 + 180K^2 L^3 \eta_l^2 \sigma_g^2 + 360KL^5 \eta_l^2 \rho^2 + 72K^3 L^7 \eta_l^4 \rho^2 + 6K^3 L^3 \eta_l^2 \rho^2 + 6KL\sigma^2_g + 6KL^3 \rho^2 )]$. If we choose the learning rates $\eta_l = \frac{1}{\sqrt{R}KL}$, $\eta_g = \sqrt{KS}$ and perturbation amplitude $\rho$ proportional to the learning rate, e.g., $\rho = \frac{1}{\sqrt{R}}$, we have:
		\begin{equation}
			\frac{1}{R}\sum_{r=1}^{R}\E [\|F(w^{r+1})\|] = \mathcal{O}\bigg(\frac{FL}{\sqrt{RKS}}+ \frac{\sqrt{K}G^2}{\sqrt{RS}} + \frac{L^2 \sigma^2}{R^{3/2}K} + \frac{\sqrt{K}L^2}{R^{3/2} \sqrt{S}}\bigg).\nonumber
		\end{equation}
	\end{theorem}
	
	\begin{equation}
		\begin{split}
			& \E[\|f(\tw^{r+1})\|] \\
			& \overset{\text{(a)}}{\leq} f(\tw^r ) - \frac{K\eta_g \eta_l}{2} \|\nabla f(\tw^r )\|^2 + K\eta_g \eta_l L^2 \mathcal{E}_{w} + K\eta_g \eta_l L^2 \mathcal{E}_{\delta} - \frac{\eta_g \eta_l}{2KN}\E_r \bigg[\bigg\|\sum_{i,k}\nabla f_i (\tw_{i,k}^r )\bigg\|^2 \bigg] + \frac{L}{2}\eta_g^2 \E_r [\|\Delta^r \|^2 ]\\
			& \overset{\text{(b)}}{\leq} f(\tw^r ) - \frac{K\eta_g \eta_l}{2} \|\nabla f(\tw^r )\|^2 + K\eta_g \eta_l L^2 \mathcal{E}_{w} + K\eta_g \eta_l L^2 \mathcal{E}_{\delta} + \frac{K\eta_g^2 \eta_l^2 L^3 \rho^2}{2S}\sigma_l^2 \\
			& - \frac{\eta_g \eta_l}{2KN}\E_r \bigg[\bigg\|\sum_{i,k}\nabla f_i (\tw_{i,k}^r )\bigg\|^2 \bigg] +  \frac{\eta_g^2 LS}{2N}\sum_i \bigg\|\sum_{j=1}^{K-1}\nabla f_i (\tw_{i,k}^r )\bigg\|^2 + \frac{\eta_g^2 LS(S-1)}{2N^2}\bigg\|\sum_{j=0}^{K-1}\nabla f_i (\tw_{i,j}^r )\bigg\|^2 \\
			& \overset{\text{(c)}}{\leq} f(\tw^r ) - \frac{K\eta_g \eta_l}{2} \|\nabla f(\tw^r )\|^2 + K\eta_g \eta_l L^2 \mathcal{E}_{w} + K\eta_g \eta_l L^2 \mathcal{E}_{\delta} + \frac{K\eta_g^2 \eta_l^2 L^3 \rho^2}{2S}\sigma_l^2 + \frac{L \eta_g^2 \eta_l^2}{2NS}\sum_{i} \|\sum_k \nabla f_i (\tw_{i,k}^r )\|^2 \\
			& \overset{\text{(d)}}{\leq} f(\tw^r ) - K\eta_g \eta_l \bigg(\frac{1}{2} - 30K^2 L^2 \eta_l^2 - \frac{L\eta_g \eta_l}{2S}(3K + 180K^3 L^2 \eta_l^2 )\bigg) \|\nabla f(\tw^r )\|^2 \\
			& + K\eta_g \eta_l \bigg(10KL^4 \eta_l^2 \rho^2 \sigma_l^2 + 90K^2 L^2 \eta_l^2 \sigma_g^2 + 180 K^2 L^4 \eta_l^2 \rho^2 
			+ 120 K^4 L^6 \eta_l^6 \rho^2 + 16K^3 \eta_l^4 L^6 \rho^2 +\frac{L^3 \eta_g \eta_l \rho^2}{2S}\sigma^2 \bigg)\\
			& + \frac{K\eta_g^2 \eta_l^2}{S} (30KL^5 \eta_l^2 \rho^2 \sigma_l^2 + 180K^2 L^3 \eta_l^2 \sigma_g^2 + 360KL^5 \eta_l^2 \rho^2 + 72K^3 L^7 \eta_l^4 \rho^2 + 6K^3 L^3 \eta_l^2 \rho^2 + 6KL\sigma^2_g + 6KL^3 \rho^2 )\\
			& \overset{\text{(e)}}{\leq} f(\tw^r ) - CK\eta_g \eta_l \|\nabla f(\tw^r )\|^2 \\
			& + K\eta_g \eta_l \bigg(10KL^4 \eta_l^2 \rho^2 \sigma_l^2 + 90K^2 L^2 \eta_l^2 \sigma_g^2 + 180 K^2 L^4 \eta_l^2 \rho^2 
			+ 120 K^4 L^6 \eta_l^6 \rho^2 + 16K^3 \eta_l^4 L^6 \rho^2 +\frac{L^3 \eta_g \eta_l \rho^2}{2S}\sigma^2 \bigg)\\
			& + \frac{K\eta_g^2 \eta_l^2}{S} (30KL^5 \eta_l^2 \rho^2 \sigma_l^2 + 180K^2 L^3 \eta_l^2 \sigma_g^2 + 360KL^5 \eta_l^2 \rho^2 + 72K^3 L^7 \eta_l^4 \rho^2 + 6K^3 L^3 \eta_l^2 \rho^2 + 6KL\sigma^2_g + 6KL^3 \rho^2 ),\nonumber
		\end{split}
	\end{equation}
	where (a) is from Lemma~\ref{lemma:decentsam}; (b) is from \ref{lemma:d2}; (c) is based on taking the expectation of $r$-th round and if the learning rates satisfy that $KL\eta_g \eta_l \leq \frac{S-1}{S}$; (d) is from Lemmas~\ref{lemma:deltadrift}, \ref{lemma:xdrift} and \ref{lemma:tt} and (e) holds because there exists a constant $C > 0$ satisfying $(\frac{1}{2} - 30K^2 L^2 \eta_l^2 - \frac{L\eta_g \eta_l}{2S}(3K + 180K^3 L^2 \eta_l^2 ))>C>0$.
	
	Summing the above result for $r = [R]$ and multiplying both sides by $\frac{1}{CK\eta_g \eta_l R}$, we have
	\begin{equation}
		\begin{split}
			& \frac{1}{R}\sum_{r=1}^{R}\E [\|F(w^{r+1})\|] \leq \frac{f(\tw^r ) - f(\tw^{r+1})}{CK\eta_g \eta_l R}\\
			& + \frac{1}{C}\bigg(10KL^4 \eta_l^2 \rho^2 \sigma_l^2 + 90K^2 L^2 \eta_l^2 \sigma_g^2 + 180 K^2 L^4 \eta_l^2 \rho^2 
			+ 120 K^4 L^6 \eta_l^6 \rho^2 + 16K^3 \eta_l^4 L^6 \rho^2 +\frac{L^3 \eta_g \eta_l \rho^2}{2S}\sigma^2 \\
			&+ \frac{\eta_g \eta_l}{S}(30KL^5 \eta_l^2 \rho^2 \sigma_l^2 + 180K^2 L^3 \eta_l^2 \sigma_g^2 + 360KL^5 \eta_l^2 \rho^2 + 72K^3 L^7 \eta_l^4 \rho^2 + 6K^3 L^3 \eta_l^2 \rho^2 + 6KL\sigma^2_g + 6KL^3 \rho^2 )\bigg)\\
			& \leq \frac{F}{CK\eta_g \eta_l R} \\
			& + \frac{1}{C}\bigg(10KL^4 \eta_l^2 \rho^2 \sigma_l^2 + 90K^2 L^2 \eta_l^2 \sigma_g^2 + 180 K^2 L^4 \eta_l^2 \rho^2 
			+ 120 K^4 L^6 \eta_l^6 \rho^2 + 16K^3 \eta_l^4 L^6 \rho^2 +\frac{L^3 \eta_g \eta_l \rho^2}{2S}\sigma^2 \\
			&+ \frac{\eta_g \eta_l}{S}(30KL^5 \eta_l^2 \rho^2 \sigma_l^2 + 180K^2 L^3 \eta_l^2 \sigma_g^2 + 360KL^5 \eta_l^2 \rho^2 + 72K^3 L^7 \eta_l^4 \rho^2 + 6K^3 L^3 \eta_l^2 \rho^2 + 6KL\sigma^2_g + 6KL^3 \rho^2 )\bigg), \nonumber
		\end{split}
	\end{equation}
	where the second inequality uses $F= f(\tw^0)-f^* \leq f(\tw^r )-f(\tw^{r+1})$. If we choose the learning rates $\eta_l = \frac{1}{\sqrt{R}KL}$, $\eta_g = \sqrt{KS}$ and perturbation amplitude $\rho$ proportional to the learning rate, e.g., $\rho = \frac{1}{\sqrt{R}}$, we have:
	\begin{equation}
		\begin{split}
			\frac{1}{R}\sum_{r=1}^{R}\E [\|F(w^{r+1})\|] & = \mathcal{O}\bigg(\frac{FL}{\sqrt{RKS}}+ \frac{\sigma_g^2}{R} + \frac{\sqrt{K}\sigma_g^2}{\sqrt{RS}} +\frac{\sqrt{KS}\sigma_g^2}{R^{3/2}} + \frac{L^2 \sigma_l^2}{R^{3/2}K} + \frac{L^2 \sigma_l^2}{R^{3/2}\sqrt{KS}}\\
			& + \frac{L^2 \sigma_l^2}{R^{5/2}\sqrt{KS}} + \frac{L^2}{R^2} + \frac{1}{R^4 K^2}+ \frac{L^2}{R^3 K} + \frac{\sqrt{KS}}{R^{5/2} SK^2} + \frac{\sqrt{KS}}{R^{7/2}S K^2} + \frac{\sqrt{K}}{R^{5/2}\sqrt{S}} + \frac{\sqrt{K}L^2}{R^{3/2}\sqrt{S}}\bigg),\nonumber
		\end{split}
	\end{equation}
	If the number of sampling clients are larger than the number of epochs, i.e., $S \geq K$, and omitting the larger order of each part, we have:
	\begin{equation}
		\frac{1}{R}\sum_{r=1}^{R}\E [\|F(w^{r+1})\|] = \mathcal{O}\bigg(\frac{FL}{\sqrt{RKS}}+ \frac{\sqrt{K}G^2}{\sqrt{RS}} + \frac{L^2 \sigma^2}{R^{3/2}K} + \frac{\sqrt{K}L^2}{R^{3/2} \sqrt{S}}\bigg).\nonumber
	\end{equation}
	This completes the proof. \hfill~$\Box$

	\section{Generalization Bounds}\label{Sec:generalizationbound}
	The generalization bound of \texttt{FedSAM} follows the margin-based generalization bounds in \cite{neyshabur2018pac, bartlett2017spectrally, farnia2018generalizable}. We consider the margin-based error for analyzing the generalization error in \texttt{FedSAM} with general neural network as follows:
	\begin{equation}\label{Eq:margin}
		L_\gamma^{\text{SAM}} (F(w)) := \frac{1}{N}\sum_{i=1}^{N}\mathbb{P}_i \bigg(f_i (w+\delta_i ,\mathbf{X}) [Y] - \max_{j \neq Y}F_i (w+\delta_i ,\mathbf{X})[j] \leq \gamma \bigg).
	\end{equation}

	Our generalization bound is based on the two following Lemmas in \cite{chatterji2019intriguing} and \cite{neyshabur2018pac}:
	\begin{lemma}\label{lemma:PAC}
		(\cite{chatterji2019intriguing}). Let $F(w)$ be any predictor function with parameters $w$ and $\mathcal{P}$ be a prior distribution on parameters $w$. Then, for any $\gamma, m, \zeta>0$, with probability $1-\zeta$ over training set $\mathcal{M}$ of size $m$, for any parameter $w$ and any perturbation distribution $\mathcal{Q}$ over parameters such that $\mathbb{P}_{\delta \sim \mathcal{Q}}[\max_{\mathbf{X}}|F(w+\delta)- F(w)| \leq \frac{\gamma}{4}]\geq \frac{1}{2}$, we have:
		\begin{equation}
			L^{\text{SAM}} (F(w)) \leq \hat{L}^{\text{SAM}}_\gamma (F(w)) + \sqrt{\frac{2\text{KL}(w + \delta\|\mathcal{P})+\log \frac{m}{\zeta}}{2(m-1)}}.\nonumber
		\end{equation}
		where $\text{KL}(\cdot \| P)$ is the KL-divergence.
	\end{lemma}

	\begin{lemma}\label{lemma:output}
		(\cite{neyshabur2018pac}). Let norm of input $\mathbf{X}$ be bounded by $A$. For any $A >0$, let $F(w)$ be a neural network with ReLU activations and depth $d$ with $h$ units per hidden-layer. Then for any $w , \mathbf{X} \in\mathcal{X}$, and any perturbation $\bdelta$ s.t. $\|\delta_j \| \leq \|W_j \|$, where $\delta_j$ is the size of layer $j$, the change in the output of the network can be bounded as follows:
		\begin{equation}
			\|F(w + \delta ,\mathbf{X}) - F(w, \mathbf{X})\|_2 \leq eA\prod_{j=1}^{d}\|W_j \| \sum_{j=1}^{d}\frac{\|\delta_j \|_2}{\|W_j \|_2}. \nonumber
		\end{equation}
	\end{lemma}
	
	Lemma~\ref{lemma:PAC} gives a data-independent deterministic bound which depends on the maximum change of the output function over the domain after a perturbation. Lemma~\ref{lemma:output} bounds the change in the output a network based on the magnitude of the perturbation.
	
	\begin{theorem}
		Let input $\mathbf{X}$ be an $n\times n$ image whose norm is bounded by $A$, $f(w)$ be the classification function with $d$ hidden-layer neural network with $h$ units per hidden-layer, and satisfy $1$-Lipschitz activation $\theta (0) =0$. We assume the constant $M\geq 1$ for each layer $W_j$ satisfies:
		\begin{equation}
			\frac{1}{M} \leq \frac{\|W_j \|}{\phi_{w}} \leq M, \nonumber
		\end{equation}
		where $\phi_{w} :=(\prod_{j=1}^{d}\|W_j \| )^{1/d}$ denotes the geometric mean of $f(w)$'s spectral norms across all layers. Then, for any margin value $\gamma$, size of local training dataset on each client $m$, $\zeta >0$, with probability $1-\zeta$ over the training set, any parameter of SAM local optimizer $\tw = w + \delta$ such that $\max_{\mathbf{X} \in \mathcal{D}_i}\|F_i (w) - f(\tw)\| \leq \frac{\gamma}{8}$, we can obtain the following generalization bound:
		\begin{equation}
			\mathcal{L}^{\text{SAM}} (F(w)) \leq \hat{\mathcal{L}}^{\text{SAM}}_{\gamma}(F(w + \delta))+ \mathcal{O}\bigg(\frac{32Bd^2 h \log(dh)Q(F(w)) + d\log\frac{Nmd\log (M)}{\zeta}}{\gamma^2 m}\bigg),\nonumber
		\end{equation}
		where $Q(F(w)) := \prod_{j=1}^{d}\|W_j \|\sum_{i=1}^{d}\frac{\|W_j \|_F^2}{\|W_j \|}$ and $\|W_j \|_F^2$ is the Frobenius norm.
	\end{theorem}

	\noindent\textit{Proof.} Based on Lemma~\ref{lemma:PAC}, we choose the perturbation $\delta_j$ of each layer which is a zero-mean multivariate Gaussian distribution with diagonal covariance matrix, i.e., $\mathcal{N}(0, \lambda_j^2 I)$ and $\lambda_j = \frac{\|\tilde{W}_j \|}{\epsilon_{\tilde{W}}}\lambda$, where $\epsilon_{\tilde{W}} := (\prod_{j=1}^{d}\|W_j \| )^{1/d}$ is the geometric average of spectral norms across all layers. We consider $F(\tilde{W})$ with weights $\tilde{W}$. Since $(1+\frac{1}{d})^d \leq e$ and $\frac{1}{e} \leq (1-\frac{1}{d})^{d-1}$, for any weight vector of $W_j$ such that $|\|W_j \|_2 - \|\tilde{W}_j \|_2 | \leq \frac{1}{d}$ for every $j$, we have:
	\begin{equation}
		(1/e)^{\frac{d}{d-1}}\prod_{j=1}^{d}\|\tilde{W}_j \| \leq \prod_{j=1}^{d}\|W_{j}\| \leq e\prod_{j=1}^{d}\|\tilde{W}_i \| .\nonumber
	\end{equation}
	Then, for the $j$th layer's random perturbation vector $\delta_{j} \sim \mathcal{N}(0, \lambda_j^2 I)$, we have the following bound from \cite{tropp2012user} with $h$ representing the width of the $j$th hidden layer:
	\begin{equation}
		\mathbb{P}\bigg(\epsilon_{\tilde{W}}\frac{\|\delta_j \|}{\|\tilde{W}_j \|} >t\bigg) \leq 2he^{-\frac{t^2}{2h\lambda^2}}.\nonumber
	\end{equation}
	Based on \cite{farnia2018generalizable}, we now use a union bound over all layers for a maximum union probability of $1/2$, which implies the normalized $\epsilon_{\tilde{W}}\frac{\|\delta_j \|}{\|\tilde{W}_j \|}$ for each layer can be upper-bounded by $\lambda \sqrt{2h\log(4hd)}$. Then, for any $W$ satisfying $|\|W_j \| - \|\tilde{W}_j \| | \leq \frac{1}{d}\|\tilde{W}_j \|$ for all layer $j$'s, we obtain the following:
	\begin{equation}
		\|F(W+\delta, \mathbf{X}) - F(W,\mathbf{X})\| \leq eA \bigg(\prod_{j=1}^{d}\|w_{j}\| \bigg) \sum_{j=1}^{d}\frac{\|\delta_j \|}{\|W_j \|_2} \leq 4e dA \epsilon_{\tilde{W}}^{d-1} \lambda \sqrt{2h\log(4hd)} \leq \frac{\gamma}{8}.\nonumber
	\end{equation}
	where the last inequality is from choosing $\lambda = \frac{\gamma}{32edA\epsilon_{\tilde{W}}^{d-1}\sqrt{h\log(4hd)}}$, where the perturbation satisfies the Lemma~\ref{lemma:output}. Then, we can bound the KL-divergence in Lemma~\ref{lemma:PAC} as follows:
	\begin{equation}
		\begin{split}
			\text{KL}(w+\delta||\mathcal{P}) & \leq \sum_{j=1}^{d}\frac{\|W_j \|_F}{2\lambda_j^2} = \frac{32^2 e^2 d^2 A^2 \epsilon_{\tilde{W}}^{2d}h\log(4hd)}{\gamma^2}\sum_{j=1}^{d}\frac{\|W_j \|_F}{\|\tilde{W}_j \|}\\
			& \leq \frac{32^2 e^2 d^2 A^2 \prod_{j=1}^{d}\|W_j \|h\log(4hd)}{\gamma^2}\sum_{j=1}^{d}\frac{\|W_j \|_F}{\|\tilde{W}_j \|} = \mathcal{O}\bigg(\frac{d^2 A^2 h\log(hd)\prod_{j=1}^{d}\|W_j \|}{\gamma^2}\sum_{j=1}^{d}\frac{\|W_j \|_F}{\|W_j \|}\bigg).\nonumber
		\end{split}
	\end{equation}
	Based on \cite{farnia2018generalizable}, we have the following result given a fixed underlying distribution $\mathcal{P}$ and any $\zeta >0$ with probability $1-\zeta$ for any $W$:
	\begin{equation}
		\mathcal{L}^{\text{SAM}} (F(w)) \leq \hat{\mathcal{L}}^{\text{SAM}}_{\gamma}(F(w + \delta))+ \mathcal{O}\bigg(\frac{d^2 A^2 h\log(hd)\prod_{j=1}^{d}\|W_j \|\sum_{j=1}^{d}\frac{\|W_j \|_F}{\|W_j \|} + \log\frac{m}{\zeta}}{m\gamma^2}\bigg).\nonumber
	\end{equation}
	
	Now, we use a cover of size $\mathcal{O}(d\log(M)^d d)$ points, and hence it can demonstrate that for a fixed underlying distribution for any $\zeta >0$, with probability $1-\zeta$, we have:
	\begin{equation}
		\mathcal{L}^{\text{SAM}} (F(w)) \leq \hat{\mathcal{L}}^{\text{SAM}}_{\gamma}(F(w + \delta))+ \mathcal{O}\bigg(\frac{d^2 A^2 h\log(hd)\prod_{j=1}^{d}\|W_j \|\sum_{j=1}^{d}\frac{\|W_j \|_F}{\|W_j \|} + d\log\frac{dm\log(M)}{\zeta}}{m\gamma^2}\bigg).\nonumber
	\end{equation}
	To apply the above result to the FL network of $N$ clients, we apply a union bound to have the bound hold simultaneously for the distribution of every client, which proves for every $\zeta > 0$ with probability at least $1-\zeta$, the average SAM loss of the clients satisfies the following margin-based bound:
	\begin{equation}
		\mathcal{L}^{\text{SAM}} (F(w)) \leq \hat{\mathcal{L}}^{\text{SAM}}_{\gamma}(F(w + \delta))+ \mathcal{O}\bigg(\frac{d^2 A^2 h\log(hd)\prod_{j=1}^{d}\|W_j \|\sum_{j=1}^{d}\frac{\|W_j \|_F}{\|W_j \|} + d\log\frac{dNm\log(M)}{\zeta}}{m\gamma^2}\bigg).\nonumber
	\end{equation}
	This completes the proof. \hfill~$\Box$

	\section{Convergence Analysis of \texttt{MoFedSAM}}\label{Sec:convergencemofedsam}
	\subsection{Description of \texttt{FedSAM} Algorithm and Key Lemmas}
	We outline the \texttt{MoFedSAM} algorithm in Algorithm~\ref{ALG:MoFedSAM}. In round $r$, we sample $\S^r \subseteq [N]$ clients with $|\S^r | = S$ and then perform the following updates:
	\begin{itemize}
		\item Starting from the shared global parameters $w_{i,0}^r = w^{r-1}$, we update the local parameters for $k \in [K]$:
		\begin{equation}
			\begin{split}
				\tw_{i,k}^r & = w_{i,k-1}^r + \rho \frac{g_{i,k-1}^r}{\|g_{i,k-1}^r \|} \\
				v_{i,k-1}^r & = \beta \tg_{i,k-1}^r + (1-\beta) \Delta^r \\
				w_{i,k}^r & = w_{i,k-1}^r - \eta_l v_{i,k-1}^r ,\nonumber
			\end{split}
		\end{equation}
		\item After $K$ times local epochs, we obtain the following:
		\begin{equation}
			\Delta_i^r = w_{i,K}^r - w^r .\nonumber
		\end{equation}
		\item Compute the new global parameters using only updates from the clients $i \in \S^r$ and a global step-size $\eta_g$:
		\begin{equation}
			\begin{split}
				\Delta^{r+1} & = \frac{1}{\eta_l KS}\sum_{i\in\mathcal{S}^r}\Delta_i^r \\
				w^{r+1} & = w^{r} + \eta_g \Delta^r . \nonumber
			\end{split}
		\end{equation}
	\end{itemize}

	To prove the convergence of \texttt{MoFedSAM}, we first propose some lemmas for \texttt{MoFedSAM} as follows:
	\begin{lemma}\label{lemma:moxdrift}
		(Bounded $\mathcal{E}_{w}$ of \texttt{MoFedSAM}). Suppose our functions satisfies Assumptions~\ref{ass:smooth}-\ref{ass:sigmag}. Then, for any $i \in [N]$, $k \in [K]$ and $r \in [R]$ the updates of \texttt{MoFedSAM} for any learning rate satisfying $\eta_l \leq \frac{1}{\sqrt{30}\beta KL}$ have the drift due to $w_{i,k} - w$:
		\begin{equation}
			\begin{split}
				\mathcal{E}_{w} & = \frac{1}{N}\sum_{i}\E [\|w_{i,k} - w\|^2 ]\\
				& \leq 5K \eta_l^2 (2\beta^2 L^2 \eta_l^2 \rho^2 \sigma_l^2 + 7K\beta^2 \eta_l^2 (3\sigma_g^2 +6L^2 \rho^2 ) + 14K(1-\beta)^2 \eta_l^2 \|\nabla f(\tw) \|^2 ) + 28\beta^2 K^3  L^4 \eta_l^4 \rho^2 . \nonumber
			\end{split}
		\end{equation}
	\end{lemma}
	
	\noindent\textit{Proof.} Recall that the local update on client $i$ is $w_{i,k} = w_{i,k-1} - \beta \eta_l g_{i,k-1} + (1-\beta)\Delta^r$. Then, we have:
	\begin{equation}
		\begin{split}
			& \E \|w_{i,k} - w\|^2 = \E\|w_{i,k-1} - w - \eta_l(\beta \tg_{i,k-1} + (1-\beta)\Delta )\|^2 \\
			& \overset{\text{(a)}}{\leq} \E\|w_{i,k-1} - w - \beta\eta_l ( \tg_{i,k-1} - \nabla f_i (\tw_{i,k-1}) + \nabla f_i (\tw_{i,k-1}) - \nabla f_i (\tw) + \nabla f_i (\tw) - \nabla f(\tw) + \nabla f(\tw)) + \eta_l (1-\beta)\Delta \|^2 \\
			& \overset{\text{(b)}}{\leq} \bigg(1+\frac{1}{2K-1} + 2\beta^2 L^2 \eta_l^2 \bigg) \E\|w_{i,k-1} - w\|^2 + 2\beta^2 L^2 \eta_l^2 \rho^2 \sigma_l^2 + 7K^2 \beta \eta_l^2 \E\|\nabla f_i (\tw_{i,k-1}) - \nabla f_i (\tw)\|^2 \\
			& + 7K\beta^2 \eta_l^2 (3\sigma_g^2 + 6L^2 \rho^2 ) + 7K\beta^2 \eta_l^2 \|\nabla f(\tw)\|^2 + 7K\eta_l^2 (1-\beta)^2 \|\Delta\|^2 \\
			& \overset{\text{(c)}}{\leq} \bigg(1+\frac{1}{2K-1} + 2\beta^2 L^2 \eta_l^2 + 14K\beta^2 L^2 \eta_l^2 \bigg) \E\|w_{i,k-1} - w\|^2 + 2\beta^2 L^2 \eta_l^2 \rho^2 \sigma_l^2 + 7K(1-\beta)^2 \eta_l^2 \|\Delta \|^2 \\
			& + 14K\beta^2 L^2 \eta_l^2 \E \|\delta_{i,k-1} - \delta \|^2 + 7K\beta^2 \eta_l^2 (3\sigma_g^2 +6L^2 \rho^2 ) + 7\beta^2 K \E\|\nabla f(\tw)\|^2 \\
			& \overset{\text{(d)}}{\leq} \bigg(1 + \frac{1}{2K-1}+ 2\beta^2 L^2 \eta_l^2 +14\beta^2 KL^2 \eta_l^2 \bigg)\E\|w_{i,k-1} - w\|^2 + 2\beta^2 L^2 \eta_l^2 \rho^2 \sigma_l^2 + 14K\beta^2 L^2 \eta_l^2 \E \|\delta_{i,k} - \delta \|^2 \\
			& + 7K\beta^2 \eta_l^2 (3\sigma_g^2 +6L^2 \rho^2 ) + 7K(1-\beta)^2 \eta_l^2 \|\Delta \|^2 + 7\beta^2 K\E\|\nabla f(\tw)\|^2\\
			& \overset{\text{(e)}}{\leq} \bigg(1 + \frac{1}{2K-1}+ 2\beta^2 L^2 \eta_l^2 +14\beta^2 KL^2 \eta_l^2 \bigg)\E\|w_{i,k-1} - w\|^2 + 2\beta^2 L^2 \eta_l^2 \rho^2 \sigma_l^2 + 14K\beta^2 L^2 \eta_l^2 \E \|\delta_{i,k} - \delta \|^2 \\
			& + 7K\beta^2 \eta_l^2 (3\sigma_g^2 +6L^2 \rho^2 ) + 14K(1-\beta)^2 \eta_l^2 \|\nabla f(\tw) \|^2 ,\nonumber
		\end{split}
	\end{equation}
	where (a) follows from the fact that $\tg_{i,k-1}$ is an unbiased estimator of $\nabla f_i (\tw_{i,k-1})$ and Lemma~\ref{lemma:independent}; (b) is from Lemmas~\ref{lemma:dependent} and \ref{lemma:sigmag}; (c) is from Assumption~\ref{ass:sigmal}; Lemma~\ref{lemma:dependent}; (d) is from Assumption~\ref{ass:sigmag} and (e) is due to the fact that $\Delta \approx \nabla f(\tw)$ and $\beta < \frac{1}{2}$.
	
	Averaging over the clients $i$ and learning rate satisfies $\eta_l \leq \frac{1}{\sqrt{30}\beta KL}$, we have:
	\begin{equation}
		\begin{split}
			& \mathcal{E}_w \leq \bigg(1 + \frac{1}{2K-1}+ 2\beta^2 L^2 \eta_l^2 +14\beta^2 KL^2 \eta_l^2 \bigg)\E\|w_{i,k-1} - w\|^2 + 2\beta^2 L^2 \eta_l^2 \rho^2 \sigma_l^2 + 14K\beta^2 L^2 \eta_l^2 \E \|\delta_{i,k} - \delta \|^2 \\
			& + 7K\beta^2 \eta_l^2 (3\sigma_g^2 +6L^2 \rho^2 ) + 14K(1-\beta)^2 \eta_l^2 \|\nabla f(\tw) \|^2\\
			& \overset{\text{(a)}}{\leq} \bigg(1+\frac{1}{K-1}\bigg)\frac{1}{N}\sum_{i\in [N]}\E\|w_{i,k-1} - w\|^2 + 2\beta^2 L^2 \eta_l^2 \rho^2 \sigma_l^2 \\
			&+ 14K\beta^2 L^2 \eta_l^2 \frac{1}{N}\sum_{i\in[N]} \E\|\delta_{i,k} - \delta\|^2 + 7K\beta^2 \eta_l^2 (3\sigma_g^2 +6L^2 \rho^2 ) + 14K(1-\beta)^2 \eta_l^2 \|\nabla f(\tw) \|^2\\
			& \leq \sum_{\tau = 0}^{k-1}\bigg(1+\frac{1}{K-1}\bigg)^\tau [2\beta^2 L^2 \eta_l^2 \rho^2 \sigma_l^2 + 7K\beta^2 \eta_l^2 (3\sigma_g^2 +6L^2 \rho^2 ) + 14K(1-\beta)^2 \eta_l^2 \|\nabla f(\tw) \|^2 ] + 14K\beta^2 L^2 \eta_l^2 \frac{1}{N}\sum_{i\in[N]} \E\|\delta_{i,k} - \delta \|^2 \\
			& \overset{\text{(b)}}{\leq} 5K (2\beta^2 L^2 \eta_l^2 \rho^2 \sigma_l^2 + 7K\beta^2 \eta_l^2 (3\sigma_g^2 +6L^2 \rho^2 ) + 14K(1-\beta)^2 \eta_l^2 \|\nabla f(\tw) \|^2 ) + 28\beta^2 K^3  L^4 \eta_l^4 \rho^2 ,\nonumber
		\end{split}
	\end{equation}
	where (a) is due to the fact that $\eta_l \leq \frac{1}{\sqrt{30}\beta KL}$ and $\beta \leq \frac{1}{2}$ and (b) is from Lemma~\ref{lemma:deltadrift}. \hfill~$\Box$
	
	\subsection{Convergence Analysis of Full client participation \texttt{MoFedSAM}}
	\begin{lemma}\label{lemma:mod2}
		For the full client participation scheme, we can bound $\E [\|\Delta^r \|^2 ]$ as follows:
		\begin{equation}
			\E_r [\|\Delta^r \|^2 ] \leq \frac{2\beta^2 L^2 \rho^2}{KN}\sigma_l^2 + \frac{2}{K^2 N^2}\E_r \bigg[\bigg\|\sum_{i,k}\beta \nabla f_i(\tw_{i,k}^r ) + (1+\beta)\Delta^r \bigg\|^2 \bigg]. \nonumber
		\end{equation}
	\end{lemma}
	
	\noindent\textit{Proof.} For the full client participation strategy, we have:
	\begin{equation}
		\begin{split}
			\E_r [\|\Delta^{r+1} \|^2 ] &\overset{\text{(a)}}{\leq} \frac{1}{K^2 N^2 \eta_l^2}\E_r \bigg[\bigg\|\sum_{i,k}\beta\eta_l \tg_{i,k}^r + (1-\beta)\eta_l \Delta^r \bigg\|^2 \bigg]\\
			& \overset{\text{(b)}}{\leq} \frac{\beta^2}{K^2 N^2} \E_r \bigg[\bigg\|\sum_{i,k}(\tg_{i,k}^r - \nabla f_i (\tw_{i,k}^r )) \bigg\|^2 \bigg] + \frac{1}{K^2 N^2}\E_r \bigg[\bigg\|\sum_{i,k} \beta\nabla f_i(\tw_{i,k}) + (1-\beta)\Delta^r \bigg\|^2 \bigg]\\
			& \overset{\text{(c)}}{\leq} \frac{\beta^2 L^2 \rho^2}{KN}\sigma_l^2 + \frac{1}{K^2 N^2}\E_r \bigg[\bigg\|\sum_{i,k}\beta \nabla f_i(\tw_{i,k}^r ) + (1-\beta)\Delta^r \bigg\|^2 \bigg]\\
			& \overset{\text{(d)}}{\leq} \frac{\beta^2 L^2 \rho^2}{KN}\sigma_l^2 + \frac{2(1-\beta)^2}{KN}\|f(\tw^r )\|^2 + \frac{\beta^2}{K^2 N^2 }\E_r \bigg[\bigg\|\sum_{i,k}f_i (\tw_{i,k}^r )\bigg\|^2 \bigg],\nonumber
		\end{split}
	\end{equation}
	where (a) is from Lemma~\ref{lemma:dependent}; (b) is from Lemma~\ref{lemma:independent} and (c) is from Lemma~\ref{lemma:SAMvariance}. \hfill~$\Box$
	
	\begin{lemma}\label{lemma:decentmosam}
		(Descent Lemma of full client participation \texttt{MoFedSAM}). For all $r \in R-1$ and $i \in \S^r$, with the choice of learning rate, the iterates generated by \texttt{MoFedSAM} under full client participation in Algorithm~\ref{ALG:MoFedSAM} satisfy:
		\begin{equation}
			\begin{split}
				\E_r [f(\tw^{r+1})] & \leq f(\tw^r ) - K\eta_g \eta_l \bigg(\frac{1}{2} - 20K^2 L^2 \eta_l^2 B^2 \bigg)\|\nabla f(\tw^r )\|^2 + K\eta_g \eta_l (6K^2 \eta_l^2 \beta^4 \rho^2 + 5K^2 \eta_l \beta^4 \rho^2 \sigma^2 \\
				&+ 20 K^3 \eta_l^3 \beta^2 G^2 + 16K^3 \eta_l^4 \beta^6 \rho^2 + \frac{\eta_g \eta_l \beta^3 \rho^2}{N}\sigma^2 )  \nonumber
			\end{split}
		\end{equation}
		where the expectation is w.r.t. the stochasticity of the algorithm.
	\end{lemma}
	
	\noindent\textit{Proof.}
	\begin{equation}\label{Eq:mofedsam}
		\begin{split}
			\E_{r} [F(w^{r+1})] & = \E_r [f(\tw^{r+1} )] \leq f(\tw^r ) + \E_r \langle \nabla f(\tw^r ), \tw^{r+1} - \tw^r ]\rangle +\frac{L}{2}\E_r [\|\tw^{r+1} - \tw^r \|^2 ]\\
			& \overset{\text{(a)}}{=} f(\tw^r ) + \eta_g \E_r \langle \nabla f(\tw^r ) ,  -\Delta^{r+1} + \beta \nabla f(\tw^r ) - \beta \nabla f(\tw^r )\rangle + \frac{L}{2}\eta_g^2 \E_r [\|\Delta^{r+1} \|^2]\\
			& \overset{\text{(b)}}{=} f(\tw^r ) - \beta \eta_g \|\nabla f(\tw^r )\|^2 + \eta_g \langle \nabla f(\tw^r ) ,\E_r [-\Delta^{r+1} + \beta \nabla f(\tw^r ) ]\rangle + \frac{L}{2}\eta_g^2 \E_r [\|\Delta^{r+1} \|^2 ], 
		\end{split}
	\end{equation}
	where (a) is from the iterate update given in Algorithm~\ref{Alg:FedSAM} and (b) results from the unbiased estimators. 
	
	For the third term, we bound it as follows:
	\begin{equation}\label{Eq:third}
		\begin{split}
			& \langle \nabla f(\tw^r ) ,\E_r [-\Delta^{r+1} + \beta \nabla f(\tw^r ) ]\rangle \\
			& = -(1-\beta)\|\nabla f(\tw^r )\|^2 + \bigg\langle \sqrt{\beta}\nabla f(\tw^r ), \E_r \bigg[-\frac{\sqrt{\beta}}{KN \eta_l}\sum_{i,k}(\eta_l \nabla f_i (\tw_{i,k}^r ) - \eta_l \nabla f_i (\tw^r ))\bigg]\bigg\rangle\\
			& \overset{\text{(a)}}{=} \bigg(\frac{3\beta}{2} - 1\bigg)\|\nabla f(\tw^r )\|^2 + \frac{\beta}{2K^2 N^2}\E_r \bigg\|\sum_{i,k}(\nabla f_i (\tw_{i,k}^r ) - \nabla f_i (\tw^r ))\bigg\|^2 - \frac{\beta}{2K^2 N^2}\E_r \bigg\|\sum_{i,k}\nabla f_i (\tw_{i,k}^r )\bigg\|^2\\
			& \overset{\text{(b)}}{\leq} \bigg(\frac{3\beta}{2} - 1\bigg)\|\nabla f(\tw^r )\|^2 + \frac{\beta}{2KN}\sum_{i,k}\E_r \|\nabla f_i (\tw_{i,k}^r ) - \nabla f_i (\tw^r )\|^2 - \frac{\beta}{2K^2 N^2}\E_r \bigg\|\sum_{i,k}\nabla f_i (\tw_{i,k}^r )\bigg\|^2 \\
			& \overset{\text{(c)}}{\leq} \bigg(\frac{3\beta}{2} - 1\bigg)\|\nabla f(\tw^r )\|^2 + \frac{\beta}{2KN}\sum_{i,k}\E_r \|\nabla f_i (\tw_{i,k}^r ) - \nabla f_i (\tw^r )\|^2 - \frac{\beta}{2K^2 N^2}\E_r \bigg\|\sum_{i,k}\nabla f_i (\tw_{i,k}^r ) \bigg\|^2 \\
			& \overset{\text{(d)}}{\leq} \bigg(\frac{3\beta}{2} - 1\bigg)\|\nabla f(\tw^r )\|^2 + \frac{\beta L^2}{2KN}\sum_{i,k}\E_r \|\tw_{i,k}^r - \tw^r \|^2  - \frac{\beta}{2K^2 N^2}\E_r \bigg\|\sum_{i,k}\nabla f_i (\tw_{i,k}^r ) \bigg\|^2 \\
			& \overset{\text{(e)}}{\leq} \bigg(\frac{3\beta}{2} - 1\bigg)\|\nabla f(\tw^r )\|^2 + \beta L^2 (\mathcal{E}_{w} + \mathcal{E}_{\delta}) - \frac{\beta}{2K^2 N^2}\E_r \bigg\|\sum_{i,k}\nabla f_i (\tw_{i,k}^r ) \bigg\|^2 ,
		\end{split}
	\end{equation}
	where (a) is from that $\Delta^r = \nabla f(\tw^r )$ and $\nabla f(\tw^r ) = \sum_i \nabla f_i (\tw^r )$; (b), (c) and (e) are from Lemma~\ref{lemma:dependent} and (d) is from Assumption~\ref{ass:smooth}. Plugging \eqref{Eq:third} into \eqref{Eq:mofedsam}, we have:
	\begin{equation}
		\begin{split}
			&\E_r [f(\tw^{r+1} )] \\
			& \leq f(\tw^r ) - \bigg(\eta_g -\frac{\beta \eta_g}{2} \bigg) \|\nabla f(\tw^r)\|^2 + \beta L^2 \eta_g (\mathcal{E}_{w} + \mathcal{E}_{\delta}) - \frac{\beta \eta_g}{2K^2 N^2}\E_r \bigg\|\sum_{i,k}\nabla f_i (\tw_{i,k}^r )\bigg\|^2 + \frac{L \eta_g^2}{2}\E_r [\|\Delta^{r+1} \|^2 ]\\
			& \overset{\text{(a)}}{\leq} f(\tw^r ) - \bigg(\frac{3\beta \eta_g}{4} - \frac{2(1-\beta)^2 L \eta_g}{KN} \bigg)\|\nabla f(\tw^r)\|^2 +\beta L^2 \eta_g (\mathcal{E}_{w}+\mathcal{E}_{\delta})+ \frac{\beta^2 L^3 \rho^2 \eta_g^2}{2KN}\sigma_l^2 \\
			& - \frac{\beta \eta_g}{2K^2 N^2}\E_r \bigg\|\sum_{i,k}\nabla f_i (\tw_{i,k}^r ) \bigg\|^2 + \frac{L\beta^2 \eta_g^2}{2K^2 N^2}\E_r \bigg\|\sum_{i,k} \nabla f_i (\tw_{i,k}^r ) \bigg\|^2\\
			& \overset{\text{(b)}}{\leq} f(\tw^r ) - \beta \eta_g \bigg(\frac{3}{4} - \frac{2(1-\beta)L}{KN} - 70(1-\beta)K^2 L^2 \eta_l^2 \bigg)\|\nabla f(\tw^r )\|^2 \\
			&+ \beta \eta_g \bigg(10\beta^2 L^4 \eta_l^2 \rho^2 \sigma_l^2 + 35\beta^2 KL^2 \eta_l^2 (3\sigma_g^2 + 6L^2 \rho^2 ) + 28\beta^2 K^3 L^6 \eta_l^4 \rho^2 + 2K^2 L^4 \eta_l^2 \rho^2 + \frac{\beta L^3 \eta_g^2 \rho^2}{2KN}\sigma_l^2 \bigg)\\
			& \overset{\text{(c)}}{\leq} f(\tw^r ) - C\beta \eta_g \|\nabla f(\tw^r )\|^2 \\
			& + \beta \eta_g \bigg(10\beta^2 L^4 \eta_l^2 \rho^2 \sigma_l^2 + 35\beta^2 KL^2 \eta_l^2 (3\sigma_g^2 + 6L^2 \rho^2 ) + 28\beta^2 K^3 L^6 \eta_l^4 \rho^2 + 2K^2 L^4 \eta_l^2 \rho^2 + \frac{\beta L^3 \eta_g^2 \rho^2}{2KN}\sigma_l^2 \bigg), \nonumber
		\end{split}
	\end{equation}
	(a) is from Lemma~\ref{lemma:mod2}; (b) is from Lemmas~\ref{lemma:deltadrift}, \ref{lemma:moxdrift} and due to the fact that $\eta_g \leq \frac{1}{\beta L}$ and (c) is due to the fact that the condition $\frac{3}{4} - \frac{2(1-\beta)L}{KN} - 70(1-\beta)K^2 L^2 \eta_l^2 > C >0$ and $\beta \leq \frac{1}{2}$ hold. \hfill~$\Box$
	
	\begin{theorem}
		(Convergence of \texttt{MoFedSAM}). Let constant local and global learning rates $\eta_l \leq \frac{1}{\sqrt{30}\beta KL} $, $\eta_g \leq \frac{1}{\beta L}$ and $\beta \leq \frac{1}{2}$ and the condition $\frac{3}{4} - \frac{2(1-\beta)L}{KN} - 70(1-\beta)K^2 L^2 \eta_l^2 > C >0$ holds. Under Assumptions~\ref{ass:smooth}-\ref{ass:sigmal} and with full client participation, the sequence of outputs $\{w^r \}$ generated by \texttt{FedGSAM} satisfies:
		\begin{equation}
			\min_{r\in [R]}\E\|\nabla F(w^r )\|^2 \leq \frac{f^0 - f^*}{C\beta \eta_g} + \Phi ,\nonumber
		\end{equation}
		where $\Phi = \frac{1}{C}(20\beta^3 L^4 \eta_l^2 \rho^2 \sigma^2 + 25\beta^2 K^2 L^2 \eta_l^2 G^2 + 20\beta^2 K^4 L^5 \eta_l^4 \rho^2 + 4\beta^2 KL^4 \eta_l^2 \rho^2 + \frac{\beta L^3 \rho^2 \eta_g^2}{2KN} )$. If we choose the learning rates $\eta_l = O(\frac{1}{\sqrt{RK}\beta L})$, $\eta_g = O(\frac{\sqrt{KN}}{\sqrt{R}\beta L})$ and the perturbation amplitude $\rho$ proportional to the learning rate, e.g., $\rho = \frac{1}{\sqrt{R}}$, we have:
		\begin{equation}
			\frac{1}{R}\sum_{r=1}^{R}\E [\|F(w^{r+1})\|] = \mathcal{O}\bigg(\frac{F\beta L}{\sqrt{RKN}}+ \frac{\beta^2 \sigma_g^2}{R} + \frac{L \sigma_l^2}{R^2 \beta} + \frac{L^2}{R^2 \beta^2}\bigg).\nonumber
		\end{equation}
	\end{theorem}
	
	\noindent\textit{Proof.} Summing the result of Lemma~\ref{lemma:decentmosam} for $r = [R]$ and multiplying both sides by $\frac{1}{C\beta\eta_g R}$, we have:
	\begin{equation}
		\begin{split}
			\frac{1}{R}\sum_{r=1}^{R}&\E [\|F(w^{r+1})\|] \leq \frac{F}{C\beta\eta_g R}\\
			& + \frac{1}{C}\bigg(10\beta^2 L^4 \eta_l^2 \rho^2 \sigma_l^2 + 35\beta^2 KL^2 \eta_l^2 (3\sigma_g^2 + 6L^2 \rho^2 ) + 28\beta^2 K^3 L^6 \eta_l^4 \rho^2 + 2K^2 L^4 \eta_l^2 \rho^2 + \frac{\beta L^3 \eta_g^2 \rho^2}{2KN}\sigma_l^2 \bigg) ,\nonumber
		\end{split}
	\end{equation}
	where it is from that $F= f(\tw^0) - f^* \leq f(\tw^r )-f(\tw^{r+1})$. If we choose the learning rates $\eta_l = O(\frac{1}{\sqrt{RK}\beta L})$, $\eta_g = O(\frac{\sqrt{KN}}{\sqrt{R}\beta L})$ and the perturbation amplitude $\rho$ proportional to the learning rate, e.g., $\rho = \frac{1}{\sqrt{R}}$, we have
	\begin{equation}
		\frac{1}{R}\sum_{r=1}^{R}\E [\|F(w^{r+1})\|] = \mathcal{O}\bigg( \frac{F\beta L}{\sqrt{RKN}}+ \frac{\beta^2 \sigma_g^2}{R} + \frac{L^2 \sigma_l^2}{R^2 K} + \frac{L\sigma_l^2}{R^2 \beta} + \frac{\beta L^2}{R^2} + \frac{KL^2}{R^3 \beta^2} + \frac{L^2}{R^2 \beta^2} \bigg).\nonumber
	\end{equation}
	If we omit the larger order of each part, we have:
	\begin{equation}
		\frac{1}{R}\sum_{r=1}^{R}\E [\|F(w^{r+1})\|] = \mathcal{O}\bigg(\frac{F\beta L}{\sqrt{RKN}}+ \frac{\beta^2 \sigma_g^2}{R} + \frac{L \sigma_l^2}{R^2 \beta} + \frac{L^2}{R^2 \beta^2}\bigg).\nonumber
	\end{equation}
	This completes the proof. \hfill~$\Box$

	\subsection{Convergence Analysis of Partial client participation \texttt{MoFedSAM}}
	\begin{lemma}\label{lemma:moforth}
		For the partial client participation, we can bound $\E_r [\|\Delta^r \|^2 ]$ as follows:
		\begin{equation}
			\E_r [\|\Delta^r \|^2 ] \leq \frac{KL^2 \eta_l^2 \rho^2}{S}\sigma_l^2 + \frac{\eta_l^2}{S^2} \bigg[\bigg \|\sum_{i} \mathbb{P}\{i\in\S^r \} \sum_{j=0}^{K-1} \nabla f_i(\tw_{i,j}^r ) \bigg\|^2 \bigg]. \nonumber
		\end{equation}
	\end{lemma}
	
	\noindent\textit{Proof.}
	\begin{equation}
		\begin{split}
			& \E_r [\|\Delta^r \|^2 ] \overset{\text{(a)}}{\leq} \frac{1}{K^2 S^2 \eta_l^2}\E_r \bigg[\bigg\|\sum_{i\in\S^r}\sum_{k} \beta \eta_l \tg_{i,k}^r + (1-\beta) \eta_l \Delta^r \bigg\|^2 \bigg] \\
			& = \frac{1}{K^2 S^2 \eta_l^2}\E_r \bigg[\bigg\|\sum_{i} \mathbb{I}\{i\in\S^r \}\sum_{k} \beta \eta_l \tg_{i,k}^r - (1-\beta)\Delta^r \bigg\|^2 \bigg]\\
			& \overset{\text{(b)}}{=} \frac{\beta^2}{K^2 S^2} \E_r \bigg[\bigg\|\sum_{i}\sum_{j=0}^{K-1}(\tg_{i,j}^r - \nabla f_i (\tw_{i,j}^r )) \bigg\|^2 \bigg] + \frac{1}{K^2 S^2}\E_r \bigg[\bigg\|\sum_{i}\mathbb{I}\{i\in\S^r \}\sum_{j=0}^{K-1} \beta \nabla f_i(\tw_{i,j}^r ) + (1-\beta)\Delta^r \bigg\|^2 \bigg]\\
			& \overset{\text{(c)}}{\leq} \frac{\beta^2 L^2 \rho^2}{KS}\sigma_l^2 + \frac{2(1-\beta)^2}{KS}\|f(\tw^r )\|^2 +  \frac{2\beta^2}{K^2 S^2}\bigg[\bigg\|\sum_{i}\mathbb{P}\{i\in\S^r \} \sum_{j=0}^{K-1} \nabla f_i(\tw_{i,j}) \bigg\|^2 \bigg]\\
			& = \frac{\beta^2 L^2 \rho^2}{KS}\sigma_l^2 + \frac{2(1-\beta)^2}{KS}\|f(\tw^r )\|^2 +  \frac{2\beta^2}{K^2 SN}\sum_{i=1}^{N} \E_r \bigg\|\sum_{j=0}^{K-1}\nabla f_i (\tw_{i,j}^r )\bigg\|^2 + \frac{2\beta^2 (S-1)}{K^2 SN^2} \E_r \bigg\|\sum_{i=1}^{N} \sum_{j=0}^{K-1}\nabla f_i (\tw_{i,j}^r )\bigg\|^2 ,\nonumber
		\end{split}
	\end{equation}
	where (a) is from Lemma~\ref{lemma:dependent}; (b) is from Lemma~\ref{lemma:independent} and (c) is from Lemma~\ref{lemma:SAMvariance}.\hfill~$\Box$

	\begin{lemma}\label{lemma:partdecentmosam}
		(Descent Lemma of partial client participation \texttt{MoFedSAM}). For all $r \in R-1$ and $i \in \S^r$, with the choice of learning rate, the iterates generated by \texttt{MoFedSAM} under partial client participation in Algorithm~\ref{ALG:MoFedSAM} satisfy:
		\begin{equation}
			\begin{split}
				\E_r [f(\tw^{r+1})] & \leq f(\tw^r ) - K\eta_g \eta_l \bigg(\frac{1}{2} - 20K^2 L^2 \eta_l^2 B^2 \bigg)\|\nabla f(\tw^r )\|^2 + K\eta_g \eta_l (6K^2 \eta_l^2 \beta^4 \rho^2 + 5K^2 \eta_l \beta^4 \rho^2 \sigma^2 \\
				&+ 20 K^3 \eta_l^3 \beta^2 G^2 + 16K^3 \eta_l^4 \beta^6 \rho^2 + \frac{\eta_g \eta_l \beta^3 \rho^2}{N}\sigma^2 )  \nonumber
			\end{split}
		\end{equation}
		where the expectation is w.r.t. the stochasticity of the algorithm.
	\end{lemma}
	
	\noindent\textit{Proof.}
	\begin{equation}\label{Eq:mopartfedsam}
		\begin{split}
			\E_{r} [F(w^{r+1})] & = \E_r [f(\tw^{r+1} )] \leq f(\tw^r ) + \E_r \langle \nabla f(\tw^r ), \tw^{r+1} - \tw^r ]\rangle +\frac{L}{2}\E_r [\|\tw^{r+1} - \tw^r \|^2 ]\\
			& = f(\tw^r ) - \beta \eta_g \|\nabla f(\tw^r )\|^2 + \eta_g \langle \nabla f(\tw^r ) ,\E_r [-\Delta^{r+1} + \beta \nabla f(\tw^r ) ]\rangle + \frac{L}{2}\eta_g^2 \E_r [\|\Delta^{r+1} \|^2 ].
		\end{split}
	\end{equation}

	Similar to full client participation strategy, we bound the third term in \eqref{Eq:mopartfedsam} as follows:
	\begin{equation}\label{Eq:mopartthird}
		\langle \nabla f(\tw^r ) ,\E_r [-\Delta^{r+1} + \beta \nabla f(\tw^r ) ]\rangle \leq \bigg(\frac{3\beta}{2} - 1\bigg)\|\nabla f(\tw^r )\|^2 + \beta L^2 (\mathcal{E}_{w} + \mathcal{E}_{\delta}) - \frac{\beta}{2K^2 N^2}\E_r \bigg\|\sum_{i,k}\nabla f_i (\tw_{i,k}^r ) \bigg\|^2 ,
	\end{equation}
	
	Plugging \eqref{Eq:mopartthird} into \eqref{Eq:mopartfedsam}, we have:
	\begin{equation}
		\begin{split}
			&\E_r [f(\tw^{r+1} )] \\
			& \leq f(\tw^r ) - \bigg(\eta_g -\frac{\beta \eta_g}{2} \bigg) \|\nabla f(\tw^r)\|^2 + \beta L^2 \eta_g (\mathcal{E}_{w} + \mathcal{E}_{\delta}) - \frac{\beta \eta_g}{2K^2 N^2}\E_r \bigg\|\sum_{i,k}\nabla f_i (\tw_{i,k}^r )\bigg\|^2 + \frac{L \eta_g^2}{2}\E_r [\|\Delta^{r+1} \|^2 ]\\
			& \overset{\text{(a)}}{\leq} f(\tw^r ) - \bigg(\frac{3\beta \eta_g}{4} - \frac{2(1-\beta)^2 L \eta_g}{KS} \bigg)\|\nabla f(\tw^r)\|^2 +\beta L^2 \eta_g (\mathcal{E}_{w}+\mathcal{E}_{\delta})+ \frac{\beta^2 L^3 \rho^2 \eta_g^2}{2KS}\sigma_l^2 \\
			& - \frac{\beta \eta_g}{2K^2 N^2}\E_r \bigg\|\sum_{i,k}\beta \nabla f_i (\tw_{i,k}^r ) \bigg\|^2 + \frac{L\beta^2 \eta_g^2}{2K^2 SN}\sum_{i}\E_r \bigg\|\sum_{k} \nabla f_i (\tw_{i,k}^r ) \bigg\|^2 + \frac{L\beta^2 (S-1) \eta_g^2}{K^2 SN^2}\E_r \bigg\|\sum_{i,k}\nabla f_i (\tw_{i,k}^r )\bigg\|^2 \\
			& \overset{\text{(b)}}{\leq} f(\tw^r ) - \beta \eta_g \bigg(\frac{3}{4} - \frac{2(1-\beta)L}{KN} - 70(1-\beta)K^2 L^2 \eta_l^2 - \frac{90\beta L^3 \eta_g \eta_l^2}{S} - \frac{3\beta L \eta_g}{2S}\bigg)\|\nabla f(\tw^r )\|^2 \\
			&+ \beta \eta_g \bigg(10\beta^2 L^4 \eta_l^2 \rho^2 \sigma_l^2 + 35\beta^2 KL^2 \eta_l^2 (3\sigma_g^2 + 6L^2 \rho^2 ) + 28\beta^2 K^3 L^6 \eta_l^4 \rho^2 + 2K^2 L^4 \eta_l^2 \rho^2 + \frac{\beta L^3 \eta_g^2 \rho^2}{2KS}\sigma_l^2 \\
			& + \frac{\beta L\eta_g}{K^2 SN}\bigg(30NK^2 L^4 \eta_l^2 \rho^2 \sigma_l^2 + 270NK^3 L^2 \eta_l^2 \sigma_g^2 + 540 NK^2 L^4 \eta_l^2 \rho^2 + 72K^4 L^6 \eta_l^4 \rho^2 + 6NK^4 L^2 \eta_l^2 \rho^2 + 4NK^2 \sigma_g^2 + 3NK^2 L^2 \rho^2  \bigg)\bigg)\\
			& \overset{\text{(c)}}{\leq} f(\tw^r ) - C\beta \eta_g \|\nabla f(\tw^r )\|^2 \\
			&+ \beta \eta_g \bigg(10\beta^2 L^4 \eta_l^2 \rho^2 \sigma_l^2 + 35\beta^2 KL^2 \eta_l^2 (3\sigma_g^2 + 6L^2 \rho^2 ) + 28\beta^2 K^3 L^6 \eta_l^4 \rho^2 + 2K^2 L^4 \eta_l^2 \rho^2 + \frac{\beta L^3 \eta_g^2 \rho^2}{2KS}\sigma_l^2 \\
			& + \frac{\beta L\eta_g}{K^2 SN}\bigg(30NK^2 L^4 \eta_l^2 \rho^2 \sigma_l^2 + 270NK^3 L^2 \eta_l^2 \sigma_g^2 + 540 NK^2 L^4 \eta_l^2 \rho^2 + 72K^4 L^6 \eta_l^4 \rho^2 + 6NK^4 L^2 \eta_l^2 \rho^2 + 4NK^2 \sigma_g^2 + 3NK^2 L^2 \rho^2  \bigg)\bigg), \nonumber
		\end{split}
	\end{equation}
	(a) is from Lemma~\ref{lemma:mod2}; (b) is from Lemmas~\ref{lemma:deltadrift}, \ref{lemma:moxdrift} and due to the fact that $\eta_g \leq \frac{S}{2\beta L(S-1)}$ and (c) is due to the fact that the condition $\frac{3}{4} - \frac{2(1-\beta)L}{KN} - 70(1-\beta)K^2 L^2 \eta_l^2 - \frac{90\beta L^3 \eta_g \eta_l^2}{S} - \frac{3\beta L \eta_g}{2S} > C >0$ and $\beta \leq \frac{1}{2}$ hold. \hfill~$\Box$

	\begin{theorem}
		Let constant local and global learning rates $\eta_l$ and $\eta_g$ be chosen as such that $\eta_l \leq \frac{1}{\sqrt{30}\beta KL}$, $\eta_g \leq \frac{S}{2\beta L(S-1)}$ and the condition $\frac{3}{4} - \frac{2(1-\beta)L}{KN} - 70(1-\beta)K^2 L^2 \eta_l^2 - \frac{90\beta L^3 \eta_g \eta_l^2}{S} - \frac{3\beta L \eta_g}{2S} >0$ holds. Under Assumption~\ref{ass:smooth}-\ref{ass:sigmal} and with partial client participation, the sequence of outputs $\{w^r \}$ generated by \texttt{MoFedSAM} satisfies:
		\begin{equation}
			\min_{r\in [R]}\E\|\nabla F(w^r )\|^2 \leq \frac{F^0 - F^*}{C\beta\eta_g} + \Phi ,\nonumber
		\end{equation}
		where $\Phi = \frac{1}{C}[10\beta^2 L^4 \eta_l^2 \rho^2 \sigma_l^2 + 35\beta^2 KL^2 \eta_l^2 (3\sigma_g^2 + 6L^2 \rho^2 ) + 28\beta^2 K^3 L^6 \eta_l^4 \rho^2 + 2K^2 L^4 \eta_l^2 \rho^2 + \frac{\beta L^3 \eta_g^2 \rho^2}{2KS}\sigma_l^2 + \frac{\beta L\eta_g}{K^2 SN}\bigg(30NK^2 L^4 \eta_l^2 \rho^2 \sigma_l^2 + 270NK^3 L^2 \eta_l^2 \sigma_g^2 + 540 NK^2 L^4 \eta_l^2 \rho^2 + 72K^4 L^6 \eta_l^4 \rho^2 + 6NK^4 L^2 \eta_l^2 \rho^2 + 4NK^2 \sigma_g^2 + 3NK^2 L^2 \rho^2  \bigg)]$. If we choose the learning rates $\eta_l = \frac{1}{\sqrt{RK}\beta L}$, $\eta_g = \frac{\sqrt{KS}}{\sqrt{R}\beta L}$ and perturbation amplitude $\rho$ proportional to the learning rate, e.g., $\rho = \frac{1}{\sqrt{R}}$, we have:
		\begin{equation}
			\frac{1}{R}\sum_{r=1}^{R}\E [\|F(w^{r+1})\|] = \mathcal{O}\bigg(\frac{FL}{\sqrt{RKS}}+ \frac{\sqrt{K}G^2}{\sqrt{RS}} + \frac{L^2 \sigma^2}{R^{3/2}K} + \frac{\sqrt{K}L^2}{R^{3/2} \sqrt{S}}\bigg).\nonumber
		\end{equation}
	\end{theorem}

	\noindent\textit{Proof.} Summing the above result for $r = [R]$ and multiplying both sides by $\frac{1}{C\beta \eta_g R}$, we have
	\begin{equation}
		\begin{split}
			& \frac{1}{R}\sum_{r=1}^{R}\E [\|F(w^{r+1})\|] \leq \frac{f(\tw^r ) - f(\tw^{r+1})}{C\beta \eta_g R}\\
			&+ \frac{1}{C}\bigg(10\beta^2 L^4 \eta_l^2 \rho^2 \sigma_l^2 + 35\beta^2 KL^2 \eta_l^2 (3\sigma_g^2 + 6L^2 \rho^2 ) + 28\beta^2 K^3 L^6 \eta_l^4 \rho^2 + 2K^2 L^4 \eta_l^2 \rho^2 + \frac{\beta L^3 \eta_g^2 \rho^2}{2KS}\sigma_l^2 \\
			& + \frac{\beta L\eta_g}{K^2 SN}\bigg(30NK^2 L^4 \eta_l^2 \rho^2 \sigma_l^2 + 270NK^3 L^2 \eta_l^2 \sigma_g^2 + 540 NK^2 L^4 \eta_l^2 \rho^2 + 72K^4 L^6 \eta_l^4 \rho^2 + 6NK^4 L^2 \eta_l^2 \rho^2 + 4NK^2 \sigma_g^2 + 3NK^2 L^2 \rho^2  \bigg)\bigg)\\
			& \leq \frac{F}{C\beta \eta_g R} + \frac{1}{C}\bigg(10\beta^2 L^4 \eta_l^2 \rho^2 \sigma_l^2 + 35\beta^2 KL^2 \eta_l^2 (3\sigma_g^2 + 6L^2 \rho^2 ) + 28\beta^2 K^3 L^6 \eta_l^4 \rho^2 + 2K^2 L^4 \eta_l^2 \rho^2 + \frac{\beta L^3 \eta_g^2 \rho^2}{2KS}\sigma_l^2 \\
			& + \frac{\beta L\eta_g}{K^2 SN}\bigg(30NK^2 L^4 \eta_l^2 \rho^2 \sigma_l^2 + 270NK^3 L^2 \eta_l^2 \sigma_g^2 + 540 NK^2 L^4 \eta_l^2 \rho^2 + 72K^4 L^6 \eta_l^4 \rho^2 + 6NK^4 L^2 \eta_l^2 \rho^2 + 4NK^2 \sigma_g^2 + 3NK^2 L^2 \rho^2  \bigg)\bigg), \nonumber
		\end{split}
	\end{equation}
	where the second inequality uses $F= f(\tw^0)-f^* \leq f(\tw^r )-f(\tw^{r+1})$. If we choose the learning rates $\eta_l = \frac{1}{\sqrt{RK}\beta L}$, $\eta_g = \frac{\sqrt{KS}}{\sqrt{R}\beta L}$ and perturbation amplitude $\rho$ proportional to the learning rate, e.g., $\rho = \frac{1}{\sqrt{R}}$, we have:
	\begin{equation}
		\begin{split}
			\frac{1}{R}\sum_{r=1}^{R}\E [\|F(w^{r+1})\|] & = \mathcal{O}\bigg(\frac{\beta FL}{\sqrt{RKS}}+ \frac{\sigma_g^2}{R} + \frac{\beta \sqrt{K}\sigma_g^2}{\sqrt{RS}} +\frac{\sqrt{KS}\sigma_g^2}{R^{3/2}} + \frac{L^2 \sigma_l^2}{R^2 K} + \frac{L\sigma_l^2}{R^2 \beta} + \frac{L^2 \sqrt{KS}\sigma_l^2}{R^{5/2}}\\
			&+ \frac{\beta L^2}{R^2} + \frac{\beta L^3}{R^2}+ \frac{L^2}{R^3 \beta^2} + \frac{L^2}{R^2 \beta^2} + \frac{L^3}{R^2 \beta \sqrt{KS}} + \frac{\sqrt{K}L^2}{R^{7/2}\sqrt{S}\beta^4} + \frac{K^{3/2}L}{R^{5/2}\sqrt{S}}\bigg),\nonumber
		\end{split}
	\end{equation}
	If the number of sampling clients are larger than the number of epochs, i.e., $S \geq K$, and omitting the larger order of each part, we have:
	\begin{equation}
		\frac{1}{R}\sum_{r=1}^{R}\E [\|F(w^{r+1})\|] = \mathcal{O}\bigg(\frac{\beta FL}{\sqrt{RKS}}+ \frac{\sqrt{K}G^2}{\sqrt{RS}} + \frac{L^2 \sigma^2}{R^{3/2}K} + \frac{\sqrt{K}L^2}{R^{3/2} \sqrt{S}}\bigg).\nonumber
	\end{equation}
	This completes the proof. \hfill~$\Box$

	\section{Experimental Setup}\label{Sec:Setup}
	\begin{table}[htbp]
		\centering
		\caption{Datasets and models.}
		\begin{adjustbox}{max width=\textwidth}
			\begin{tabular}{ccccc}
				\toprule
				Dataset   & Task                              & Clients & Total samples & Model                     \\\midrule
				EMNIST \cite{cohen2017emnist}   & Handwritten character recognition & 100/50  & 81,425        & 2-layer CNN + 2-layer FFN \\
				CIFAR-10 \cite{krizhevsky2009learning}  & Image classification             & 100/50  & 60,000        & ResNet-18  \cite{he2016deep}               \\
				CIFAR-100 \cite{krizhevsky2009learning}  & Image classification             & 100/50  & 60,000        & ResNet-18  \cite{he2016deep}              \\\toprule
			\end{tabular}
		\end{adjustbox}
		\label{Tab:dataset}
	\end{table}
	
	We ran the experiments on a CPU/GPU cluster, with RTX 2080Ti GPU, and used PyTorch \cite{paszke2019pytorch} to build and train our models. The description of datasets is introduced in Table~\ref{Tab:dataset}. 
	
	\subsection{Dataset Description}
	EMNIST \cite{cohen2017emnist} is a 62-class image classification dataset. In this paper, we use 20\% of the dataset, and we divide this dataset to each client based on Dirichlet allocation of parameter $0.6$ over 100 client by default. We train the same CNN as in \cite{reddi2020adaptive, dieuleveut2021federated}, which includes two convolutional layers with 3$\times$3 kernels, max pooling, and dropout, followed by a 128 unit dense layer.
	
	CIFAR-10 and CIFAR-100 \cite{krizhevsky2009learning} are labeled subsets of the 80 million images dataset. They both share the same 60,000 input images. CIFAR-100 has a finer labeling, with 100 unique labels, in comparison to CIFAR-10, having 10 unique labels. The Dirichlet allocation of these two datasets are also 0.6. For both of them, we train ResNet-18 \cite{he2016deep} architecture.
	
	\subsection{Hyperparameters}
	For each algorithm and each dataset, the learning rate was set via grid search on the set $\{10^{-0.5}, 10^{-1}, 10^{-1.5}, 10^{-2}\}$. \texttt{FedCM}, \texttt{MimeLite} and \texttt{MoFedSAM} momentum term $\beta$ was tuned via grid search on $\{0.01, 0.1, 0.2, 0.5, 1\}$. The global learning rate $\eta_g = 1$, and local learning rate $\eta_l = 0.1$ by default.

	\section{Additional Experiments}\label{Sec:add}
	\subsection{Training accuracy on different datasets}
	
	\begin{figure*}[htbp]
		\centering
		\begin{minipage}{0.32\columnwidth}
			\centering
			\includegraphics[width=\textwidth]{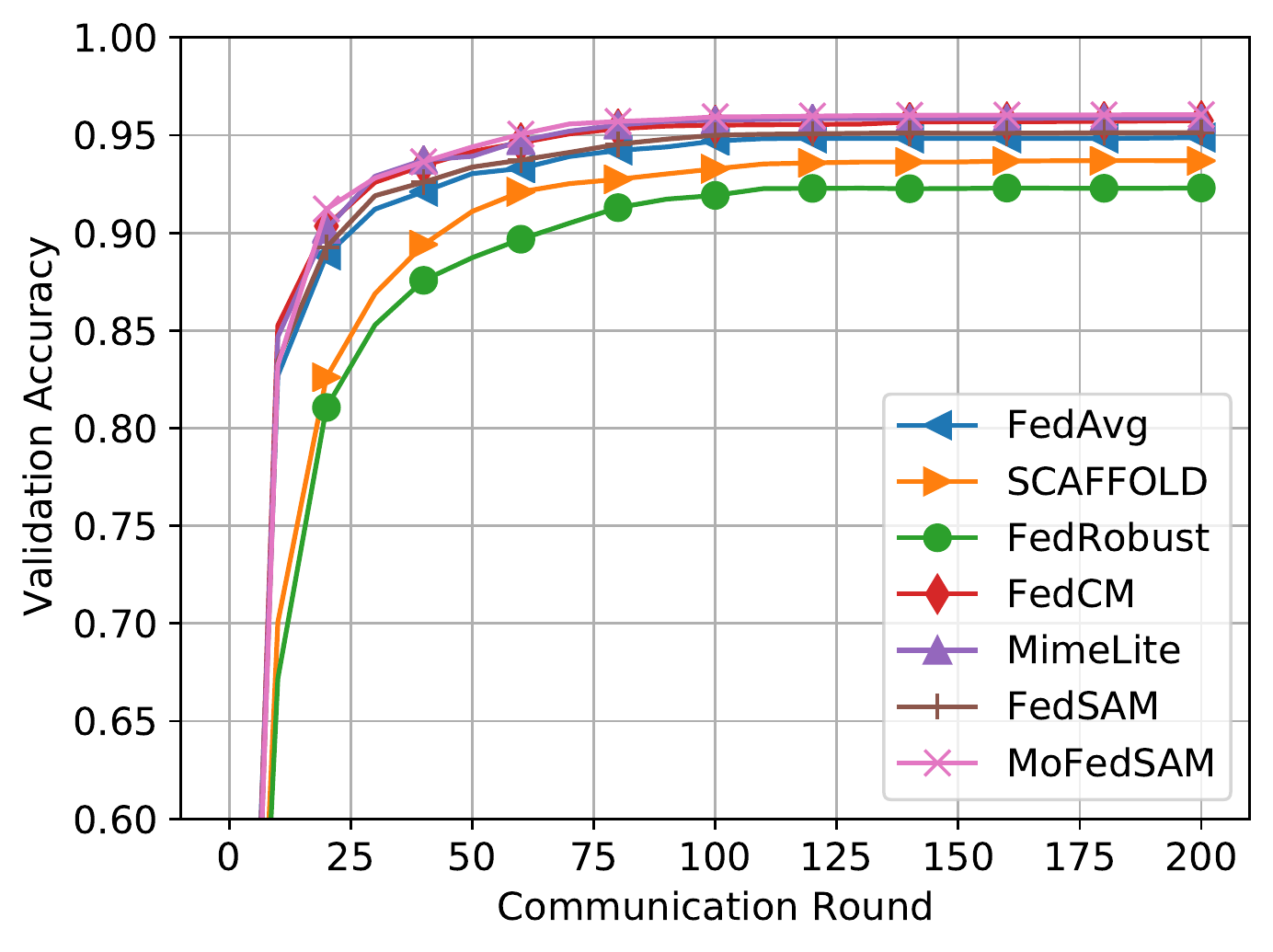}
			\subcaption[first]{EMNIST dataset.}
		\end{minipage}
		\hfill
		\begin{minipage}{0.32\columnwidth}
			\centering
			\includegraphics[width=\textwidth]{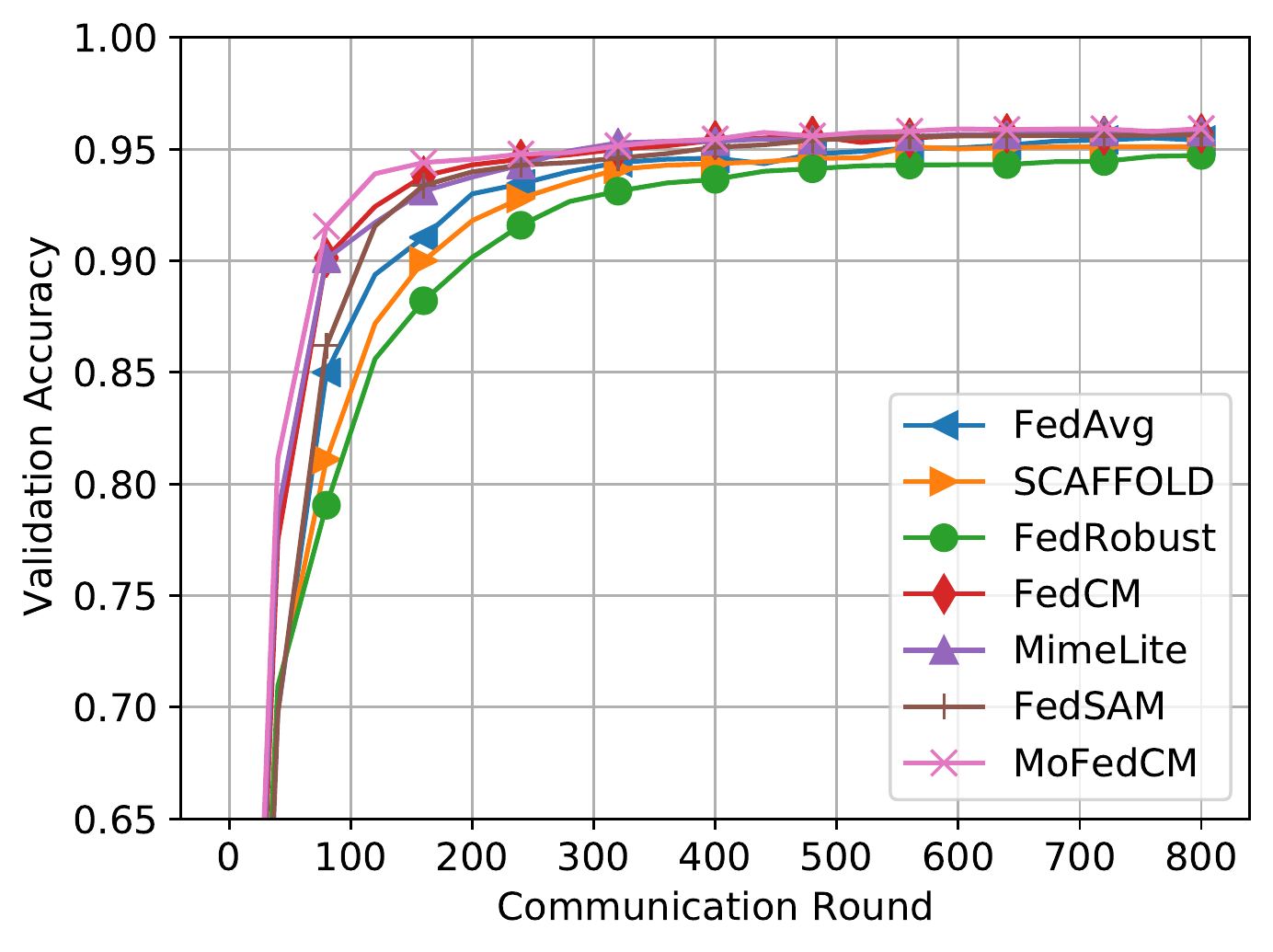}
			\subcaption[second]{CIFAR-10 dataset.}
		\end{minipage}%
		\hfill
		\begin{minipage}{0.32\columnwidth}
			\centering
			\includegraphics[width=\textwidth]{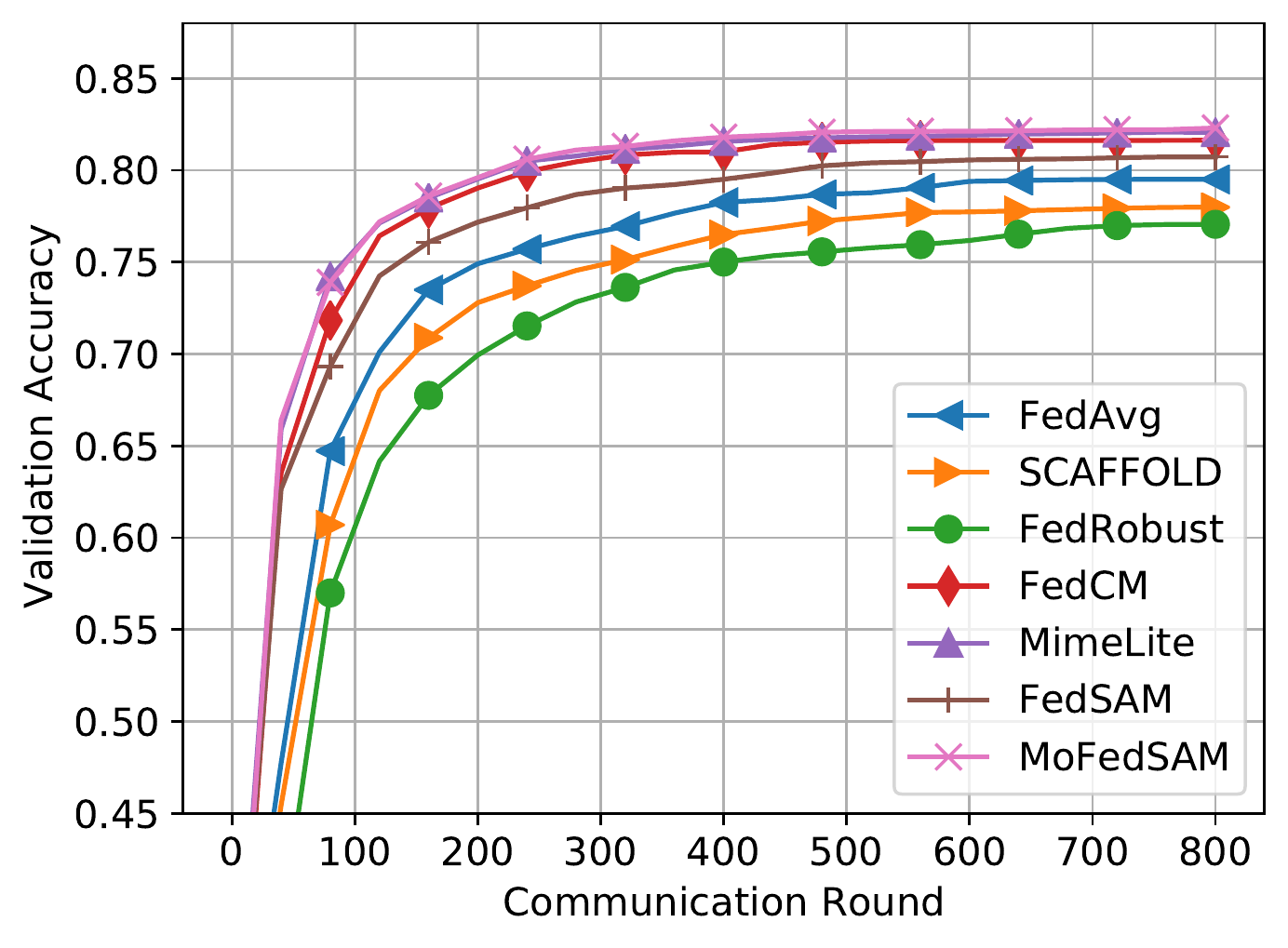}
			\subcaption[third]{CIFAR-100 dataset.}
		\end{minipage}
		\caption{Training accuracy on different datasets.}
		\label{Fig:datasettrain}
	\end{figure*}
	
	\begin{table*}[htbp]
		\centering
		\caption{Impact of the heterogeneity on EMNIST dataset (IID, Dirichlet 0.6 and Dirichlet 0.3).}
		\begin{adjustbox}{max width=0.97\textwidth}
			\begin{tabular}{*{10}{c}}
				\toprule 
				\multirow{2}{*}{Algorithm}       & \multicolumn{3}{c}{IID}              & \multicolumn{3}{c}{Dirichlet 0.6}            & \multicolumn{3}{c}{Dirichlet 0.3}           \\\cmidrule{2-10}
				& Train          & Validation     & Round & Train  & Validation     & Round & Train     & Validation     & Round \\ \midrule
				\texttt{FedAvg} & 96.98(0.73) & 89.95(1.95) & 32 & 95.07 (0.94) & 84.38 (4.03) & 43 & 93.66(1.27) & 82.83(4.42) & 61   \\
				\texttt{SCAFFOLD}  & 96.04(1.01) & 88.79(2.38) & 51 & 93.85 (1.31) & 84.09 (4.56) & 69 & 92.85(1.68) & 82.01(4.95) & 88 \\
				\texttt{FedRobust} & 95.63(0.56) & 87.67(1.63) & 66 & 93.17 (0.62) & 83.70 (3.37) & 91 & 92.10(1.00) & 81.80(3.79) & 103 \\
				\texttt{FedCM}     & 97.47(0.87) & 91.13(2.07) & 18 & 96.22 (1.16) & 84.85 (4.22) & 25 & 94.83(1.29) & 83.09(4.58) & 47 \\
				\texttt{MimeLite}  & 97.26(0.85) & 91.29(2.11) & 16 & 95.73 (0.49) & 84.88 (3.04) & 38 & 94.90(1.33) & 83.14(4.55) & 46  \\
				\texttt{FedSAM}    & 97.42(0.49) & 90.22(1.50) & 22 & 96.16 (1.14) & 84.75 (4.11) & 28 & 94.32(0.91) & 82.97(3.56) & 53 \\
				\texttt{MoFedSAM}  & 97.58(0.51) & 91.52(1.53) & 13 & 96.42 (0.42) & 85.07 (2.95) & 24 & 94.98(0.95) & 83.28(3.59) & 41 \\\toprule
			\end{tabular}
		\end{adjustbox}
		\label{Tab:heterogeneousemnist}
	\end{table*}
	
	\begin{table*}[htbp]
		\centering
		\caption{Impact of the heterogeneity on CIFAR-100 dataset (IID, Dirichlet 0.6 and Dirichlet 0.3).}
		\begin{adjustbox}{max width=0.97\textwidth}
			\begin{tabular}{*{10}{c}}
				\toprule 
				\multirow{2}{*}{Algorithm}       & \multicolumn{3}{c}{IID}              & \multicolumn{3}{c}{Dirichlet 0.6}            & \multicolumn{3}{c}{Dirichlet 0.3}           \\\cmidrule{2-10}
				& Train & Validation  & Round & Train & Validation & Round & Train & Validation & Round \\ \midrule
				\texttt{FedAvg} & 84.68 (1.46) & 58.97 (3.56) & 253 & 79.57 (1.84) & 53.57 (5.40) & 302 & 77.61 (1.99) & 51.22 (6.17) & 593  \\
				\texttt{SCAFFOLD}  & 83.41 (2.07) & 57.16 (4.32) & 327 & 78.49 (2.02) & 51.49 (5.87) & 551 & 76.30 (2.67) & 48.89 (6.59) & - \\
				\texttt{FedRobust} & 82.58 (1.35) & 55.87 (3.35) & 378 & 76.80 (1.70) & 49.06 (4.75) & 893 & 75.26 (1.87) & 47.92 (5.86) & - \\
				\texttt{FedCM}     & 87.05 (1.48) & 59.64 (3.73) & 149 & 82.46 (2.00) & 55.73 (5.11) & 189 & 79.91 (2.02) & 52.57 (6.28) & 410 \\
				\texttt{MimeLite}  & 87.42 (1.56) & 59.87 (3.67) & 143 & 82.53 (2.08) & 55.82 (5.04) & 182 & 79.96 (2.00) & 52.60 (6.31) & 397 \\
				\texttt{FedSAM}  & 85.65 (1.27) & 59.11 (3.11) & 228 & 81.04 (1.59) & 54.69 (4.36) & 245 & 78.05 (1.71) & 51.78 (5.43) & 561  \\
				\texttt{MoFedSAM}  & 87.82 (1.32) & 60.02 (3.20) & 129 & 82.62 (1.53) & 56.60 (4.42) & 124 & 80.09 (1.77) & 52.90 (5.62) & 373 \\\toprule
			\end{tabular}
		\end{adjustbox}
		\label{Tab:heterogeneouscifar100}
	\end{table*}
	
	Figure~\ref{Fig:datasettrain} shows that the training accuracy on different datasets. Comparing with the validation accuracy results in Figure~\ref{Fig:dataset}, the performance divergence is not clear. The reason is because the global model is easy to overfit the training dataset. Since the distribution of validation dataset on each client is different from training datasets, compared benchmarks perform less generalization. This indicates that our proposed algorithms benefits. Although \texttt{FedSAM} does not show better performance compared to the momentum FL, i.e., \texttt{FedCM} and \texttt{MimeLite}, it saves more transmission costs, since it does not need to download $\Delta^r$. For example, on CIFAR-100 dataset, FedCM achieves 85.26\% training accuracy with 4.41\% deviation of local models, however, it obtains 54.09\% validation accuracy with 14.38\% deviation. For \texttt{MoFedSAM} algorithm, it can achieve 86.02\% training accuracy with 3.23\% deviation of local models, and 55.13\% validation accuracy with 3.25\% deviation of local models.
	
	Tables~\ref{Tab:heterogeneous}, \ref{Tab:heterogeneousemnist} and \ref{Tab:heterogeneouscifar100} aim to show the impact of heterogeneous degrees of FL. From these results, we can clearly see that increasing the degree of heterogeneity makes huge degradation of learning performance. However, it does not effect the training accuracy significantly. For example, on CIFAR-10 dataset, \texttt{FedSAM} obtains 95.42\%, 94.20\%, and 92.90\% training accuracy, when heterogeneity is IID, Dirichlet 0.6 and Dirichlet 0.3, and 87.36\%, 82.55\% and 79.82\% for validation accuracy. More specifically, the influence of heterogeneity for our proposed algorithms are less than compared benchmarks, which is due to the fact that the more generalized global model, the less impact of distribution shift.

	\subsection{Impact of hypeparameters}
	\begin{figure*}[htbp]
		\centering
		\begin{minipage}{0.24\columnwidth}
			\centering
			\includegraphics[width=\textwidth]{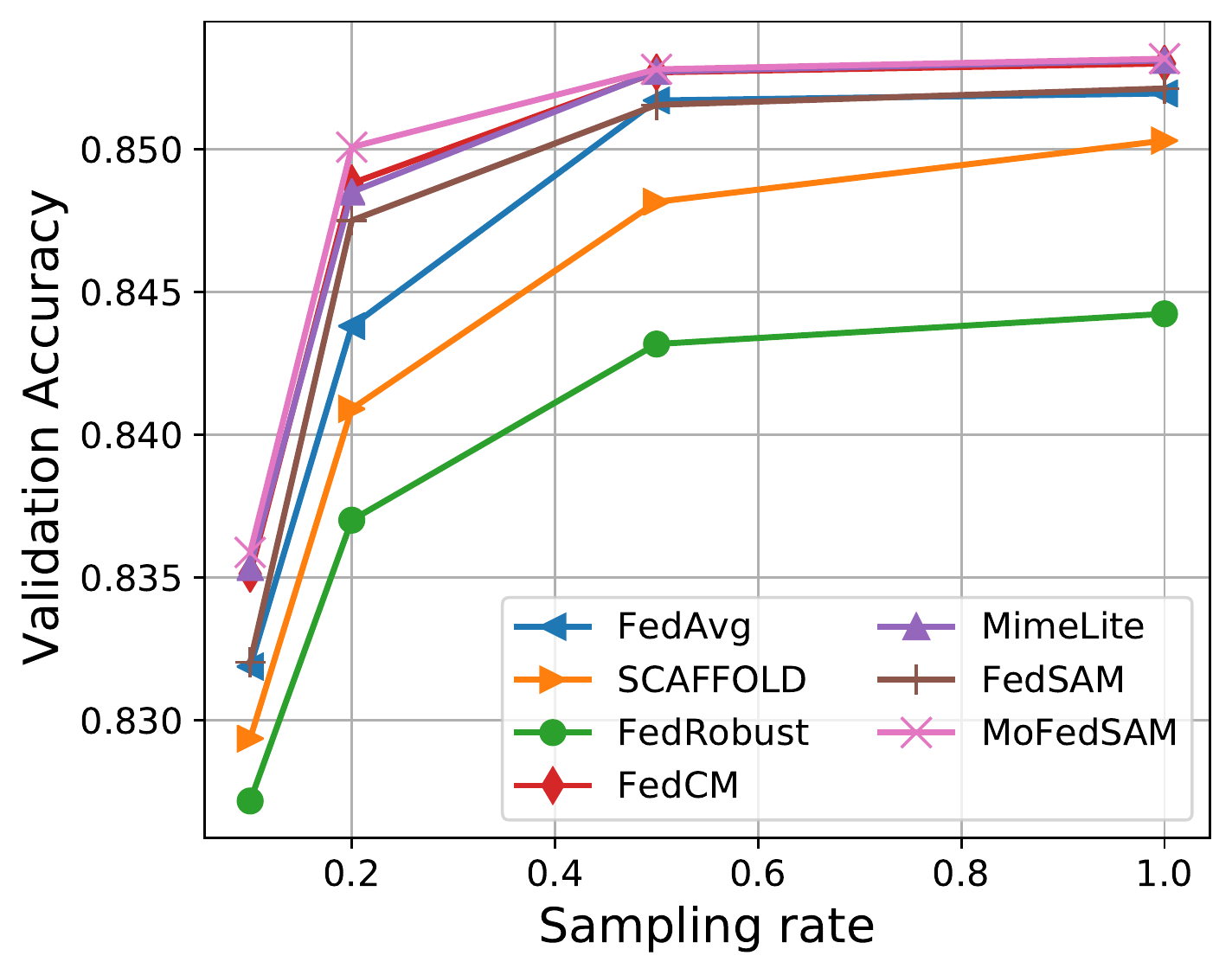}
			\subcaption[first]{Impact of $S$.}
			\label{Fig:rhocifaremnist}
		\end{minipage}%
		\hfill
		\begin{minipage}{0.24\columnwidth}
			\centering
			\includegraphics[width=\textwidth]{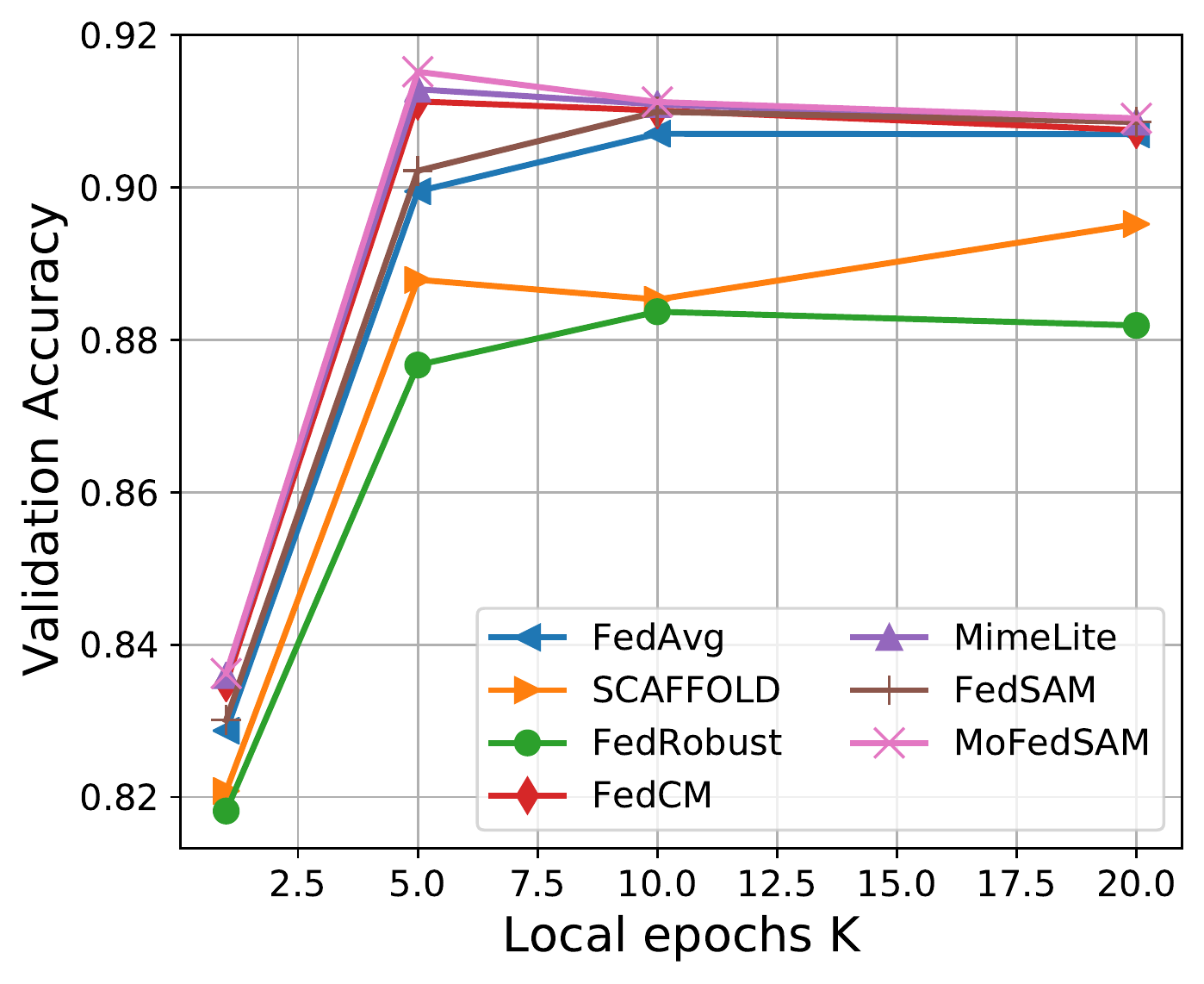}
			\subcaption[second]{Impact of $K$.}
			\label{Fig:Kemnist}
		\end{minipage}
		\hfill
		\begin{minipage}{0.24\columnwidth}
			\centering
			\includegraphics[width=\textwidth]{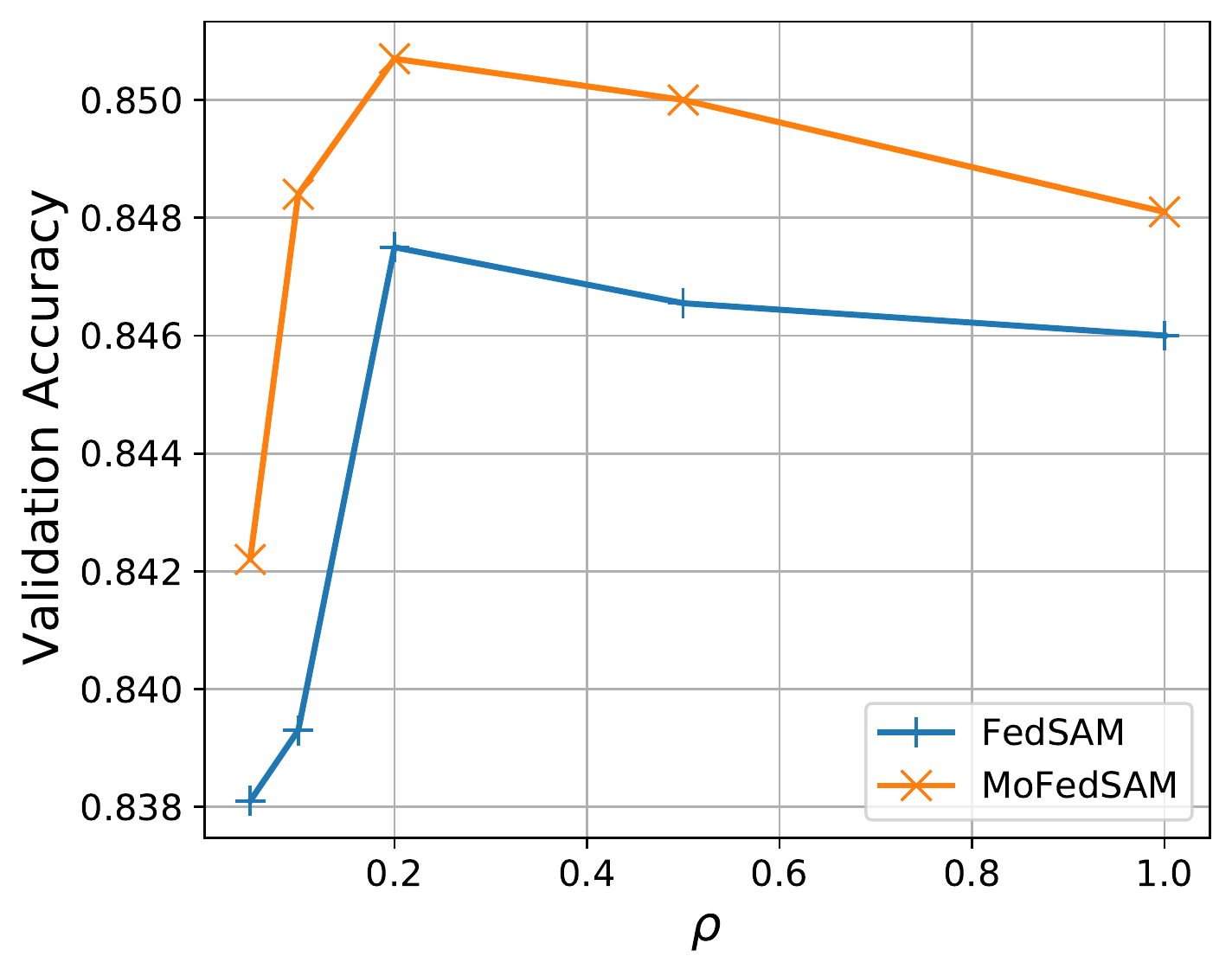}
			\subcaption[third]{Impact of $\rho$.}
			\label{Fig:rhoemnist}
		\end{minipage}
		\hfill
		\begin{minipage}{0.24\columnwidth}
			\centering
			\includegraphics[width=\textwidth]{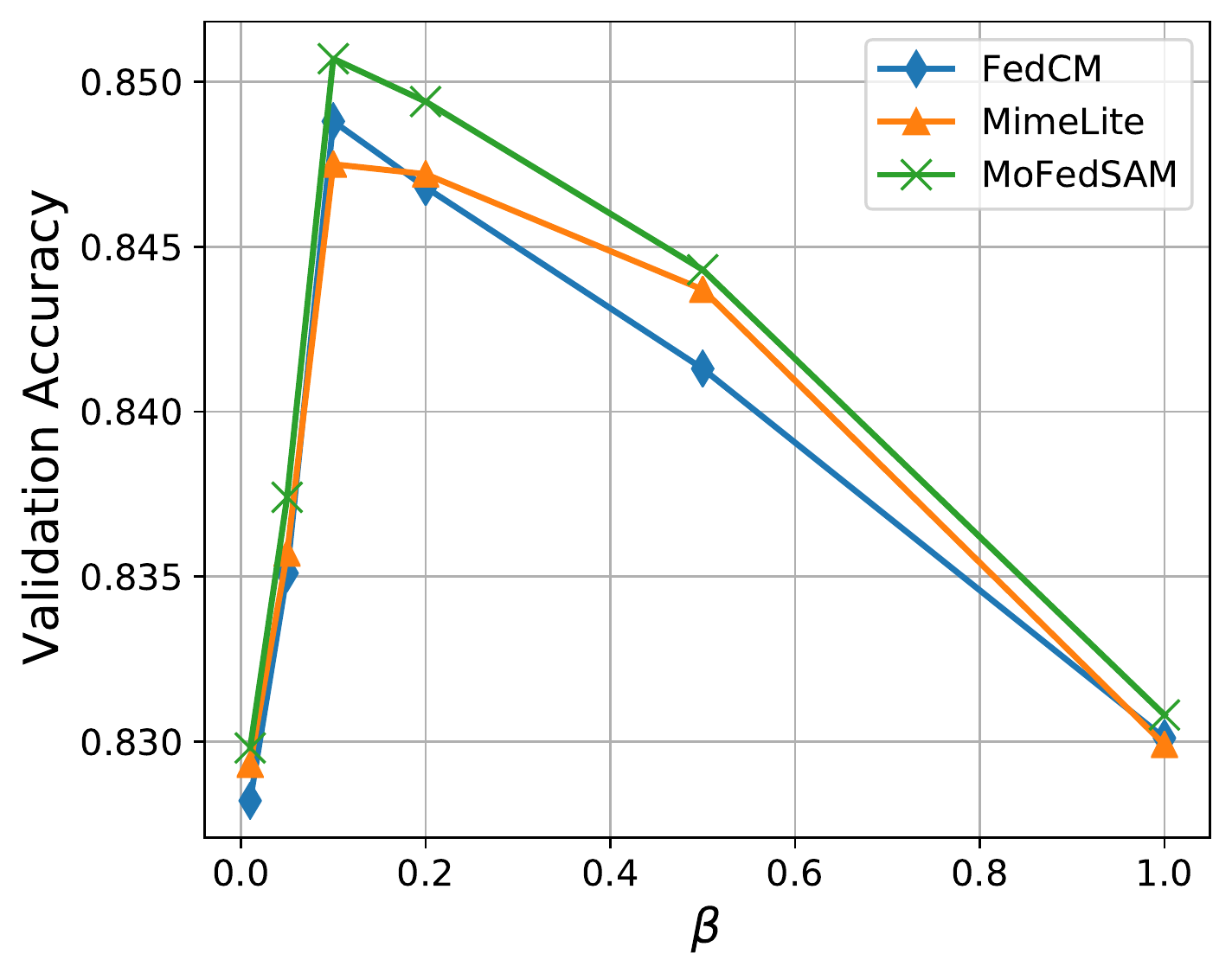}
			\subcaption[fourth]{Impact of $\beta$.}
			\label{Fig:betaemnist}
		\end{minipage}
		\caption{Impacts of different parameters on EMNIST dataset.}
		\label{Fig:impactemnist}
	\end{figure*}
	
	\begin{figure*}[htbp]
		\centering
		\begin{minipage}{0.24\columnwidth}
			\centering
			\includegraphics[width=\textwidth]{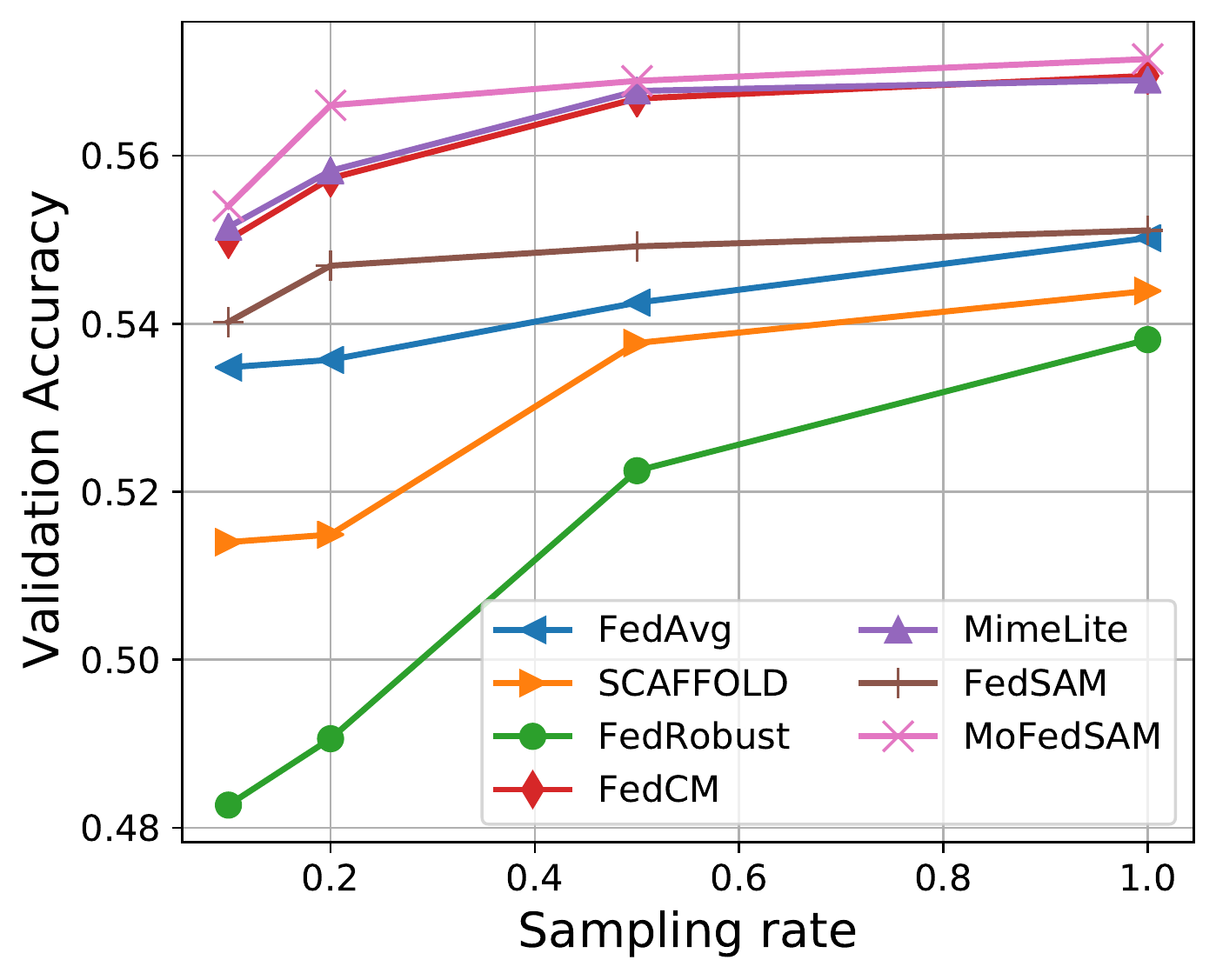}
			\subcaption[second]{Impact of $S$.}
			\label{Fig:rhocifar100}
		\end{minipage}%
		\hfill
		\begin{minipage}{0.24\columnwidth}
			\centering
			\includegraphics[width=\textwidth]{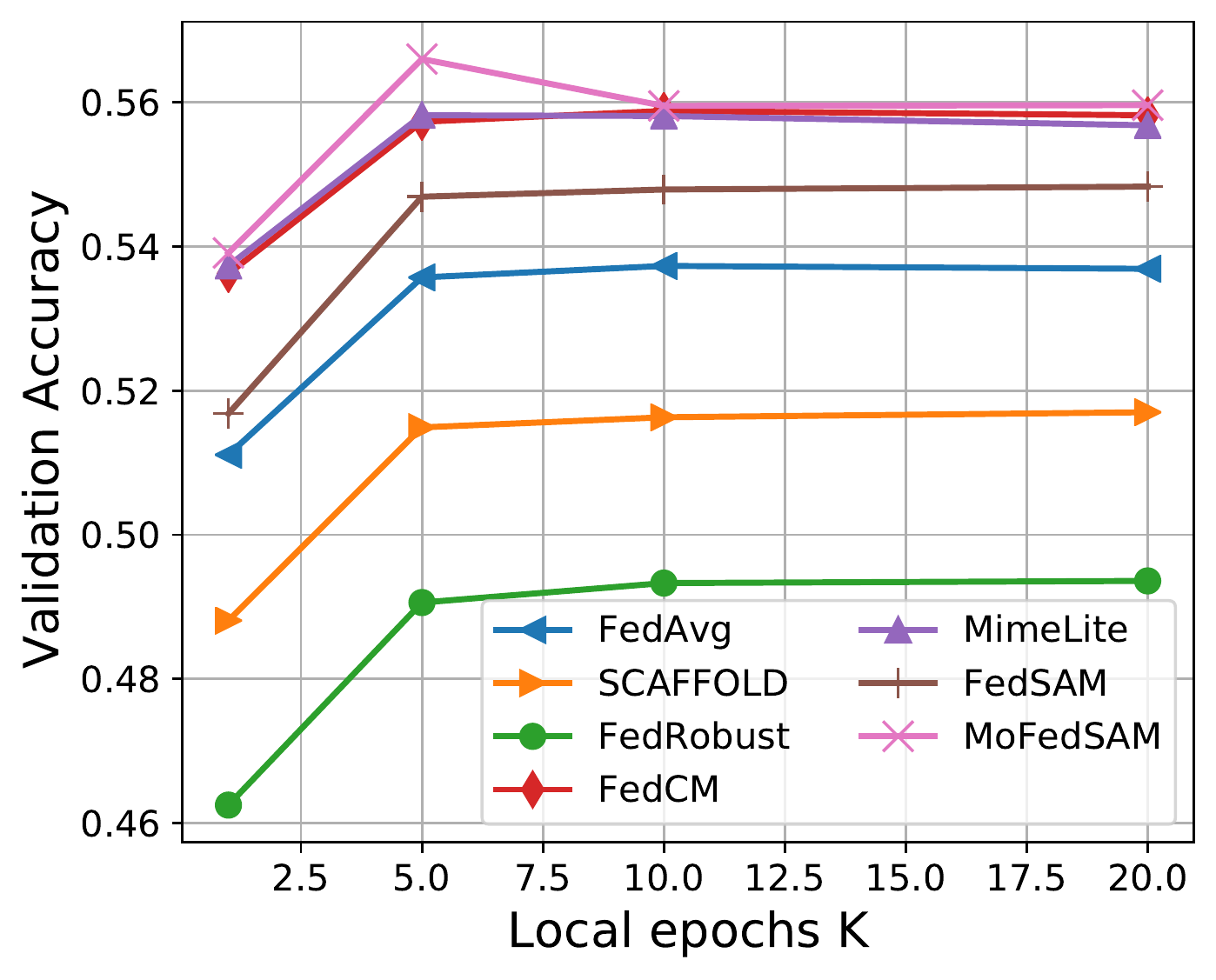}
			\subcaption[third]{Impact of $K$.}
			\label{Fig:Kcifar100}
		\end{minipage}
		\hfill
		\begin{minipage}{0.24\columnwidth}
			\centering
			\includegraphics[width=\textwidth]{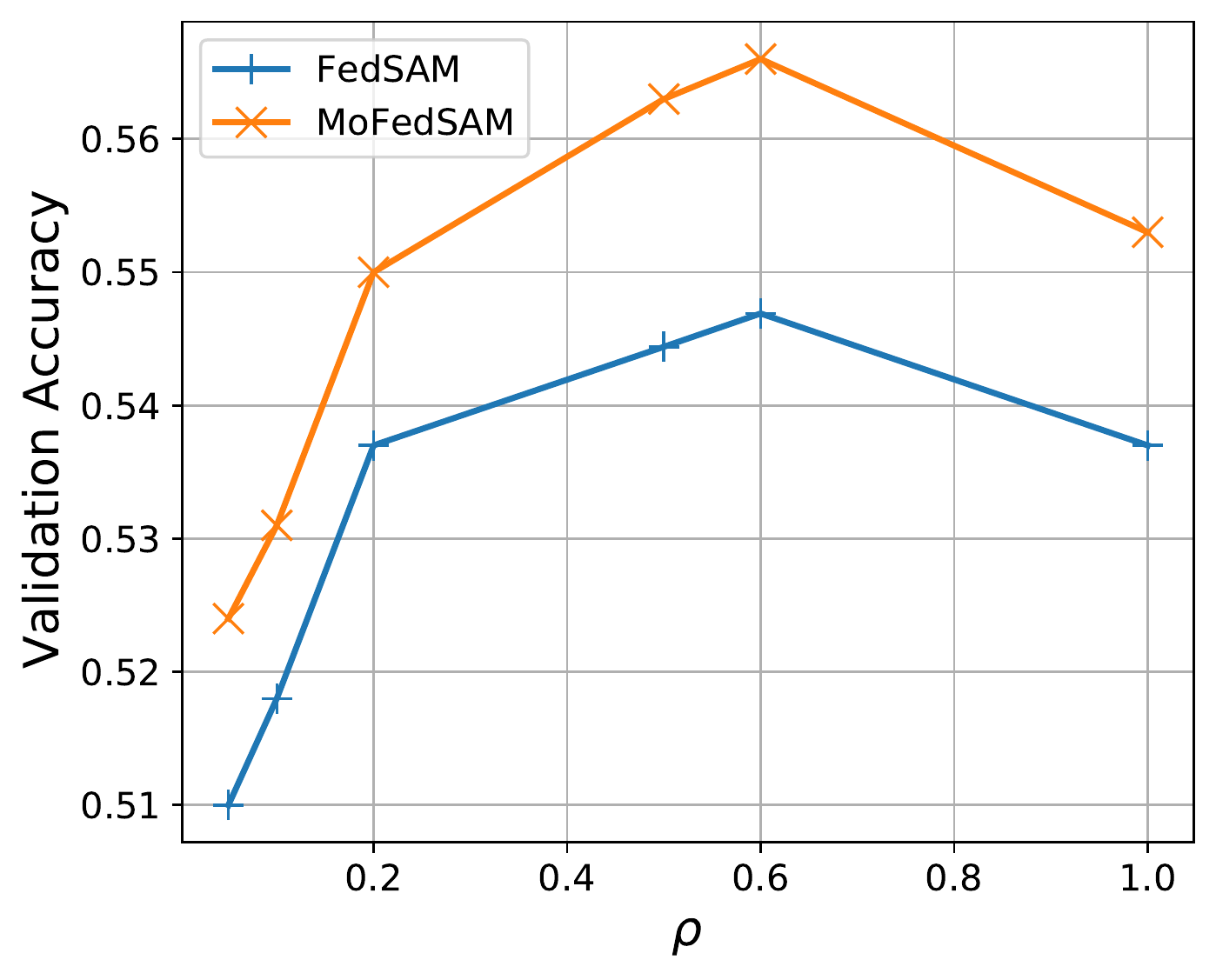}
			\subcaption[fourth]{Impact of $\rho$.}
			\label{Fig:Scifar100}
		\end{minipage}
		\hfill
		\begin{minipage}{0.24\columnwidth}
			\centering
			\includegraphics[width=\textwidth]{figures/rho_cifar100.pdf}
			\subcaption[fourth]{Impact of $\beta$.}
			\label{Fig:betacifar100}
		\end{minipage}
		\caption{Impacts of different parameters on CIFAR-100 dataset.}
		\label{Fig:impactcifar100}
	\end{figure*}
	
	Figures~\ref{Fig:impactcifar10}-\ref{Fig:impactcifar100} aim to show the impacts of different hyperparameters, i.e., the number of participated clients $S$ in each communication round, the number of local epochs $K$, the perturbation control parameter $\rho$ of SAM optimizer, and the momentum parameter $\beta$. We can see that increasing $S$ can improve the performance. However, increasing $K$ cannot guarantee increasing the performance. For $\rho$ and $\beta$, they depend on the different algorithms and datasets. By grid searching, it is not difficult to find the suitable value to optimize the performance.
	
\end{document}